\pdfoutput=1

\documentclass[11pt]{article}

\usepackage{styles/acl}

\usepackage{times}
\usepackage{latexsym}

\usepackage[T1]{fontenc}

\usepackage[utf8]{inputenc}

\usepackage{microtype}

\usepackage{todonotes}
\usepackage{subcaption}                     
\usepackage{graphicx}
\usepackage[nolist]{acronym}                
\usepackage{hyperref}                       
\usepackage{xurl}                           
\usepackage{booktabs}                       
\usepackage{multirow}                       
\usepackage{amsfonts}                       
\usepackage{nicefrac}                       
\usepackage{enumitem}                       
\usepackage{pgfplots}                       
\pgfplotsset{compat=1.17} 
\usepackage{pgfplotstable}

\author{Daniele Giofr\'e \and Sneha Ghantasala \\
  Thomson Reuters Labs \\
  \texttt{daniele.giofre} \and \texttt{sneha.ghantasala@thomsonreuters.com} \\
}

\title{Legal-HNet: Mixing Legal Long-Context Tokens with Hartley Transform}

\begin{document}
\maketitle


\begin{acronym}[UMLX]
    \acro{MLM}{Masked Language Modeling}
    \acro{RTD}{Replaced Token Detection}
    \acro{GAN}{Generative Adversarial Network}
    
    \acro{US}{United States}
    \acro{EU}{European Union}
    \acro{ECtHR}{European Court to Human Rights}

    \acro{NLP}{Natural Language Processing}
    \acro{ML}{Machine Learning}
    
    \acro{LM}{Language Model}
    \acro{SOTA}{state-of-the-art}
    
    \acro{BERT}{Bidirectional Encoder Representations from Transformers}
    
    \acro{TC}{Text Classification}
    \acro{QA}{Question Answering}
    \acro{NER}{Named Entity Recognition}
    
    \acro{FNet}{Fourier Network}
    \acro{HNet}{Hartley Network}
    \acro{NLP}{Natural Language Processing}
    \acro{FT}{Fourier Transform}
    \acro{HT}{Hartley Transform}
    \acro{FCT}{Fourier Cosine Transform}
    \acro{FFT}{Fast Fourier Transform}
    \acro{FHT}{Fast Hartley Transform}
    \acro{RAED}{Right-side Attention Encoder-Decoder}
    \acro{HF}{HuggingFace}
    \acro{LT}{Laplace Transform}
    \acro{HP}{HyperParameters}
    \acro{RFT}{Real Fourier Transform}
    \acro{IFT}{Imaginary Fourier Transform}
    \acro{FF}{Feed-Forward}
    
    \acro{CNN}{Convolutional Neural Network}
    \acro{NN}{Neural Network}
    \acro{ED}{Encoder-Decoder}
\end{acronym}

\begin{abstract}

Since its introduction, the transformers architecture has seen great adoption in NLP applications, but it also has limitations. Although the self-attention mechanism allows for generating very rich representations of the input text, its effectiveness may be limited in specialized domains such as legal, where, for example, language models often have to process very long texts. In this paper, we explore alternatives to replace the attention-based layers with simpler token-mixing mechanisms: Hartley and Fourier transforms.  Using these non-parametric techniques, we train models with long input documents from scratch in the legal domain setting. We also introduce a new hybrid Seq2Seq architecture, a no-attention-based encoder connected with an attention-based decoder, which performs quite well on existing summarization tasks with much less compute and memory requirements. We believe that similar, if not better performance, as in the case of long correlations of abstractive text summarization tasks, can be achieved by adopting these simpler infrastructures. This not only makes training models from scratch accessible to more people, but also contributes to the reduction of the carbon footprint during training.

\end{abstract}

\acresetall 

\section{Introduction}

The Transformer architecture has gained rapid and widespread adoption in the \ac{NLP} field thanks to its outstanding results. At its core is a self-attention mechanism - an inductive bias that links each token in the input through a weighted basis for the relevance of every other token. Each hidden unit is represented on the basis of the hidden units in the previous layer. Many papers have described this abstraction, on an increasingly complex and higher-order basis than the previous one, as the main mechanism to define more complex features such as syntactic and semantic relations \cite{ tenney2019bert, vig-belinkov-2019-analyzing,clark-etal-2019-bert, voita-etal-2019-analyzing}.

Although the self-attention mechanism often works well, it is greedy for both computational and memory resources, and there are many studies in the literature where different  techniques have been adopted to replace the layer with quadratic time complexity \cite{beltagy2020longformer,child2019generating,choromanski2020masked,choromanski2021rethinking, dosovitskiy2021image,jiao2020tinybert,pmlr-v119-katharopoulos20a,kim2020fastformers,kitaev2020reformer,vyas2020fast, wang2020linformer}.

Our starting point is the work by \citet{lee-thorp_fnet_2021}, the so-called \ac{FNet}, with the aim of addressing the long-document problem in legal texts. They prove that Transformer encoder architectures can be sped-up, with limited accuracy costs, by replacing the self-attention sub-layers with simple linear \ac{FT} that “mix” input tokens.

Although there are many other studies addressing the replacement of the self-attention sub-layer with a linear transformation, we believe that \ac{FNet} provides a simple and elegant way to address it, achieving the best compromise between speed and accuracy.

In this paper, we explore the diverse Fourier-class Transform using a different real output propagation to avoid the complex number and to improve the original \ac{RFT}  architecture made in \citet{lee-thorp_fnet_2021}.
We show that by replacing the \ac{RFT} layer with a \ac{HT} \cite{bracewell_computing_1995} we obtain better metrics than \ac{FNet} for most of the tasks where the two \ac{NN} architectures were compared. We call this \ac{LM} architecture \ac{HNet}\footnote{\citet{lee-thorp_fnet_2021} have tried a \ac{HT}, and because, in the paper, they got no improvement in GLUE benchmark, they gave little room for further testing}.

Given that \ac{HNet} does not require any further computational steps in comparison with \ac{FNet}, it can be used for longer sequence lengths at the same computational cost as BERT-based solutions, which can be particularly effective both to solve the problem of long documents in legal domain, and to afford high training and inference costs or the creation of ad-hoc language models (i.e. with a specific vocabulary and domain). Furthermore, in our downstream tasks, \ac{HNet} reduces the performance gap introduced by the original \ac{FNet} architecture. 

We trained three new legal domain language models using the \ac{FNet} and \ac{HNet} architectures using different sequence lengths as input (4096 and 8192 tokens for FNet and 4096 tokens for HNet). We evaluated them for two public summarization datasets, BillSum and PubMed, and for classification and question \& answering tasks using the LexGlue benchmark. We added the latter benchmark for the sake of completeness, since it is the equivalent of the  GLUE benchmark in the legal \ac{NLP} world (it allows us to make an analogy with the GLUE benchmark reported in in the FNet paper). Notably, the aforementioned benchmark is based on short texts, while the motivation of this study lies in improving performance for long-text problems.





 Indeed, to test the ability to capture long-distance dependencies in the text, we mainly evaluated \ac{HNet} (and the two \ac{FNet} \ac{LM}s) on the task of automatic (abstractive) summarization, as done in other recent studies \cite{joel_paper, zhang_pegasus_2020}. Abstractive summarization refers to the task of capturing the most important concepts/ideas from the (long) document and then rewriting it in a shorter passage in a grammatical and logically coherent way \cite{chen_multi-task_2019}. Since an abstractive summarisation task requires an \ac{ED} model architecture and few studies have been carried out in the literature regarding those based on Fourier-class Transformers \cite{kiruluta_new_2021}, we have also proposed a new Seq2Seq architecture that adopts the lightweight implementation provided by Fourier-class based Encoders and utilises the performance of the attention based Decoders.

We used the BillSum benchmark, as a domain-specific summarization task; and the PubMed benchmark, to evaluate the model's ability outside the legal context (i.e., in the biomedical context). We decided on the biomedical domain because like the legal one, it requires a specific and complex vocabulary so as to prove that our \ac{LM}s can be open-domain. The proposed models achieved in both cases better metrics than a Transformer architecture and approached the recent \ac{SOTA} (i.e. BudgetLongformer)  the well-known previous one  (i.e. PEGASUS, which was trained using the Gap-Sentences generation task and with more resources and data respect our \ac{LM}s in the pretraining phase\footnote{our \ac{LM}s has only been pretrained on 450k legal documents as well as for the tokenizer.}), see Tables \ref{tab:billsum_results}, and \ref{tab:pubmed_results}. 
Furthermore, both BudgetLongformer and Pegasus, in addition to having many more trainable parameters and a greater environmental impact, they also require a larger memory allocation, which limits the maximum number of tokens provided as input by both the encoder and the decoder.

It is important to note that this performance was achieved with computationally and memory-light allocation models, i.e. \ac{HNet} and \ac{FNet}, trained for minimal pretraining phase: one million steps and with a data catalogue of less than 500k documents (only legal documents), 
which means that \ac{HNet} can be trained faster and cheaper using less documents. For example, RoBERTa \cite{liu_roberta_2019} or PEGASUS \cite{zhang_pegasus_2020}, which are both models that \ac{HNet} and \ac{FNet} are compared to, required substantially longer training and greater related costs. RoBERTa was trained for 1024 GPU days, whereas both \ac{HNet} and \ac{FNet} only needed 20 GPU days each (estimated using a 16GB NVIDIA V100 GPUs)



\subsection*{Contributions}
The contributions of this paper are six-fold:
\vspace{-4mm}
\begin{itemize}[leftmargin=8pt, itemsep=0em]
    \item We investigated further into the use of \ac{HT} instead of \ac{FT} in the case of \ac{LM}s in the legal domain, calling it \ac{HNet}, and finding that it performs slightly better with equal CO2 impact and the capability of handling long documents.
    \item We have trained from scratch and built 3 new legal \ac{LM}s for handling long legal documents: one \ac{HNet} with input sequence up to 4096 tokens and two \ac{FNet}(s) with input sequence up to 4096 and 8192 tokens.
    \item We proposed and evaluated a new Encoder-Decoder diagram where no-attention based encoders can be connected with attention based decoders, to achieve the best compromise in terms of efficiency (a light encoder) and quality (a performing decoder with cross attention).
    \item We approach \ac{SOTA} on two public benchmarks, BillSum and Pubmed, using in the worst scenario 50 percent less computational resources.
    \item We note that models based on Fourier-class transforms become truly beneficial when tasks require long documents, for example,  with an average length much longer than 512 tokens, both for obvious cost reasons and because of the reducing performance gap with respective attention mechanism-based models.
    \item  On the LexGLUE benchmark \cite{chalkidis_lexglue_2021}, despite the obvious emphasis on covering classification tasks for short documents, our models achieve equivalent or even better performance to those found by \citet{lee-thorp_fnet_2021}. Indeed, we measured a performance, with respect to the BERT-based model, of 92\% overall for the base FNet-4096, of 94\% for the base HNet-4096, and of 87\% for the \ac{LM} released by \citet{lee-thorp_fnet_2021}.
\end{itemize}

\subsection*{Main Research Questions}
In this work, we pose and examine six main research questions:


\noindent \textbf{RQ1}: \emph{Is it possible to create a domain specific (e.g. legal) FNet based \ac{LM} from scratch, reducing costs and carbon footprint compared with self-attention based \ac{LM}s?}

\noindent \textbf{RQ2}: \emph{Does using the Hartley transform instead of Fourier in the FNet based architecture to create a domain specific (e.g. legal) \ac{LM} from scratch lead to better results?}

\noindent
\textbf{RQ3}:
\emph{Is it possible to create a ``competitive'' Seq2Seq model, which can approach the performance of \ac{SOTA} ones, using Fourier/Hartley layers?}

\noindent \textbf{RQ4}:
\emph{How do our models compare with other models on the challenging summarization task? Particularly in the case of a legal domain-specific benchmark such as BillSum?}

\noindent \textbf{RQ5}:
\emph{ How well do our models generalize to other domains, for example in the biomedical domain, as evaluated by the PubMed summarization benchmark?}

\noindent \textbf{RQ6}:
\emph{How do our \acp{LM} compare with other models on the \acf{TC} benchmark LexGLUE?}

\section{Related Work}
\label{sec:related_work}

\subsection*{Hartley Transforms in Neural Networks}

As shown in \citet{lee-thorp_fnet_2021} , the transformer encoder architectures can be sped-up, with limited accuracy costs, by replacing the self-attention sub-layers with a simple linear transformation: Fourier transform. Fourier layers are basically a simpler token mixing mechanism. In the \ac{FNet} article, although \ac{HT} was tested on the GLUE benchmark \cite{wang_glue_2018}, it obtained comparable results to \ac{FT}, giving no room for further investigations. \ac{FT} acts on real-valued functions getting complex-valued functions. 
On the other hand, \ac{HT} is a way to transform real-valued functions to real-valued functions \cite{bracewell_computing_1995}. Hartley related transforms have also shown to perform better than their corresponding \ac{FT} counterparts in \ac{CNN} related optimizations \cite{hartley_stochastic_cnns, hartley_spectral_pooling_cnns}, and in  multi-task learning with multiple
languages and modalities \cite{lee-thorp_sparse_2022,armitage_mlm_2020}.

\subsection*{Domain-Specific Language Models}
Previous research has shown that domain specific pretraining of language models from scratch show better results compared to continual pretraining of general-domain language models \cite{domain_specific_pretraining_biomedical_nlp}. Domain-specific pretraining on datasets of specialized domains such as 
law \cite{chalkidis_legal-bert_2020, xiao_lawformer_2021}, 
biology \cite{lee_biobert_2019},
scientific articles \cite{beltagy_scibert_2019}, or
clinical documents \cite{li_clinical-longformer_2022}  have shown promising results.

\subsection*{Long Document Processing}
In the legal domain, texts tend to span multiple pages, ranging from 10s to 100s of pages, which translates to tens of thousands tokens. The quadratic time and memory requirement of the attention typically used in the transformer architecture \cite{vaswani_attention_2017} prohibits efficient processing of sequences longer than 512 tokens on current hardware.

This has led to research focusing on tweaking the attention layers by making the attention matrices sparse, like the efficient transformers \cite{tay_efficient_2020, child_generating_2019, beltagy_longformer_2020, zaheer_big_2021, roy_efficient_2021, kitaev_reformer_2020, tay_synthesizer_2021}. While other researchers focus on replacing the attention mechanism with simpler token mixing mechanisms, like \ac{FNet}s \cite{lee-thorp_fnet_2021}, and MLP-Mixer \cite{tolstikhin_mlp-mixer_2021,fusco_pnlp-mixer_2022}.

\section{Datasets}
\label{sec:datasets}
In this section, we briefly introduce the datasets used in our experiments.

\subsection{Pretraining}
Training a new language model from scratch requires documents in the range of hundreds of thousands with considerable diversity, as minimum range\footnote{Massive \ac{LM}s can be fed up-to hundreds of millions of documents \cite{brown_language_2020}}. Our domain specific legal language model uses 7 different datasets: 3 public datasets and 4 internal datasets. It was trained on a total of around half a million documents. Since different datasets have different number of documents, we tried to ensure the model is not biased towards a particular dataset by making sure the number of documents from the main datasets are in the similar range. All documents with less than 200 tokens are filtered out. CUAD \cite{hendrycks_cuad_2021}, Posture \cite{song_multi-label_2022} and EUR-Lex \cite{chalkidis_et_al_2019_eurlex57k_nodate} datasets are the three public datasets used in the pretraining. Since the EUR-Lex dataset is used in the LexGLUE benchmarking tasks, we use only the training and validation datasets of the same. For further information about the datasets refer Table \ref{tab:pretraining_data}.

\begin{table}[t]
\centering
\resizebox{\columnwidth}{!}{
    \begin{tabular}{lrrr}
    \toprule
    Pretraining Subset                    & Dataset Size & \# Doc Token Avg. & \# Documents (train/val) \\ 
    \midrule
    \bf Total                           & \bf 9.2GB  &\bf 2.7k & \bf 450k / 54.2k\\ 
    CL Opinions$^{1,a}$                         & 2.7GB  &  2.3k & 148k / 24.9k\\
    CL Docket Entries and Court Filings$^{1,a}$ & 2.3GB &  3.9k & 103k / 9.4k\\
    CUAD$^{2,a}$ \cite{hendrycks_cuad_2021}     & 1.6GB &  9.3k & 22.4k / 2.0k\\
    Practical Law Tax and Legal Commentary$^{1,b}$ & 1.5GB  &  2.0k & 100k / 12.4k\\
    Laws$^{1,a}$                                & 0.5GB  &  3.2k & 11.8k / 1.0k\\
    Posture50k$^{2,a}$ \cite{song_multi-label_2022}    & 0.3GB  &  1.3k & 20.1k / 1.7k\\
    EUR-Lex$^{2,b}$ \cite{chalkidis_et_al_2019_eurlex57k_nodate}   & 0.2GB  &  0.6k & 43.1k / 2.8k\\
    \bottomrule
    \end{tabular}
}
\caption{The datasets used for pretraining our models: (1) internal datasets; (2) public dataset; (a) US English; (b) UK English. 
CL is short for Court Listener. }
\label{tab:pretraining_data}
\end{table}

\subsection{BillSum}

\citet{kornilova_billsum_2019} introduced a legislative summarization dataset from 21K \ac{US} bills from 1993 to 2018. It is challenging due to the technical nature and complex structure of the bills. Additionally, the bills are rather long, ranging from 1k to 4k tokens with their summaries being up to $\sim$ 1k tokens long (see Appendix \ref{sec:data_details} for more details). 

\subsection{PubMed}
\citet{cohan_discourse-aware_2018} introduced another challenging summarization dataset in a specialized domain (scientific articles from the biomedical domain). It includes 133K scientific papers together with their abstracts in English. The papers are 3k words long on average and the summaries (abstracts) 200 words. Thus, similar to the BillSum dataset, this dataset is well suited as a test bed for methods capable of long document summarization. Note, that in this dataset the domain is vastly different from the legal domain (see Appendix \ref{sec:data_details} for more details).

\subsection{LexGLUE}
\citet{chalkidis_lexglue_2021} recently introduced a benchmark for the English legal domain called LexGLUE. LexGLUE contains six \ac{TC} tasks and one \ac{QA} task comprising diverse legal data such as \ac{US} court decisions and contracts, terms of service documents, \ac{EU} legislation and cases from the \ac{ECtHR}. There exists a public leaderboard of diverse models on GitHub\footnote{\url{https://github.com/coastalcph/lex-glue}}, with Legal-BERT \cite{chalkidis_legal-bert_2020} performing best. 

The LexGLUE benchmark focuses on evaluating \acp{LM} in legal \ac{TC} and \ac{QA} tasks, and is, now, one of the benchmark standards in the legal \ac{NLP}. For these reasons, we feel it is important for the community to evaluate our \ac{LM}s against that benchmark, despite the fact that, in LexGLUE, 4 out of 7 tasks involve documents with input lengths lower than 512 tokens on average. The remaining 3 tasks, the ECtHR A and B tasks and the SCOTUS tasks involve documents with a longer span, but  the median of the first two tasks is also less than 1000 tokens. 
Usually, legal documents are much longer than 512 tokens and thus this distribution might not be representative of real-world tasks. One can also speculate that shorter input length tasks may be better handled by short-input models (e.g., BERT, RoBERTa, Legal-BERT, etc.).

\section{Neural Network Architecture}
\label{sec:nn_architecture}
In this section, we describe the \ac{LM} architectures used.  
Please, see the Appendices \ref{appendix:pretraining}
for the setup of the pretraining \ac{LM} experiments, and \ref{appendix:downstream_benchmark}
for the downstream benchmarks experiments.
In all our experiments, we used the \ac{HF} transformers library \cite{wolf_transformers_2020} available under an Apache 2.0 license.

\subsection{HNet\label{hnetmodel}}
\begin{figure}[htbp]
    \centering
    \resizebox{\columnwidth}{!}{
        \includegraphics{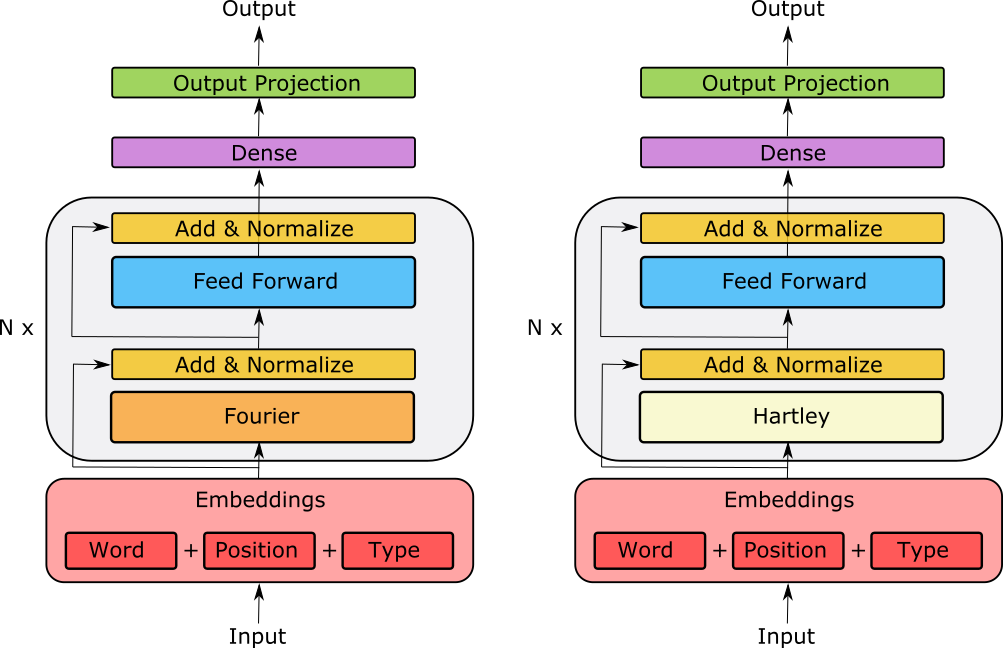}
    }
    \vspace{-7mm}
    \caption{On the left, \ac{FNet} encoder architecture with N encoder blocks. On the right, HNet encoder architecture with N encoder blocks.}
    \label{fig:HnetandFnetdiagram}
    \vspace{-5mm}
\end{figure}

\ac{HNet}, as \ac{FNet}, is an attention-free Transformer architecture, wherein each layer consists of a Hartley mixing sub-layer followed by a feed-forward sub-layer. The architecture is shown in Fig. \ref{fig:HnetandFnetdiagram}. As you can see, the Fourier sub-layer of each Transformer encoder layer is replaced with a Hartley sub-layer, which applies a 2D \ac{FFT} to its (sequence length, hidden dimension) embedding input – one 1D FFT along the sequence dimension, $F_{seq}$, and one 1D FFT along the hidden dimension, $F_h$. 
From the 2D \ac{FFT} thus produced, its real part is taken and subtracted from its imaginary part to obtain a 2D \ac{FHT}:
\begin{equation}
\label{eq:Fnetequation}
H = R (F_{seq} (F_{h}(x))) - I (F_{seq} (F_{h}(x))).
\end{equation}
Unlike in the \ac{FNet}, where only the real part (i.e. only the first term in Eq. \ref{eq:Fnetequation}), was taken into account, i.e. \ac{RFT}, so as not to deal with the complex numbers typical of the \ac{FT}; \ac{HT} already guarantees real-valued output including the imaginary part contributions, too. 
Since anything that can be done with \ac{FT} can be done with \ac{HT} and vice versa \cite{bracewell_computing_1995, paraskevas_hartley_2015}, 
\ac{HT} is a readily adoptable tool, when looking for a real-valued output.
We also implemented \ac{HNet} in the \ac{HF} library based on its \ac{FNet} implementation. 

In addition to the expression in Eq. \ref{eq:Fnetequation}, we tested several other real-valued configurations, but among the various experiments, \ac{HT} was the one that showed the most potential for success (see Appendix \ref{sec:fourier_architecture} for more details).

\subsection{Right-side Attention Encoder-Decoder\label{hnet-encdecmodel}}
As the \ac{FNet} paper proposed only an encoder architecture, \citet{kiruluta_new_2021} first suggested 3 different Seq2Seq architectures to overcome this shortcoming. 
In all of their proposed Seq2Seq diagrams, they removed the multi-headed self-attention layers in a transformer (both encoder and decoder): either by direct removal, i.e. moving the Fourier token mixing completely outside the
transformer by Fourier transforming both inputs, or by replacement with a Fourier layer. Although the latter solution, what they call Hybrid-FNet Seq2Seq, is their best solution, and is more similar to a classical transformer diagram, their results, as reported in Table \ref{tab:pubmed_results}, are a long way from those offered by alternatives such as PEGASUS and Longformer. 

\begin{figure}[htbp]
    \centering
    \resizebox{\columnwidth}{!}{
        \includegraphics{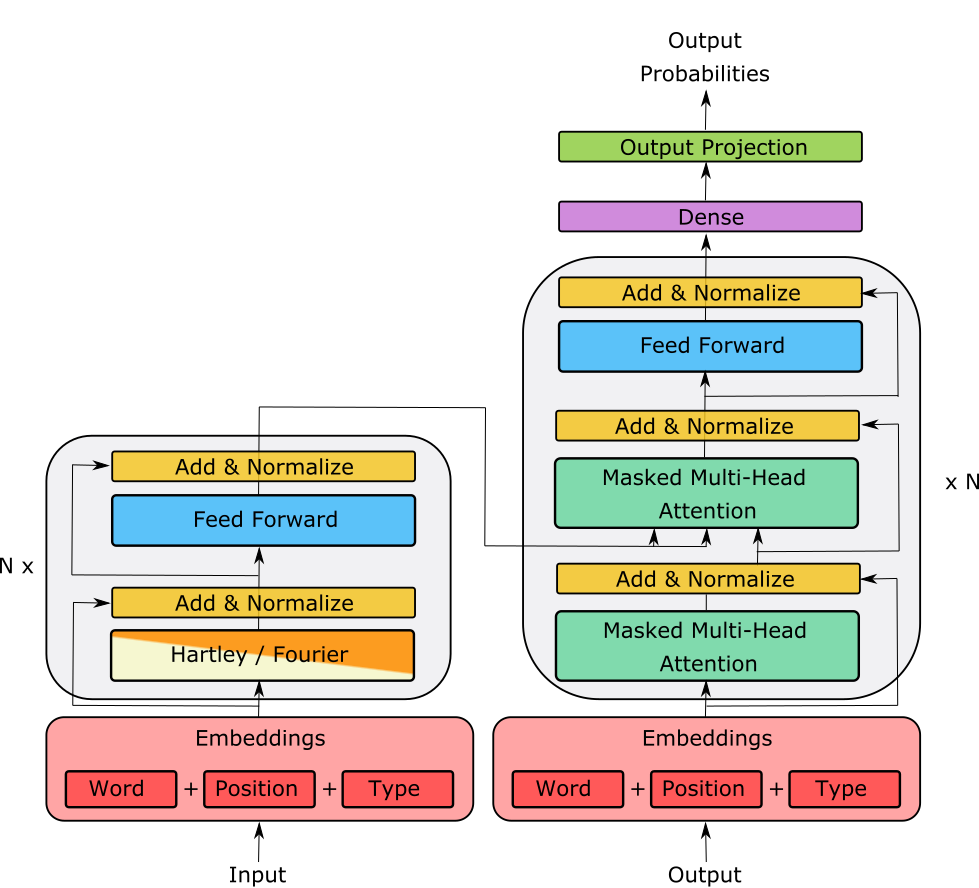}
    }
    \vspace{-7mm}
    \caption{Right-side Attention Encoder-Decoder (RAED) diagram. We tested also another configuration where we replaced these two Multi-head Attention green boxes with a Hartley/Fourier in the same manner as was done by \citet{kiruluta_new_2021}, the so-called FNet-Transformer, but without equal success as \ac{RAED} with downstream tasks.}
    \label{fig:halfatt_encdec}
    \vspace{-5mm}
\end{figure}

For these reasons, we have designed a new Seq2Seq infrastructure with an encoder without attention and a decoder with attention, which for clarity we can call the \ac{RAED} 
model (See Fig. \ref{fig:halfatt_encdec}).  It is based on the idea that an attention-based decoder gives a moderate computational load to the entire model infrastructure because it is generally powered by short text. Furthermore, in the main NLP tasks, where Seq2Seq infrastructure is involved, more emphasis is placed on the output generated by the decoder. We believe, therefore, that the multi-headed self-attention mechanism acting on the decoder tokens through a dense representation of the context via Fourier/Hartley descriptors (i.e. encoder output) is crucial to obtaining AI generated text which is indistinguishable from a human with a reduced computational cost.


\section{Results}
\label{sec:results}

In the following three sections, we present the results on the BillSum dataset, the PubMed dataset and the LexGLUE benchmark.




\subsection{BillSum}

Our results on the BillSum dataset are presented in Table \ref{tab:billsum_results}.

We observe that our RAED models
clearly exceed the baseline of the original article (DOC + SUM), even though their model is based on BERT-large which contains almost 2 times more encoder parameters and its pretraining cost is 10 times more expensive. 
Furthermore, our RAED-HNet-4096 base  model outperforms a transformer model, and competes with the PEGASUS-base model \cite{zhang_pegasus_2020} (35.14 vs. 37.78 Rouge-L), which is pretrained using the Gap-Sentences task specifically designed for abstractive summarization. Vice versa, it fails to reach the new \ac{SOTA}, i.e. BudgetLongformer \cite{joel_paper}. This has been trained with an ELECTRA task, which unlike the \ac{MLM} task turns out to be very efficient in summarization tasks. However, our models become very attractive solutions when cost and environmental impact are also taken into account, without compromising too much on quality. In particular, they almost halve training costs and reduce model operational costs by 17\%. 

We conclude that the \ac{RAED} architecture is effective, with minimal compute requirements for long-input summarization in-domain.

\begin{table*}[t]
\centering
\resizebox{0.9\textwidth}{!}{
    \begin{tabular}{lrrrr}
        \toprule
        Model (max-in-len->max-gen-len)                                          & \# Enc. Params $\downarrow$   & Rouge-1 $\uparrow$ & Rouge-L $\uparrow$ & Speed (T$\uparrow$ / I$\uparrow$) [it/s]\\ 
        \midrule
        DOC + SUM (BERT large)  \cite{kornilova_billsum_2019}
            & 340M  & 40.80 & 33.73 & OOM \\ 
        Transformer base  \cite{zhang_pegasus_2020}
            & 223M  & 44.05 & 30.98 & 0.80 (3.8x) / 0.21 \\ 
        PEGASUS base  \cite{zhang_pegasus_2020}
            & 223M  & 51.42 & 37.78 & 0.80 (3.8x) / 0.21 (1.0x) \\ 
        BudgetLongformer base (diverse)  \cite{joel_paper}
            & 255M & 55.45 & 43.23 & 0.85 (4.0x) / 0.22 (1.0x) \\  
        RAED-FNet-4096 base (4096->512)   
            & 182M & 47.64 & 34.51 & 1.40 (6.7x) / 0.26 (1.2x) \\  
        RAED-HNet-4096 base (4096->512)   
            & 182M & 45.92 & 35.14 & 1.40 (6.7x) / 0.26 (1.2x)\\  
        \bottomrule
    \end{tabular}
}
\caption{Results on the BillSum dataset. Enc. Params is short for Encoder Parameters. T and I are abbreviations for Train and Inference. We report the f1-scores of the Rouge metrics, and the speeds as number of iterations over a second, performed on a p3.2xlarge AWS instance with 1XV100 Nvidia GPU, batch size of 8, and an input sequence length of 1024 for the encoder and of 256 for the decoder (in parentheses, speed-ups compared to transformer architecture inference).}
\label{tab:billsum_results}
\end{table*}

\subsection{PubMed}

Our results on the PubMed dataset are presented in Table \ref{tab:pubmed_results}. 

Similar to the results on BillSum, our RAED models clearly outperform the Transformer-base model, in particular our RAED-HNet-4096 base (20.10 and vs. 19.02 Rouge-L). Furthermore, our RAED architecture exceeds the most performing Seq2Seq architecture, which includes \ac{FNet} layers, proposed by \citet{kiruluta_new_2021}.

As in BillSum, our LMs do not reach both PEGASUS and BudgetLongformer in terms of performance, but as described above they can be excellent alternatives when longer documents need to be processed and/or there are cost and speed constraints.

Note, that we pretrain, as BudgetLongformer, on a much narrower domain than PEGASUS (legal text vs. C4). For example, our tokenizer and model has never seen medical data during its pretraining phase. In addiction, our tokenizer has 3 times less tokens than the PEGASUS tokenizer (96k) and has half the number of tokens as of the BudgetLongformer tokenizer (64k).  

In conclusion, the RAED architecture is effective even on an out-of-domain downstream summarization task.
\begin{table}[t]
\centering
\resizebox{\columnwidth}{!}{
    \begin{tabular}{lrrrr}
        \toprule
        Model  (max-in-len->max-gen-len) & \# Enc. Params $\downarrow$   & Rouge-1 $\uparrow$  & Rouge-L $\uparrow$ \\ 
        \midrule
        Transformer base \cite{zhang_pegasus_2020}                                      & 223M & 33.94  & 19.02 \\ 
        PEGASUS base \cite{zhang_pegasus_2020}                                          & 223M & 39.98  & 25.23 \\ 
        BudgetLongformer base \cite{joel_paper}      
            & 255M & 41.16 & 26.53 \\ 
        Hybrid-FNet base \cite{kiruluta_new_2021}
            & 172M & 35.60 & 14.50 \\ 
        RAED-FNet-4096 base (4096->512) 
            & 182M & 37.24 & 19.98 \\ 
        RAED-FNet-8192 base (6144->512)
            & 185M & 37.32 & 20.17 \\ 
        RAED-HNet-4096 base (4096->512)
            & 182M & 37.52 & 20.10 \\ 
        \bottomrule
    \end{tabular}
}
\caption{Results on the PubMed dataset. Enc. Params is short for Encoder Parameters. We report the f1-scores of the Rouge metrics.
}
\label{tab:pubmed_results}
\end{table}


\subsection{LexGLUE}



Our results on the LexGLUE benchmark are presented in Table 
\ref{tab:lexglue_results} in Appendix \ref{sec:detailed_results}.

While, in Table \ref{tab:lexglue_perf_results}, we reported the performance of the various \ac{LM}s against a reference \ac{LM}, based on defined downstream metrics (e.g. $\mu$-f1-score):
\[
P=\left<\frac{metrics_{LM, task}}{metrics_{LM_{ref}, task}}\right>_{task}.
\]
This is similar to what was carried out by \citet{lee-thorp_fnet_2021} during the evaluation of the GLUE benchmark for \ac{FNet}. As we can see our models have a far better performance than \ac{FNet} one\footnote{checkpoint released by Google in HF, https://huggingface.co/google/fnet-base.}, which was not pre-trained on legal documents. 

\begin{table}[t]
\centering
\resizebox{\columnwidth}{!}{
    \begin{tabular}{lrrrr}
        \toprule
        Model    & Reference Model  & P ($\mu$-f1, m-f1) $\uparrow$\\ 
        \midrule
        FNet \cite{lee-thorp_fnet_2021}  &  
        BERT \cite{chalkidis_lexglue_2021} & 87.4, 82.4 \\ 
        FNet-4096   &        
        BERT \cite{chalkidis_lexglue_2021} & 91.5, 87.5 \\ 
        HNet-4096   & 
        BERT \cite{chalkidis_lexglue_2021} & 93.9, 88.9 \\ 
        [0.3cm]
 
        FNet-4096   & 
        RoBERTa \cite{chalkidis_lexglue_2021} & 91.6, 88.6 \\ 
        HNet-4096   & 
        RoBERTa \cite{chalkidis_lexglue_2021} & 93.8, 89.5 \\ 
        [0.3cm]
 
        FNet-4096   & 
        CaseLaw-BERT$^{*}$ \cite{chalkidis_lexglue_2021} & 89.2, 85.1 \\ 
        HNet-4096   & 
        CaseLaw-BERT$^{*}$ \cite{chalkidis_lexglue_2021} & 91.9, 86.7 \\ 
        [0.3cm]
 
        FNet-4096   & 
        Legal-BERT$^{*}$ \cite{chalkidis_lexglue_2021} & 88.2, 83.8 \\ 
        HNet-4096   & 
        Legal-BERT$^{*}$ \cite{chalkidis_lexglue_2021} & 91.4, 85.4 \\ 
        \bottomrule
    \end{tabular}
}
\caption{Results on the LexGLUE Benchamrk. We evaluated the Performance of our \ac{LM}s as \citet{lee-thorp_fnet_2021} did in the original paper with the GLUE benchmark. \ac{LM}s with asterix have seen these testing data during the \ac{MLM} pre-training stage.
}
\label{tab:lexglue_perf_results}
\end{table}

In particular we note that our legal \ac{FNet}, FNet-4096, achieves almost 92\% performance against BERT/RoBERTa models, similar to what the \ac{FNet} authors get for the GLUE benchmark.
While the legal model based on a Hartley layer, i.e. HNet-4096, achieves 94\% performance for same reference models, reducing the gap between Attention-based and Fourier-based models.

While the performance compared to models with legal domain knowledge is lower, such as CaseLaw-BERT and legal-BERT. In support of our models, the latter have a contaminated testing set since a part of those documents were seen during \ac{MLM} pretraining.

As already observed in \citet{joel_paper}, models trained only on long documents or with emphasis on long-context, hardly reach the \ac{SOTA} for these LexGLUE tasks. This, indeed, suggested that the LexGLUE benchmark needs a more accurate description of the first 512 tokens (e.g. 4 over 7 tasks have a median token length far below 512, as shown in Appendix \ref{sec:data_details}),  
which could be obtained through a pretraining dataset with a comparable distribution of token inputs.
We avoided focusing too much on this point since the purpose of the paper is to solve legal long documents as input.

Finally, we used the \ac{HP} provided by the benchmark, which we can merely speculate that they are strongly related to BERT-like models (i.e. attention mechanism-like). Most of the models, for which the LexGLUE benchmark was performed, come from a BERT-checkpoint.\footnote{
Note that Legal-Bert, CaseLaw-Bert, Longformer and BigBird have been warm started from the BERT or RoBERTa checkpoint. Thus, they have been trained on short documents extensively, at least, during the first pretraining phase.
}
It is not possible to understand how these \ac{HP}s were chosen by \citet{chalkidis_lexglue_2021}, but it is known how they can strongly influence the downstream performance \cite{liu_empirical_2021, dodge_fine-tuning_2020}.

\section{Conclusions and Future Work}
\label{sec:conclusions_future_work}

\subsection{Answers to Main Research Questions}

\noindent \textbf{RQ1}: \emph{Is it possible to create a domain specific (e.g. legal) \ac{FNet} based \ac{LM} from scratch, reducing costs and carbon footprint compared with self-attention based \ac{LM}s?}
Yes, it is possible to pretrain a domain-specific (legal) \ac{FNet} based \ac{LM} from scratch with lesser compute and memory, achieving comparable performance with self-attention based \ac{LM}s.

\noindent \textbf{RQ2}: \emph{Does using the Hartley transform instead of Fourier in the \ac{FNet} based architecture to create a domain specific (e.g. legal) \ac{LM} from scratch lead to better results?}
Yes. As can be seen from the result Tables \ref{tab:billsum_results}, \ref{tab:pubmed_results}, \ref{tab:lexglue_perf_results} and \ref{tab:lexglue_results}, for similar maximum sequence length models, Hartley transform based models show better results.

\noindent \textbf{RQ3}: \emph{Is it possible to create a ``competitive'' Seq2Seq model, which can approach the performance of \ac{SOTA} ones, using Fourier/Hartley layers?}
Yes. As can be seen from Table \ref{tab:pubmed_results}, the proposed \ac{RAED} architecture achieves comparable performance with the \ac{SOTA} ones.

\noindent \textbf{RQ4}: \emph{How do our models compare with other models on the challenging summarization task? Particularly in the case of a legal domain-specific benchmark such as BillSum?}
As can be seen from Table \ref{tab:billsum_results}, although our \ac{LM}s do not beat the \ac{SOTA}, they achieve comparable metrics with lesser compute, memory and data requirements.

\noindent \textbf{RQ5}: \emph{How well do our models generalize to other domains, for example in the biomedical domain, as evaluated by the PubMed summarization benchmark?}
As can be seen from Table \ref{tab:pubmed_results}, although our models do not beat the \ac{SOTA}, it achieves comparable metrics with lesser compute, memory and data requirements.

\noindent \textbf{RQ6}: \emph{How do our \acp{LM} compare with other models on the \acf{TC} benchmark LexGLUE?}
As can be seen from Tables \ref{tab:lexglue_perf_results} and \ref{tab:lexglue_results}, our \ac{LM}s reduce the expected performance gap for legal tasks. 

\subsection{Conclusion}


In this work, we investigate the Fourier-class Transform as a replacement to the self-attention mechanism to solve the long-document problem in the legal domain. We trained from scratch and successfully evaluated three new legal \ac{LM}s, which can conveniently deal long documents. We found also that \ac{HT} leads to better outcomes than \ac{FT}, without additional computational cost.
In addition, we also proposed a new architecture for a Seq2Seq model involving no-attention based \ac{LM}s as encoders, going well beyond previous Fourier-based architecture outcomes, and achieving values close to current \ac{SOTA} with about half the training costs and 17\% less operational costs (i.e. inference). 

\subsection{Future Work}
A potential future experiment could explore a wider \ac{HP}s optimization, in particular during the BillSum and PubMed tasks, but also in \ac{MLM} pre-training phase. We, indeed, limit only to a narrow exploration of different learning rates and batch sizes.  For example, the tokenizer's vocabulary is another fairly incisive \ac{HP} that could be tuned (e.g. by increasing it), to accommodate a more vocabulary-rich language (as that of legal).
Probably also extending the pretraining dataset with not-legal documents could lead to more robust and generic \ac{LM}s, which might achieve further gains for domain and out-domain downstream tasks. 

In addition, future works could also check how the \ac{LM}s perform for other in-of-domain tasks such as CUAD \cite{hendrycks_cuad_2021}, Posture50K \cite{song_multi-label_2022} for \ac{TC} or the recently released MultiLexSum \cite{shen_multi-lexsum_2022} for summarization; and for other out-of-domain tasks such as by trying one of the remaining 11 summarization datasets evaluated by  \citet{zhang_pegasus_2020}.



Finally, another future work with a potential high return of investment, which we could not perform due to resource allocation constraints, is to replace the Fourier-class Transform with a \ac{LT}. \ac{LT} satisfies the causality principle\footnote{it could be relevant in Casual Language Modelling}, and provides a solution in terms of decaying exponents multiplied by a periodic part (i.e. \ac{FT}). For these reasons, it could better mimic short \& long correlations, and inherently assign a clockwise direction to the text sequence.


\section{Limitations}
\label{sec:limitations}

Fourier-class LMs have the disadvantage of not achieving the same performance as their respective cousins based on attention mechanisms, in \ac{TC} tasks. In particular, when a good description of the first N tokens is most important. For these reasons, they cannot be considered a solution if the quality of the result is prioritized.

Because of insufficient compute, 
we were not able to create the \emph{large} versions of these 3 models. Therefore, we do not know how our models or the \ac{RAED} architecture would scale in the case of \emph{large} models. We imagine that it will go to produce better results but it is not possible for us to estimate whether they are meaningfully better.

Summarization tasks were evaluated only on the basis of the Rouge score, since it was the only data available for comparison with previous works. In the case of a proper evaluation of the \ac{RAED} architecture, it would be necessary to conduct an evaluation through human experts, since it is known that the Rouge score can lead to misleading interpolations of the results. In particular, when the output of the model is a reformulation of the same concept. This unfortunately requires human involvement in the loop, which was not within the costs allocated for this research.

Out of the 9.2 GB of data used during the \ac{MLM} pretraining, only 1.7 GB was in U.K. English with a mix of European and U.K. legal texts, while the rest was based on U.S. English with U.S. legal texts.   
Apart from the differences between U.S. and U.K. languages, from a legal perspective, the U.K. and U.S. systems are based on case law, while the European system is based on civil law. Therefore, we expect the models to be able to transfer knowledge in legal downstream tasks where U.S. English and the case-law system is present, while it is more difficult for texts in U.K. English and with the civil law system.

\section*{Ethics Statement}

Pretraining language models is a very compute-heavy process and thus leaves a large carbon footprint \cite{strubell_energy_2019, patterson_carbon_2021}. Our research has the main aim to reduce such computational resource requirements without compromising too much on the output quality. 


\bibliography{bib/anthology,bib/custom,bib/references, bib/hnet}

\begin{thebibliography}{62}
\expandafter\ifx\csname natexlab\endcsname\relax\def\natexlab#1{#1}\fi

\bibitem[{Armitage et~al.(2020)Armitage, Kacupaj, Tahmasebzadeh, Swati,
  Maleshkova, Ewerth, and Lehmann}]{armitage_mlm_2020}
Jason Armitage, Endri Kacupaj, Golsa Tahmasebzadeh, Swati, Maria Maleshkova,
  Ralph Ewerth, and Jens Lehmann. 2020.
\newblock \href {https://doi.org/10.1145/3340531.3412783} {{MLM}: {A}
  {Benchmark} {Dataset} for {Multitask} {Learning} with {Multiple} {Languages}
  and {Modalities}}.
\newblock In \emph{Proceedings of the 29th {ACM} {International} {Conference}
  on {Information} \& {Knowledge} {Management}}, pages 2967--2974.
\newblock ArXiv:2008.06376 [cs, stat].

\bibitem[{Beltagy et~al.(2019)Beltagy, Lo, and Cohan}]{beltagy_scibert_2019}
Iz~Beltagy, Kyle Lo, and Arman Cohan. 2019.
\newblock \href {http://arxiv.org/abs/1903.10676} {{SciBERT}: {A} {Pretrained}
  {Language} {Model} for {Scientific} {Text}}.
\newblock \emph{arXiv:1903.10676 [cs]}.
\newblock ArXiv: 1903.10676.

\bibitem[{Beltagy et~al.(2020{\natexlab{a}})Beltagy, Peters, and
  Cohan}]{beltagy2020longformer}
Iz~Beltagy, Matthew~E. Peters, and Arman Cohan. 2020{\natexlab{a}}.
\newblock \href {http://arxiv.org/abs/2004.05150} {Longformer: The
  long-document transformer}.

\bibitem[{Beltagy et~al.(2020{\natexlab{b}})Beltagy, Peters, and
  Cohan}]{beltagy_longformer_2020}
Iz~Beltagy, Matthew~E. Peters, and Arman Cohan. 2020{\natexlab{b}}.
\newblock \href {http://arxiv.org/abs/2004.05150} {Longformer: {The}
  {Long}-{Document} {Transformer}}.
\newblock \emph{arXiv:2004.05150 [cs]}.
\newblock ArXiv: 2004.05150.

\bibitem[{Bracewell(1995)}]{bracewell_computing_1995}
Ronald~N. Bracewell. 1995.
\newblock \href {https://doi.org/10.1063/1.168534} {Computing with the
  {Hartley} transform}.
\newblock \emph{Comput. Phys.}, 9(4):373.

\bibitem[{Brown et~al.(2020)Brown, Mann, Ryder, Subbiah, Kaplan, Dhariwal,
  Neelakantan, Shyam, Sastry, Askell, Agarwal, Herbert-Voss, Krueger, Henighan,
  Child, Ramesh, Ziegler, Wu, Winter, Hesse, Chen, Sigler, Litwin, Gray, Chess,
  Clark, Berner, McCandlish, Radford, Sutskever, and
  Amodei}]{brown_language_2020}
Tom~B. Brown, Benjamin Mann, Nick Ryder, Melanie Subbiah, Jared Kaplan,
  Prafulla Dhariwal, Arvind Neelakantan, Pranav Shyam, Girish Sastry, Amanda
  Askell, Sandhini Agarwal, Ariel Herbert-Voss, Gretchen Krueger, Tom Henighan,
  Rewon Child, Aditya Ramesh, Daniel~M. Ziegler, Jeffrey Wu, Clemens Winter,
  Christopher Hesse, Mark Chen, Eric Sigler, Mateusz Litwin, Scott Gray,
  Benjamin Chess, Jack Clark, Christopher Berner, Sam McCandlish, Alec Radford,
  Ilya Sutskever, and Dario Amodei. 2020.
\newblock \href {http://arxiv.org/abs/2005.14165} {Language {Models} are
  {Few}-{Shot} {Learners}}.
\newblock \emph{arXiv:2005.14165 [cs]}.
\newblock ArXiv: 2005.14165.

\bibitem[{Chalkidis et~al.(2020)Chalkidis, Fergadiotis, Malakasiotis, Aletras,
  and Androutsopoulos}]{chalkidis_legal-bert_2020}
Ilias Chalkidis, Manos Fergadiotis, Prodromos Malakasiotis, Nikolaos Aletras,
  and Ion Androutsopoulos. 2020.
\newblock \href {http://arxiv.org/abs/2010.02559} {{LEGAL}-{BERT}: {The}
  {Muppets} straight out of {Law} {School}}.
\newblock \emph{arXiv:2010.02559 [cs]}.
\newblock ArXiv: 2010.02559.

\bibitem[{Chalkidis et~al.(2021)Chalkidis, Jana, Hartung, Bommarito,
  Androutsopoulos, Katz, and Aletras}]{chalkidis_lexglue_2021}
Ilias Chalkidis, Abhik Jana, Dirk Hartung, Michael~James Bommarito, Ion
  Androutsopoulos, Daniel~Martin Katz, and Nikolaos Aletras. 2021.
\newblock \href {https://doi.org/10.2139/ssrn.3936759} {{LexGLUE}: {A}
  {Benchmark} {Dataset} for {Legal} {Language} {Understanding} in {English}}.
\newblock {SSRN} {Scholarly} {Paper} ID 3936759, Social Science Research
  Network, Rochester, NY.

\bibitem[{{Chalkidis et al.
  (2019)}(2019)}]{chalkidis_et_al_2019_eurlex57k_nodate}
{Chalkidis et al. (2019)}. 2019.
\newblock \href {http://archive.org/details/EURLEX57K} {\emph{{EURLEX57K}}}.
\newblock Association for Computational Linguistics.

\bibitem[{Chen et~al.(2019)Chen, Ma, Mao, and Li}]{chen_multi-task_2019}
Yangbin Chen, Yun Ma, Xudong Mao, and Qing Li. 2019.
\newblock \href {https://doi.org/10.1007/s41019-019-0087-7} {Multi-{Task}
  {Learning} for {Abstractive} and {Extractive} {Summarization}}.
\newblock \emph{Data Sci. Eng.}, 4(1):14--23.

\bibitem[{Child et~al.(2019{\natexlab{a}})Child, Gray, Radford, and
  Sutskever}]{child2019generating}
Rewon Child, Scott Gray, Alec Radford, and Ilya Sutskever. 2019{\natexlab{a}}.
\newblock \href {http://arxiv.org/abs/1904.10509} {Generating long sequences
  with sparse transformers}.

\bibitem[{Child et~al.(2019{\natexlab{b}})Child, Gray, Radford, and
  Sutskever}]{child_generating_2019}
Rewon Child, Scott Gray, Alec Radford, and Ilya Sutskever. 2019{\natexlab{b}}.
\newblock \href {http://arxiv.org/abs/1904.10509} {Generating {Long}
  {Sequences} with {Sparse} {Transformers}}.
\newblock \emph{arXiv:1904.10509 [cs, stat]}.
\newblock ArXiv: 1904.10509.

\bibitem[{Choromanski et~al.(2020)Choromanski, Likhosherstov, Dohan, Song,
  Gane, Sarlos, Hawkins, Davis, Belanger, Colwell, and
  Weller}]{choromanski2020masked}
Krzysztof Choromanski, Valerii Likhosherstov, David Dohan, Xingyou Song,
  Andreea Gane, Tamas Sarlos, Peter Hawkins, Jared Davis, David Belanger, Lucy
  Colwell, and Adrian Weller. 2020.
\newblock \href {http://arxiv.org/abs/2006.03555} {Masked language modeling for
  proteins via linearly scalable long-context transformers}.

\bibitem[{Choromanski et~al.(2021)Choromanski, Likhosherstov, Dohan, Song,
  Gane, Sarlos, Hawkins, Davis, Mohiuddin, Kaiser, Belanger, Colwell, and
  Weller}]{choromanski2021rethinking}
Krzysztof Choromanski, Valerii Likhosherstov, David Dohan, Xingyou Song,
  Andreea Gane, Tamas Sarlos, Peter Hawkins, Jared Davis, Afroz Mohiuddin,
  Lukasz Kaiser, David Belanger, Lucy Colwell, and Adrian Weller. 2021.
\newblock \href {http://arxiv.org/abs/2009.14794} {Rethinking attention with
  performers}.

\bibitem[{Clark et~al.(2019)Clark, Khandelwal, Levy, and
  Manning}]{clark-etal-2019-bert}
Kevin Clark, Urvashi Khandelwal, Omer Levy, and Christopher~D. Manning. 2019.
\newblock \href {https://doi.org/10.18653/v1/W19-4828} {What does {BERT} look
  at? an analysis of {BERT}{'}s attention}.
\newblock In \emph{Proceedings of the 2019 ACL Workshop BlackboxNLP: Analyzing
  and Interpreting Neural Networks for NLP}, pages 276--286, Florence, Italy.
  Association for Computational Linguistics.

\bibitem[{Cohan et~al.(2018)Cohan, Dernoncourt, Kim, Bui, Kim, Chang, and
  Goharian}]{cohan_discourse-aware_2018}
Arman Cohan, Franck Dernoncourt, Doo~Soon Kim, Trung Bui, Seokhwan Kim, Walter
  Chang, and Nazli Goharian. 2018.
\newblock \href {https://doi.org/10.18653/v1/N18-2097} {A {Discourse}-{Aware}
  {Attention} {Model} for {Abstractive} {Summarization} of {Long} {Documents}}.
\newblock In \emph{Proceedings of the 2018 {Conference} of the {North}
  {American} {Chapter} of the {Association} for {Computational} {Linguistics}:
  {Human} {Language} {Technologies}, {Volume} 2 ({Short} {Papers})}, pages
  615--621, New Orleans, Louisiana. Association for Computational Linguistics.

\bibitem[{Dodge et~al.(2020)Dodge, Ilharco, Schwartz, Farhadi, Hajishirzi, and
  Smith}]{dodge_fine-tuning_2020}
Jesse Dodge, Gabriel Ilharco, Roy Schwartz, Ali Farhadi, Hannaneh Hajishirzi,
  and Noah Smith. 2020.
\newblock \href {http://arxiv.org/abs/2002.06305} {Fine-{Tuning} {Pretrained}
  {Language} {Models}: {Weight} {Initializations}, {Data} {Orders}, and {Early}
  {Stopping}}.
\newblock ArXiv:2002.06305 [cs].

\bibitem[{Dosovitskiy et~al.(2021)Dosovitskiy, Beyer, Kolesnikov, Weissenborn,
  Zhai, Unterthiner, Dehghani, Minderer, Heigold, Gelly, Uszkoreit, and
  Houlsby}]{dosovitskiy2021image}
Alexey Dosovitskiy, Lucas Beyer, Alexander Kolesnikov, Dirk Weissenborn,
  Xiaohua Zhai, Thomas Unterthiner, Mostafa Dehghani, Matthias Minderer, Georg
  Heigold, Sylvain Gelly, Jakob Uszkoreit, and Neil Houlsby. 2021.
\newblock \href {http://arxiv.org/abs/2010.11929} {An image is worth 16x16
  words: Transformers for image recognition at scale}.

\bibitem[{Fusco et~al.(2022)Fusco, Pascual, and Staar}]{fusco_pnlp-mixer_2022}
Francesco Fusco, Damian Pascual, and Peter Staar. 2022.
\newblock \href {https://doi.org/10.48550/arXiv.2202.04350} {{pNLP}-{Mixer}: an
  {Efficient} all-{MLP} {Architecture} for {Language}}.
\newblock ArXiv:2202.04350 [cs].

\bibitem[{Gu et~al.(2021)Gu, Tinn, Cheng, Lucas, Usuyama, Liu, Naumann, Gao,
  and Poon}]{domain_specific_pretraining_biomedical_nlp}
Yu~Gu, Robert Tinn, Hao Cheng, Michael Lucas, Naoto Usuyama, Xiaodong Liu,
  Tristan Naumann, Jianfeng Gao, and Hoifung Poon. 2021.
\newblock \href {https://doi.org/10.1145/3458754} {Domain-specific language
  model pretraining for biomedical natural language processing}.
\newblock \emph{ACM Trans. Comput. Healthcare}, 3(1).

\bibitem[{Hao et~al.(2020)Hao, Zhang, , 9156, , Zhang, Jianwei, Ma, , 9157, ,
  and Ma}]{hartley_spectral_pooling_cnns}
Hao, Zhang, , 9156, , Hao Zhang, Jianwei, Ma, , 9157, , and Jianwei Ma. 2020.
\newblock \href {https://doi.org/https://doi.org/10.4208/csiam-am.2020-0018}
  {Hartley spectral pooling for deep learning}.
\newblock \emph{CSIAM Transactions on Applied Mathematics}, 1(3):518--529.

\bibitem[{Hendrycks et~al.(2021)Hendrycks, Burns, Chen, and
  Ball}]{hendrycks_cuad_2021}
Dan Hendrycks, Collin Burns, Anya Chen, and Spencer Ball. 2021.
\newblock \href {http://arxiv.org/abs/2103.06268} {{CUAD}: {An}
  {Expert}-{Annotated} {NLP} {Dataset} for {Legal} {Contract} {Review}}.
\newblock ArXiv:2103.06268 [cs].

\bibitem[{Jiao et~al.(2020)Jiao, Yin, Shang, Jiang, Chen, Li, Wang, and
  Liu}]{jiao2020tinybert}
Xiaoqi Jiao, Yichun Yin, Lifeng Shang, Xin Jiang, Xiao Chen, Linlin Li, Fang
  Wang, and Qun Liu. 2020.
\newblock \href {http://arxiv.org/abs/1909.10351} {Tinybert: Distilling bert
  for natural language understanding}.

\bibitem[{Katharopoulos et~al.(2020)Katharopoulos, Vyas, Pappas, and
  Fleuret}]{pmlr-v119-katharopoulos20a}
Angelos Katharopoulos, Apoorv Vyas, Nikolaos Pappas, and Fran{\c{c}}ois
  Fleuret. 2020.
\newblock \href {https://proceedings.mlr.press/v119/katharopoulos20a.html}
  {Transformers are {RNN}s: Fast autoregressive transformers with linear
  attention}.
\newblock In \emph{Proceedings of the 37th International Conference on Machine
  Learning}, volume 119 of \emph{Proceedings of Machine Learning Research},
  pages 5156--5165. PMLR.

\bibitem[{Kim and Awadalla(2020)}]{kim2020fastformers}
Young~Jin Kim and Hany~Hassan Awadalla. 2020.
\newblock \href {http://arxiv.org/abs/2010.13382} {Fastformers: Highly
  efficient transformer models for natural language understanding}.

\bibitem[{Kiruluta et~al.(2021)Kiruluta, Lemos, and Lundy}]{kiruluta_new_2021}
Andrew Kiruluta, Andreas Lemos, and Eric Lundy. 2021.
\newblock \href {http://arxiv.org/abs/2111.15473} {New {Approaches} to {Long}
  {Document} {Summarization}: {Fourier} {Transform} {Based} {Attention} in a
  {Transformer} {Model}}.
\newblock ArXiv:2111.15473 [cs].

\bibitem[{Kitaev et~al.(2020{\natexlab{a}})Kitaev, Kaiser, and
  Levskaya}]{kitaev_reformer_2020}
Nikita Kitaev, Łukasz Kaiser, and Anselm Levskaya. 2020{\natexlab{a}}.
\newblock \href {http://arxiv.org/abs/2001.04451} {Reformer: {The} {Efficient}
  {Transformer}}.
\newblock \emph{arXiv:2001.04451 [cs, stat]}.
\newblock ArXiv: 2001.04451.

\bibitem[{Kitaev et~al.(2020{\natexlab{b}})Kitaev, Łukasz Kaiser, and
  Levskaya}]{kitaev2020reformer}
Nikita Kitaev, Łukasz Kaiser, and Anselm Levskaya. 2020{\natexlab{b}}.
\newblock \href {http://arxiv.org/abs/2001.04451} {Reformer: The efficient
  transformer}.

\bibitem[{Kornilova and Eidelman(2019)}]{kornilova_billsum_2019}
Anastassia Kornilova and Vladimir Eidelman. 2019.
\newblock \href {https://doi.org/10.18653/v1/D19-5406} {{BillSum}: {A} {Corpus}
  for {Automatic} {Summarization} of {US} {Legislation}}.
\newblock In \emph{Proceedings of the 2nd {Workshop} on {New} {Frontiers} in
  {Summarization}}, pages 48--56, Hong Kong, China. Association for
  Computational Linguistics.

\bibitem[{Kudo(2018)}]{kudo_subword_2018}
Taku Kudo. 2018.
\newblock \href {http://arxiv.org/abs/1804.10959} {Subword {Regularization}:
  {Improving} {Neural} {Network} {Translation} {Models} with {Multiple}
  {Subword} {Candidates}}.
\newblock ArXiv:1804.10959 [cs].

\bibitem[{Lan et~al.(2020)Lan, Chen, Goodman, Gimpel, Sharma, and
  Soricut}]{lan_albert_2020}
Zhenzhong Lan, Mingda Chen, Sebastian Goodman, Kevin Gimpel, Piyush Sharma, and
  Radu Soricut. 2020.
\newblock \href {http://arxiv.org/abs/1909.11942} {{ALBERT}: {A} {Lite} {BERT}
  for {Self}-supervised {Learning} of {Language} {Representations}}.
\newblock \emph{arXiv:1909.11942 [cs]}.
\newblock ArXiv: 1909.11942.

\bibitem[{Lee et~al.(2019)Lee, Yoon, Kim, Kim, Kim, So, and
  Kang}]{lee_biobert_2019}
Jinhyuk Lee, Wonjin Yoon, Sungdong Kim, Donghyeon Kim, Sunkyu Kim, Chan~Ho So,
  and Jaewoo Kang. 2019.
\newblock \href {https://doi.org/10.1093/bioinformatics/btz682} {{BioBERT}: a
  pre-trained biomedical language representation model for biomedical text
  mining}.
\newblock \emph{Bioinformatics}, page btz682.
\newblock ArXiv: 1901.08746.

\bibitem[{Lee-Thorp and Ainslie(2022)}]{lee-thorp_sparse_2022}
James Lee-Thorp and Joshua Ainslie. 2022.
\newblock \href {http://arxiv.org/abs/2205.12399} {Sparse {Mixers}: {Combining}
  {MoE} and {Mixing} to build a more efficient {BERT}}.
\newblock ArXiv:2205.12399 [cs].

\bibitem[{Lee-Thorp et~al.(2021)Lee-Thorp, Ainslie, Eckstein, and
  Ontanon}]{lee-thorp_fnet_2021}
James Lee-Thorp, Joshua Ainslie, Ilya Eckstein, and Santiago Ontanon. 2021.
\newblock \href {http://arxiv.org/abs/2105.03824} {{FNet}: {Mixing} {Tokens}
  with {Fourier} {Transforms}}.
\newblock \emph{arXiv:2105.03824 [cs]}.
\newblock ArXiv: 2105.03824.

\bibitem[{Lewis et~al.(2020)Lewis, Liu, Goyal, Ghazvininejad, Mohamed, Levy,
  Stoyanov, and Zettlemoyer}]{lewis_bart_2020}
Mike Lewis, Yinhan Liu, Naman Goyal, Marjan Ghazvininejad, Abdelrahman Mohamed,
  Omer Levy, Veselin Stoyanov, and Luke Zettlemoyer. 2020.
\newblock \href {https://doi.org/10.18653/v1/2020.acl-main.703} {{BART}:
  {Denoising} {Sequence}-to-{Sequence} {Pre}-training for {Natural} {Language}
  {Generation}, {Translation}, and {Comprehension}}.
\newblock In \emph{Proceedings of the 58th {Annual} {Meeting} of the
  {Association} for {Computational} {Linguistics}}, pages 7871--7880, Online.
  Association for Computational Linguistics.

\bibitem[{Li et~al.(2022)Li, Wehbe, Ahmad, Wang, and
  Luo}]{li_clinical-longformer_2022}
Yikuan Li, Ramsey~M. Wehbe, Faraz~S. Ahmad, Hanyin Wang, and Yuan Luo. 2022.
\newblock \href {http://arxiv.org/abs/2201.11838} {Clinical-{Longformer} and
  {Clinical}-{BigBird}: {Transformers} for long clinical sequences}.
\newblock \emph{arXiv:2201.11838 [cs]}.
\newblock ArXiv: 2201.11838.

\bibitem[{Liu and Wang(2021)}]{liu_empirical_2021}
Xueqing Liu and Chi Wang. 2021.
\newblock \href {https://doi.org/10.18653/v1/2021.acl-long.178} {An {Empirical}
  {Study} on {Hyperparameter} {Optimization} for {Fine}-{Tuning} {Pre}-trained
  {Language} {Models}}.
\newblock In \emph{Proceedings of the 59th {Annual} {Meeting} of the
  {Association} for {Computational} {Linguistics} and the 11th {International}
  {Joint} {Conference} on {Natural} {Language} {Processing} ({Volume} 1: {Long}
  {Papers})}, pages 2286--2300, Online. Association for Computational
  Linguistics.

\bibitem[{Liu et~al.(2019)Liu, Ott, Goyal, Du, Joshi, Chen, Levy, Lewis,
  Zettlemoyer, and Stoyanov}]{liu_roberta_2019}
Yinhan Liu, Myle Ott, Naman Goyal, Jingfei Du, Mandar Joshi, Danqi Chen, Omer
  Levy, Mike Lewis, Luke Zettlemoyer, and Veselin Stoyanov. 2019.
\newblock \href {http://arxiv.org/abs/1907.11692} {{RoBERTa}: {A} {Robustly}
  {Optimized} {BERT} {Pretraining} {Approach}}.
\newblock \emph{arXiv:1907.11692 [cs]}.
\newblock ArXiv: 1907.11692.

\bibitem[{Mozafari et~al.(2021)Mozafari, Clark, Gross, and
  Meyer}]{hartley_stochastic_cnns}
S.~H. Mozafari, J.~J. Clark, W.~J. Gross, and B.~H. Meyer. 2021.
\newblock \href {https://doi.org/10.1109/SiPS52927.2021.00049} {Hartley
  stochastic computing for convolutional neural networks}.
\newblock In \emph{2021 IEEE Workshop on Signal Processing Systems (SiPS)},
  pages 1--6.

\bibitem[{Nikolaus and Giofr\'e(2022)}]{joel_paper}
Joel Nikolaus and Daniele Giofr\'e. 2022.
\newblock \href {https://neurips2022-enlsp.github.io/accepted_papers.html}
  {{BudgetLongformer}: {Can} we {Cheaply} {Pretrain} a {SotA} {Legal}
  {Language} {Model} {From} {Scratch}?}

\bibitem[{Paraskevas et~al.(2015)Paraskevas, Barbarosou, and
  Chilton}]{paraskevas_hartley_2015}
Ioannis Paraskevas, Maria Barbarosou, and Edward Chilton. 2015.
\newblock \href {https://doi.org/10.1049/joe.2014.0350} {Hartley transform and
  the use of the {Whitened} {Hartley} spectrum as a tool for phase spectral
  processing}.
\newblock \emph{J. eng.}, 2015(3):95--101.

\bibitem[{Patterson et~al.(2021)Patterson, Gonzalez, Le, Liang, Munguia,
  Rothchild, So, Texier, and Dean}]{patterson_carbon_2021}
David Patterson, Joseph Gonzalez, Quoc Le, Chen Liang, Lluis-Miquel Munguia,
  Daniel Rothchild, David So, Maud Texier, and Jeff Dean. 2021.
\newblock \href {http://arxiv.org/abs/2104.10350} {Carbon {Emissions} and
  {Large} {Neural} {Network} {Training}}.
\newblock \emph{arXiv:2104.10350 [cs]}.
\newblock ArXiv: 2104.10350.

\bibitem[{Raffel et~al.(2020)Raffel, Shazeer, Roberts, Lee, Narang, Matena,
  Zhou, Li, and Liu}]{raffel_exploring_2020}
Colin Raffel, Noam Shazeer, Adam Roberts, Katherine Lee, Sharan Narang, Michael
  Matena, Yanqi Zhou, Wei Li, and Peter~J. Liu. 2020.
\newblock \href {http://arxiv.org/abs/1910.10683} {Exploring the {Limits} of
  {Transfer} {Learning} with a {Unified} {Text}-to-{Text} {Transformer}}.
\newblock \emph{arXiv:1910.10683 [cs, stat]}.
\newblock ArXiv: 1910.10683.

\bibitem[{Roy et~al.(2021)Roy, Saffar, Vaswani, and
  Grangier}]{roy_efficient_2021}
Aurko Roy, Mohammad Saffar, Ashish Vaswani, and David Grangier. 2021.
\newblock \href {https://doi.org/10.1162/tacl_a_00353} {Efficient
  {Content}-{Based} {Sparse} {Attention} with {Routing} {Transformers}}.
\newblock \emph{Transactions of the Association for Computational Linguistics},
  9:53--68.
\newblock Place: Cambridge, MA Publisher: MIT Press.

\bibitem[{Shen et~al.(2022)Shen, Lo, Yu, Dahlberg, Schlanger, and
  Downey}]{shen_multi-lexsum_2022}
Zejiang Shen, Kyle Lo, Lauren Yu, Nathan Dahlberg, Margo Schlanger, and Doug
  Downey. 2022.
\newblock \href {https://doi.org/10.48550/arXiv.2206.10883} {Multi-{LexSum}:
  {Real}-{World} {Summaries} of {Civil} {Rights} {Lawsuits} at {Multiple}
  {Granularities}}.
\newblock ArXiv:2206.10883 [cs].

\bibitem[{Song et~al.(2022)Song, Vold, Madan, and
  Schilder}]{song_multi-label_2022}
Dezhao Song, Andrew Vold, Kanika Madan, and Frank Schilder. 2022.
\newblock \href {https://doi.org/10.1016/j.is.2021.101718} {Multi-label legal
  document classification: {A} deep learning-based approach with
  label-attention and domain-specific pre-training}.
\newblock \emph{Information Systems}, 106:101718.

\bibitem[{Strubell et~al.(2019)Strubell, Ganesh, and
  McCallum}]{strubell_energy_2019}
Emma Strubell, Ananya Ganesh, and Andrew McCallum. 2019.
\newblock \href {https://doi.org/10.48550/arXiv.1906.02243} {Energy and
  {Policy} {Considerations} for {Deep} {Learning} in {NLP}}.
\newblock ArXiv:1906.02243 [cs].

\bibitem[{Tay et~al.(2021)Tay, Bahri, Metzler, Juan, Zhao, and
  Zheng}]{tay_synthesizer_2021}
Yi~Tay, Dara Bahri, Donald Metzler, Da-Cheng Juan, Zhe Zhao, and Che Zheng.
  2021.
\newblock \href {http://arxiv.org/abs/2005.00743} {Synthesizer: {Rethinking}
  {Self}-{Attention} in {Transformer} {Models}}.
\newblock \emph{arXiv:2005.00743 [cs]}.
\newblock ArXiv: 2005.00743.

\bibitem[{Tay et~al.(2020)Tay, Dehghani, Bahri, and
  Metzler}]{tay_efficient_2020}
Yi~Tay, Mostafa Dehghani, Dara Bahri, and Donald Metzler. 2020.
\newblock \href {http://arxiv.org/abs/2009.06732} {Efficient {Transformers}:
  {A} {Survey}}.
\newblock \emph{arXiv:2009.06732 [cs]}.
\newblock ArXiv: 2009.06732.

\bibitem[{Tenney et~al.(2019)Tenney, Das, and Pavlick}]{tenney2019bert}
Ian Tenney, Dipanjan Das, and Ellie Pavlick. 2019.
\newblock \href {http://arxiv.org/abs/1905.05950} {Bert rediscovers the
  classical nlp pipeline}.

\bibitem[{Tolstikhin et~al.(2021)Tolstikhin, Houlsby, Kolesnikov, Beyer, Zhai,
  Unterthiner, Yung, Steiner, Keysers, Uszkoreit, Lucic, and
  Dosovitskiy}]{tolstikhin_mlp-mixer_2021}
Ilya Tolstikhin, Neil Houlsby, Alexander Kolesnikov, Lucas Beyer, Xiaohua Zhai,
  Thomas Unterthiner, Jessica Yung, Andreas Steiner, Daniel Keysers, Jakob
  Uszkoreit, Mario Lucic, and Alexey Dosovitskiy. 2021.
\newblock \href {https://doi.org/10.48550/arXiv.2105.01601} {{MLP}-{Mixer}:
  {An} all-{MLP} {Architecture} for {Vision}}.
\newblock ArXiv:2105.01601 [cs].

\bibitem[{Vaswani et~al.(2017)Vaswani, Shazeer, Parmar, Uszkoreit, Jones,
  Gomez, Kaiser, and Polosukhin}]{vaswani_attention_2017}
Ashish Vaswani, Noam Shazeer, Niki Parmar, Jakob Uszkoreit, Llion Jones,
  Aidan~N. Gomez, Lukasz Kaiser, and Illia Polosukhin. 2017.
\newblock \href {http://arxiv.org/abs/1706.03762} {Attention {Is} {All} {You}
  {Need}}.
\newblock \emph{arXiv:1706.03762 [cs]}.
\newblock ArXiv: 1706.03762.

\bibitem[{Vig and Belinkov(2019)}]{vig-belinkov-2019-analyzing}
Jesse Vig and Yonatan Belinkov. 2019.
\newblock \href {https://doi.org/10.18653/v1/W19-4808} {Analyzing the structure
  of attention in a transformer language model}.
\newblock In \emph{Proceedings of the 2019 ACL Workshop BlackboxNLP: Analyzing
  and Interpreting Neural Networks for NLP}, pages 63--76, Florence, Italy.
  Association for Computational Linguistics.

\bibitem[{Voita et~al.(2019)Voita, Talbot, Moiseev, Sennrich, and
  Titov}]{voita-etal-2019-analyzing}
Elena Voita, David Talbot, Fedor Moiseev, Rico Sennrich, and Ivan Titov. 2019.
\newblock \href {https://doi.org/10.18653/v1/P19-1580} {Analyzing multi-head
  self-attention: Specialized heads do the heavy lifting, the rest can be
  pruned}.
\newblock In \emph{Proceedings of the 57th Annual Meeting of the Association
  for Computational Linguistics}, pages 5797--5808, Florence, Italy.
  Association for Computational Linguistics.

\bibitem[{Vyas et~al.(2020)Vyas, Katharopoulos, and Fleuret}]{vyas2020fast}
Apoorv Vyas, Angelos Katharopoulos, and François Fleuret. 2020.
\newblock \href {http://arxiv.org/abs/2007.04825} {Fast transformers with
  clustered attention}.

\bibitem[{Wang et~al.(2018)Wang, Singh, Michael, Hill, Levy, and
  Bowman}]{wang_glue_2018}
Alex Wang, Amanpreet Singh, Julian Michael, Felix Hill, Omer Levy, and Samuel
  Bowman. 2018.
\newblock \href {https://doi.org/10.18653/v1/W18-5446} {{GLUE}: {A}
  {Multi}-{Task} {Benchmark} and {Analysis} {Platform} for {Natural} {Language}
  {Understanding}}.
\newblock In \emph{Proceedings of the 2018 {EMNLP} {Workshop} {BlackboxNLP}:
  {Analyzing} and {Interpreting} {Neural} {Networks} for {NLP}}, pages
  353--355, Brussels, Belgium. Association for Computational Linguistics.

\bibitem[{Wang et~al.(2020)Wang, Li, Khabsa, Fang, and Ma}]{wang2020linformer}
Sinong Wang, Belinda~Z. Li, Madian Khabsa, Han Fang, and Hao Ma. 2020.
\newblock \href {http://arxiv.org/abs/2006.04768} {Linformer: Self-attention
  with linear complexity}.

\bibitem[{Wolf et~al.(2020)Wolf, Debut, Sanh, Chaumond, Delangue, Moi, Cistac,
  Rault, Louf, Funtowicz, Davison, Shleifer, von Platen, Ma, Jernite, Plu, Xu,
  Le~Scao, Gugger, Drame, Lhoest, and Rush}]{wolf_transformers_2020}
Thomas Wolf, Lysandre Debut, Victor Sanh, Julien Chaumond, Clement Delangue,
  Anthony Moi, Pierric Cistac, Tim Rault, Remi Louf, Morgan Funtowicz, Joe
  Davison, Sam Shleifer, Patrick von Platen, Clara Ma, Yacine Jernite, Julien
  Plu, Canwen Xu, Teven Le~Scao, Sylvain Gugger, Mariama Drame, Quentin Lhoest,
  and Alexander Rush. 2020.
\newblock \href {https://doi.org/10.18653/v1/2020.emnlp-demos.6} {Transformers:
  {State}-of-the-{Art} {Natural} {Language} {Processing}}.
\newblock In \emph{Proceedings of the 2020 {Conference} on {Empirical}
  {Methods} in {Natural} {Language} {Processing}: {System} {Demonstrations}},
  pages 38--45, Online. Association for Computational Linguistics.

\bibitem[{Xiao et~al.(2021)Xiao, Hu, Liu, Tu, and Sun}]{xiao_lawformer_2021}
Chaojun Xiao, Xueyu Hu, Zhiyuan Liu, Cunchao Tu, and Maosong Sun. 2021.
\newblock \href {https://doi.org/10.1016/j.aiopen.2021.06.003} {Lawformer: {A}
  pre-trained language model for {Chinese} legal long documents}.
\newblock \emph{AI Open}, 2:79--84.

\bibitem[{Yang et~al.(2020)Yang, Dai, Yang, Carbonell, Salakhutdinov, and
  Le}]{yang_xlnet_2020}
Zhilin Yang, Zihang Dai, Yiming Yang, Jaime Carbonell, Ruslan Salakhutdinov,
  and Quoc~V. Le. 2020.
\newblock \href {http://arxiv.org/abs/1906.08237} {{XLNet}: {Generalized}
  {Autoregressive} {Pretraining} for {Language} {Understanding}}.
\newblock \emph{arXiv:1906.08237 [cs]}.
\newblock ArXiv: 1906.08237.

\bibitem[{Zaheer et~al.(2021)Zaheer, Guruganesh, Dubey, Ainslie, Alberti,
  Ontanon, Pham, Ravula, Wang, Yang, and Ahmed}]{zaheer_big_2021}
Manzil Zaheer, Guru Guruganesh, Avinava Dubey, Joshua Ainslie, Chris Alberti,
  Santiago Ontanon, Philip Pham, Anirudh Ravula, Qifan Wang, Li~Yang, and Amr
  Ahmed. 2021.
\newblock \href {http://arxiv.org/abs/2007.14062} {Big {Bird}: {Transformers}
  for {Longer} {Sequences}}.
\newblock \emph{arXiv:2007.14062 [cs, stat]}.
\newblock ArXiv: 2007.14062.

\bibitem[{Zhang et~al.(2020)Zhang, Zhao, Saleh, and Liu}]{zhang_pegasus_2020}
Jingqing Zhang, Yao Zhao, Mohammad Saleh, and Peter~J. Liu. 2020.
\newblock \href {http://arxiv.org/abs/1912.08777} {{PEGASUS}: {Pre}-training
  with {Extracted} {Gap}-sentences for {Abstractive} {Summarization}}.
\newblock \emph{arXiv:1912.08777 [cs]}.
\newblock ArXiv: 1912.08777.

\end{thebibliography}
\bibliographystyle{styles/acl_natbib}

\appendix

\section{Fourier-class Architecture}
\label{sec:fourier_architecture}
In this section, we discuss the various Fourier-class architectures we have tested without having the hoped-for successes, with the idea of finding a better configuration for tasks requiring long documents. Specifically aside from the Hartley Layer already conveniently discussed in the main paper we tested, 4 other configurations (all of which involve real input and produce real output): 

\begin{enumerate}
    \item \textbf{Absolute value}: as also verified by \cite{lee-thorp_fnet_2021}, taking the absolute value after performing the 2D \ac{FT} strongly deteriorates the performance of the model. We observed that the \ac{MLM} losses (eval and train) achieve a convergence value around 6.
    \item \textbf{Phase}: as for the absolute value, even here propagating to the next \ac{FF} layer the phase of complex-value 2D \ac{FT} strongly deteriorates the performance of the mdoel. We observed even here that the \ac{MLM} losses (eval and train) achieve a convergence value around 6.
     \item \textbf{\ac{IFT}}: in similar way to the \ac{RFT}, we propagate the real projection of the imaginary term of the 2D \ac{FT} output (the second term of eq. \ref{eq:Fnetequation}), not surprising we got similar \ac{MLM} loss curves of \ac{RFT} and equivalent performance in the LexGLUE benchmark.
     \item \textbf{\ac{IFT} with \ac{RFT} memory}: in this case, we pretrain the weights using a RFT-based architecture, and then continue pretraining using an IFT-based architecture. During the transition from \ac{RFT} to \ac{IFT}  after an initial phase  of settling of the loss functions ($\sim$ 2k steps), the \ac{LM} showed a further marked phase of learning, which made us deduce the importance of propagating two terms of Eq. \ref{eq:Fnetequation} to the next \ac{FF} layer. Given the difficulty of training such architecture, we opted for a solution that included the two terms without two training stages (e.g. the Hartley Layer). 
 
\end{enumerate}

\section{Tokenizer}
\label{sec:tokenizer}

We trained a Unigram tokenizer \cite{kudo_subword_2018} similar to \cite{lan_albert_2020, yang_xlnet_2020, raffel_exploring_2020,zaheer_big_2021}, to encode the complicated legal language. We chose the standard 32k tokens as vocabulary size. We trained the tokenizer using the \ac{HF} tokenizers library\footnote{\url{https://github.com/huggingface/tokenizers}} on the entire Pretraining dataset ($\sim$ 23GB, $\sim$ 450k documents), covering English legal texts, mostly from the US. Additionally, we applied few preprocessing/cleaning steps on the input texts as Metaspace, lower case, individual digits and normalized punctualization\footnote{ https://github.com/moses-smt/mosesdecoder/blob/master/scripts/tokenizer/normalize-punctuation.perl, in \ac{HF}}.

\section{Pretraining}
\label{appendix:pretraining}

We trained 3 \ac{LM}s using the \ac{MLM} task, 1 HNet model with sequence length 4096 and 2 FNet models with sequence length 4096 and 8192 on the Pretraining dataset (9.2 GB, 2.7k  document tokens average, 504.2k documents). Our validation set consisted of $\sim$ 50k randomly selected examples, model was evaluated at each epoch to save compute. Please see Table \ref{tab:pretraining_data} for more details.

To maximise the legal text employed in the models during the \ac{MLM} stage by avoiding excessive pad tokens or excessive discarded text due to truncation, we concatenated all the examples and then cut them off in slices of the model's maximum sequence length (4096 and 8192, respectively). We did this in batches with multiprocessing to speed up data preparation. The last slice in each batch will not contain the established number of tokens, so we dropped it.

We trained two 4096-sequence models (HNet-4096 and FNet-4096, 85.6M parameters each) up to $\sim$ 1M of steps with a batch size of 2 for the first 500k steps and of 16 for the next steps\footnote{We played with gradient accumulation by moving it from 1 to 8.}; and one 8192-sequence model (FNet-8192, 88.8M parameters) up-to $\sim$ 800k of steps with a batch size of 1 for the first 700k steps and of 8 for the next steps. 
As can be seen, we decided to have a low batch size at the beginning and then increasing. This is because, in the first phase, we intend to update the weights of the model frequently, so as to quickly reduce losses and rapidly understand the basic relationships/characteristics most closely related to individual documents. Only at a later stage, we focus on the robustness of the model, placing more emphasis on relationships acting on multiple documents.
Please find more details in Appendix \ref{subsec:pretrained}.

HNet-4096 and Fnet-4096 took about 20 days each while FNet-8192 took 10 days on 16GB NVIDIA V100 GPU. The achieved training and evaluation losses are shown in Table \ref{tab:pretrain_losses}. 

\begin{table}[t]
\centering
\resizebox{\columnwidth}{!}{
    \begin{tabular}{llrrr}
    \toprule
    Model    & \# Steps  & Train Loss  & Eval Loss \\ 
    \midrule
    HNet-4096 &  250K  &   1.490 &     1.610 \\
    HNet-4096  & 500K  &   1.443 &     1.575 \\
    HNet-4096  & 750K  &   1.384 &     1.515 \\
    \textbf{HNet-4096}  & \textbf{1000K}  &  \textbf{1.359} &     \textbf{1.463} \\
    \midrule
    
    FNet-4096 &  250K  &   1.762 &     1.885 \\
    FNet-4096  & 500K  &   1.414 &     1.513 \\
    FNet-4096  & 750K  &   1.309 &     1.406 \\
    \textbf{FNet-4096}  & \textbf{1000K}  &  \textbf{1.281} &     \textbf{1.384} \\

    \midrule
    FNet-8192 &  250K  &   2.064 &     2.158 \\
    FNet-8192  & 500K  &   1.795 &     1.913 \\
    \textbf{FNet-8192}  & \textbf{800K}  &   \textbf{1.615} &     \textbf{1.713} \\

    \bottomrule
    \end{tabular}
}
\caption{Training and Evaluation losses for the different trained models.\label{tab:pretrain_losses}}
\end{table}

\subsection{Reproducibility Details}
\label{subsec:pretrained}

We pretrained our models with a learning rate of 5e-5, and a weight decay of 0.01 using 500 warm-up steps.
In addition, we used for both Hnet and Fnet models the same model config used in Huggingface\footnote{https://huggingface.co/docs/transformers/model\_doc/fnet} where we updated the maximum sequence and bos and eos token ids based on our tokenizer.
We used also the standard \ac{MLM} probability of 15\%.

For running the pretraining, we used an AWS p3.2xlarge instance with a 16GB NVIDIA V100 GPU. Training the three models (HNet-4096, Fnet-4096, and FNet-8192), took approx. 50 GPU days in total. Previous debug runs additionally consumed approx. 5 GPU days.

\section{Downstream Benchmarks}
\label{appendix:downstream_benchmark}


\subsection*{BillSum and PubMed}
When finetuning on the BillSum dataset \cite{kornilova_billsum_2019}  and on the PubMed summarization task \cite{cohan_discourse-aware_2018}, we trained using early stopping with patience of 3 epochs. We paired our pretrained encoder model with a randomly initialized bart-base decoder model \cite{lewis_bart_2020},  using the \ac{RAED} architecture in Fig. \ref{fig:halfatt_encdec}. We have randomly initialized the decoder since by construction we had to use a different tokenizer. This assumption proved to be valid as it led to better results than the use of the weights from the pretrained huggingface checkpoint\footnote{\url{https://huggingface.co/facebook/bart-base}}. We used the bart-base default config except 2 as no\_repeat\_ngram\_size. 
We set the maximum input length to 4096 and the maximum target length to 512. We found that, in using a lower decoder sequence length, many summaries get cut off (please, see Appendix \ref{sec:data_details} for data statistics). 
Due to high training costs, we only trained it with one random seed (42).
Our models contain 85.6M (4096-HNet/4096-FNet) and 88.8M (8192-FNet) parameters in the encoder and 96M parameters in the decoder for a total of 182M, and 185M parameters, respectively.

\subsection*{LexGLUE}
Finally, we evaluated on LexGLUE \cite{chalkidis_lexglue_2021} using the publicly available scripts without modification to ensure consistent and comparable results. Because of compute limitations, we ran each experiment with only 2 random seeds (1,2) and with the default set of \ac{HP} for equivalent comparison. We can conjecture that these set of \ac{HP} are mainly based on BERT-like models, since many models tested by \citet{chalkidis_lexglue_2021} come from a BERT model config (e.g. BERT, RoBERTa, Legal-BERT, Longformer, etc.). That also explains why many tasks in LexGLUE require less than 512 tokens.

\subsection{Reproducibility Details}

For running the finetuning experiments, we used an AWS p3.2xlarge instance with 16GB NVIDIA V100 GPUs. Running the BillSum, PubMed, and LexGLUE experiments including hyperparameter tuning took approximately 4, 19, and 10\footnote{5 days for each seed.}  GPU days  in total respectively.

\section{Detailed Results}
\label{sec:detailed_results}

\begin{table*}[t]
\centering
\resizebox{\textwidth}{!}{
    \begin{tabular}{lrrrrrrrrrrrr}
    \toprule
    model           & ECtHR A       & ECtHR B       & SCOTUS        & EUR-LEX       & LEDGAR        & UNFAIR-ToS    & CaseHOLD  & Average \\ 
    \midrule
    base models\\
    \midrule
    BERT                & 71.2 / 63.6 & 79.7 / 73.4 & 68.3 / 58.3 & 71.4 / 57.2 & 87.6 / 81.8 & 95.6 / 81.3 & 70.8 & 77.8 / 69.5\\
    RoBERTa             & 69.2 / 59.0 & 77.3 / 68.9 & 71.6 / 62.0 & 71.9 / 57.9 & 87.9 / 82.3 & 95.2 / 79.2 & 71.4 & 77.8 / 68.7\\
    DeBERTa             & 70.0 / 60.8 & 78.8 / 71.0 & 71.1 / 62.7 & 72.1 / 57.4 & 88.2 / 83.1 & 95.5 / 80.3 & 72.6 & 78.3 / 69.7\\
    BigBird             & 70.0 / 62.9 & 78.8 / 70.9 & 72.8 / 62.0 & 71.5 / 56.8 & 87.8 / 82.6 & 95.7 / 81.3 & 70.8 & 78.2 / 69.6\\
    Longformer          & 69.9 / 64.7 & 79.4 / 71.7 & 72.9 / 64.0 & 71.6 / 57.7 & 88.2 / 83.0 & 95.5 / 80.9 & 71.9 & 78.5 / 70.5\\
    CaseLawBERT         & 69.8 / 62.9 & 78.8 / 70.3 & 76.6 / 65.9* & 70.7 / 56.6 & 88.3 / 83.0 & 96.0 / 82.3 & 75.4* & 79.4 / 70.9\\
    LegalBERT-base      & 70.0 / 64.0* & 80.4 / 74.7* & 76.4 / 66.5* & 72.1 / 57.4* & 88.2 / 83.0* & 96.0 / 83.0 & 75.3* & 79.8 / 72.0\\
    FNet \cite{lee-thorp_fnet_2021}  & 57.1 / 46.4	& 65.7 / 56.4 &	60.5 / 46.5	& 65.2 / 46.5 & 85.6 / 80.1	& 95.3 / 78.0 & 50.9 & 68.6 / 57.8 \\
    FNet-4096  &  64.0  / 55.6 &	71.3  / 58.8 &	70.2  / 59.5 & 64.3  / 44.6 & 85.7  / 79.5 &	93.8  / 70.6 &	50.4 & 71.4 / 59.9 \\ 
    FNet-8192   & -  & - & 70.1 / 59.6 & - & - & - & -  & -\\ 
    HNet-4096   & 62.0  / 52.7 &	70.4  / 54.9 &	72.2  / 63.4 &	65.6  / 47.3 &	85.5 / 78.9 &	93.5  / 71.0 &	63.1 & 73.2 / 61.6\\ 
    \bottomrule
    \end{tabular}
}
\caption{Results on LexGLUE. Because of limited compute, we only ran 2 random seed for our models (seed 1, and 2). The other results are reported on GitHub\footnote{\url{https://github.com/coastalcph/lex-glue/discussions/categories/new-results}}. The asterix denotes datasets which are (partly) covered in the pretraining dataset. For each column we report the results in the format micro-averaged F1 score / macro-average F1 score. For the CaseHOLD task, both scores are the same. We performed only the Scotus task for FNet-8192 since it is the unique task which involves long-documents. 
}
\label{tab:lexglue_results}
\end{table*}

In this section, we show detailed and comprehensive results of the compared models on the LexGLUE benchmark (Table \ref{tab:lexglue_results}).

\section{Library Versions}
We used the following versions to the libraries in a pip requirements.txt format:\\
datasets==2.1.0\\
jsonlines==3.0.0\\
sentencepiece==0.1.96\\
transformers==4.18.0\\
rouge-score==0.1.1\\
torch==1.10.1\\

\section{Examples}
Example summaries are displayed in Tables 
\ref{tab:examples_hybrid_fnet-billsum}, 
\ref{tab:examples_hybrid_hnet-billsum}, \ref{tab:examples_hybrid_fnet_ccdv-pubmed-summarization}, \ref{tab:examples_hybrid_fnet_8192_ccdv-pubmed-summarization}, and \ref{tab:examples_hybrid_hnet_ccdv-pubmed-summarization}. Since the documents are very long sometimes, we truncated them to the first 2500 characters. We sorted the examples by RougeL scores and show the bottom 5\%, bottom 25\%, top 75\% and top 95\% percentile. 

\renewcommand{\arraystretch}{0.01} 

\begin{table*}[t]
\centering
\resizebox{14cm}{!}{
    \begin{tabular}{ l p{20cm} }
    \toprule
\\ 
 \midrule 
\multicolumn{2}{l}{Bottom 5\% example (Sorted by fmeasure)}\\ 
 \midrule 
Document & \tiny SECTION 1. SHORT TITLE.

    This Act may be cited as the ``Advanced Nuclear Fuel Availability 
Act''.

SEC. 2. PROGRAM.

    (a) Establishment.--The Secretary shall establish and carry out, 
through the Office of Nuclear Energy, a program to support the 
availability of HA-LEU for domestic commercial use.
    (b) Program Elements.--In carrying out the program under subsection 
(a), the Secretary--
            (1) may provide financial assistance to assist commercial 
        entities to design and license transportation packages for HA-
        LEU, including canisters for metal, gas, and other HA-LEU 
        compositions;
            (2) shall, to the extent practicable--
                    (A) by January 1, 2021, have commercial entities 
                submit such transportation package designs to the 
                Commission for certification by the Commission under 
                part 71 of title 10, Code of Federal Regulations; and
                    (B) encourage the Commission to have such 
                transportation package designs so certified by the 
                Commission by January 1, 2023;
            (3) not later than January 1, 2020, shall submit to 
        Congress a report on the Department's uranium inventory that 
        may be available to be processed to HA-LEU for purposes of such 
        program, which may not include any uranium allocated by the 
        Secretary for use in support of the atomic energy defense 
        activities of the National Nuclear Security Administration;
            (4) not later than 1 year after the date of enactment of 
        this Act, and biennially thereafter through September 30, 2025, 
        shall conduct a survey of stakeholders to estimate the quantity 
        of HA-LEU necessary for domestic commercial use for each of the 
        5 subsequent years;
            (5) shall assess options available for the Secretary to 
        acquire HA-LEU for such program, including an assessment, for 
        each such option, of the cost and amount of time required;
            (6) shall establish a consortium, which may include 
        entities involved in any stage of the nuclear fuel cycle, to 
        partner with the Department to support the availability of HA-
        LEU for domestic commercial use, including by--
                    (A) providing information to the Secretary for 
                purposes of surveys conducted under paragraph (4); and
                    (B) ... \\ 
 \midrule 
Gold & \tiny Advanced Nuclear Fuel Availability Act This bill directs the Department of Energy to develop and deploy high-assay low-enriched uranium for domestic commercial use and to develop a schedule for recovering costs associated with such development. \\ 
 \midrule 
Model & \small . advanced nuclear fuel availability act (sec. 2)this bill requires the nuclear regulatory commission (nrc) to establish and carry out, through the office of nuclear energy, a program to support the availability of ha-leu for domestic commercial use. in carrying out the program, the department of energy (doe) may provide financial assistance to assist commercial entities to design and license transportation packages for its ball, including can tribes for metal, gas, and other hamu wells. (this will: (1) submit to congress on the department's uranium inventory that may be available to be processed to unexpan for purposes of such program, and to encourage the nrc to have such transportation package designs so certified by january 1,2021; and (3) conduct biennially thereafter, to assess the quantity of uninu necessary for such program. the bill sets forth reporting requirements. additionally, doe may not make specified commitments under this bill. additionally, doj shall report on its conversion to its nuclear technology, including its licensing and regulation, ownership, or other acquisition. \\ 
 \midrule 
Metrics & Rouge1: 0.22, RougeL: 0.18, RougeLsum: 0.18
 \\ 
\\ 
 \midrule 
\multicolumn{2}{l}{Bottom 25\% example (Sorted by fmeasure)}\\ 
 \midrule 
Document & \tiny SECTION 1. SHORT TITLE.

    This Act may be cited as the ``Airport Streamlining Approval 
Process Act of 2002''.

SEC. 2. FINDINGS.

    Congress finds that--
            (1) airports play a major role in interstate and foreign 
        commerce;
            (2) congestion and delays at our Nation's major airports 
        have a significant negative impact on our Nation's economy;
            (3) airport capacity enhancement projects at congested 
        airports are a national priority and should be constructed on 
        an expedited basis;
            (4) airport capacity enhancement projects must include an 
        environmental review process that provides local citizenry an 
        opportunity for consideration of and appropriate action to 
        address environmental concerns; and
            (5) the Federal Aviation Administration, airport 
        authorities, communities, and other Federal, State, and local 
        government agencies must work together to develop a plan, set 
        and honor milestones and deadlines, and work to protect the 
        environment while sustaining the economic vitality that will 
        result from the continued growth of aviation.

SEC. 3. PROMOTION OF NEW RUNWAYS.

    Section 40104 of title 49, United States Code, is amended by adding 
at the end the following:
    ``(c) Airport Capacity Enhancement Projects at Congested 
Airports.--In carrying out subsection (a), the Administrator shall take 
action to encourage the construction of airport capacity enhancement 
projects at congested airports as those terms are defined in section 
47179.''.

SEC. 4. AIRPORT PROJECT STREAMLINING.

    (a) In General.--Chapter 471 of title 49, United States Code, is 
amended by inserting after section 47153 the following:

             ``SUBCHAPTER III--AIRPORT PROJECT STREAMLINING

``Sec. 47171. DOT as lead agency
    ``(a) Airport Project Review Process.--The Secretary of 
Transportation shall develop and implement a coordinated review process 
for airport capacity enhancement projects at congested airports.
    ``(b) Coordinated Reviews.--The coordinated review process under 
this section shall provide that all environmental reviews, analyses, 
opinions, permits, licenses, and approvals that must be issued or made 
by a Federal agency or airport sponsor for an airport capacity 
enhancement project at a congested airport will be conducted 
concurrently, to the maximum extent practicable, and completed within a 
time ... \\ 
 \midrule 
Gold & \tiny Airport Streamlining Approval Process Act of 2002 - (Sec. 3) Amends Federal transportation law to direct the Administrator of the Federal Aviation Administration (FAA) to take action to encourage the construction of airport capacity enhancement projects at congested airports.(Sec. 4) Directs the Secretary of Transportation to develop and implement a coordinated review process for such projects, which shall provide that all environmental reviews, analyses, opinions, permits, licenses, and approvals that must be issued or made by a Federal agency or airport sponsor for such a project will be conducted concurrently in cooperation with all Federal and State agencies with jurisdiction over environmental-related matters.Requires: (1) the Secretary determine the reasonable alternatives to an airport capacity enhancement project at a congested airport; and (2) any other participating Federal or State agency to consider only those alternatives the Secretary has determined are reasonable.Authorizes the Secretary, at the request of an airport sponsor for a congested airport, to approve a restriction on use of a runway to be constructed at the airport to minimize potentially significant adverse noise impacts from the runway only if the restriction is necessary and the most appropriate and cost-effective measure (taking into consideration associated environmental trade-offs) to mitigate such impacts and expedite runway construction.Authorizes the Secretary, in specified circumstances, to allow an airport sponsor carrying out such a project to make payments, out of airport revenues (including local taxes on aviation fuel), for measures to mitigate the environmental impacts of the project, including aircraft noise.Permits the FAA Administrator to accept funds from an airport sponsor to hire additional staff or obtain the services of consultants in order to facilitate the timely processing, review, and completion of environmental activities associated with an airport development project.Authorizes appropriations to facilitate timely processing, review, and completion of environmental activities associated with airport capacity enhancement projects at congested airports.Permits a person disclosing a substantial interest in an order issued by the Secretary or the head of any other pertinent Federal agency to apply for judicial review of the order. Prescribes procedures and requirements for such an appeal.(Sec. 5) Repeals the requirement that the Secretary approve a project grant application only if the chief executive officer of the State in which the project will be located certifies that there is reasonable assurance that the project will be located, designed, constructed, and operated in compliance with applicable air and water quality standards.Revises the approval criteria without the requirement of an environmental impact statement for an airport development project that does not involve the location of an airport or runway, or a major runway extension, at an existing airport. Allows such a project without an environmental impact statement if completing the project would allow airport operations involving aircraft complying with the noise standards prescribed for "stage 3" aircraft (currently "stage 2" aircraft).(Sec. 6) Authorizes the Secretary to incur obligations to make grants to an operator of a congested airport and a specified unit of local government to carry out a project to mitigate noise in the area surrounding the airport, if the project is included as a commitment in an FFA record of decision for an airport capacity enhancement project, even if that airport has not met certain regulatory requirements. \\ 
 \midrule 
Model & \small air streamlining approval process act of 2002 - directs the administrator of the federal aviation administration (faa) to develop and implement a coordinated review process for airport capacity enhancement projects at miles airports. requires the secretary of transportation to: (1) identify, as soon as practicable, all federal and state agencies that may have jurisdiction over environmental-related matters; and (2) determine the reasonable alternatives to airport operations enhancement project at airports, including local taxes on aviation fuel, for measures to mitigate the environmental impacts of such projects.authorizes waivers of certain federal or state authority to conduct an environmental review or analysis of an applicant, license, or approval for the project, subject to the process process.allows an airport refinery to make payments, out of revenues generated at the airport (including local taxes. requires any person acquired under this act to be credited as offsetting collections (to the costs of activities and services for which the funds are accepted). \\ 
 \midrule 
Metrics & Rouge1: 0.41, RougeL: 0.26, RougeLsum: 0.26
 \\ 
\\ 
 \midrule 
\multicolumn{2}{l}{Top 75\% example (Sorted by fmeasure)}\\ 
 \midrule 
Document & \tiny SECTION 1. SHORT TITLE.

    This Act may be cited as the ``Contributions Legally Interdicted 
from Noncitizens To Our Nonprofits Act of 2016'' or as the ``CLINTON 
Act of 2016''.

SEC. 2. CERTAIN CHARITABLE ORGANIZATIONS PROHIBITED FOR ACCEPTING 
              CONTRIBUTIONS FROM PERSONS CONNECTED TO FOREIGN 
              GOVERNMENTS.

    (a) In General.--Section 501 of the Internal Revenue Code of 1986 
is amended by adding at the end the following new subsection:
    ``(s) Prohibition on Acceptance of Contributions From Persons 
Connected to Foreign Governments by 501(c)(3) Organizations Established 
by Certain Federal Officials.--
            ``(1) Termination of tax-exempt status.--A Federal official 
        organization shall not be treated as described in subsection 
        (c)(3) with respect to any period after the date on which such 
        organization knowingly or willingly accepts or solicits any 
        contribution from any person connected to a foreign government. 
        If a Federal official organization accepts a contribution from 
        any person and learns that such person is connected to a 
        foreign government after such acceptance, such organization 
        shall not be treated for purposes of this subsection as 
        knowingly or willingly accepting such contribution from such 
        person if such contribution is returned to such person not 
        later than the date which is 30 days after the date on which 
        the organization so learns the status of such person.
            ``(2) Forfeiture of prohibited contributions.--There is 
        hereby imposed a tax on any Federal official organization which 
        knowingly or willingly accepts any contribution from any person 
        connected to any foreign government in an amount equal to the 
        amount of such contribution.
            ``(3) Federal official organization.--For purposes of this 
        subsection, the term `Federal official organization' means any 
        organization described in subsection (c)(3) (or which would be 
        so described without regard to paragraph (1)) if one or more 
        Federal officials established, control, or actively participate 
        in the management of, such organization. For purposes of the 
        preceding sentence, service in a merely honorary capacity shall 
        note be treated as control or active participation in 
        management.
            ``(4) Federal official.--For purposes... \\ 
 \midrule 
Gold & \tiny Contributions Legally Interdicted from Noncitizens To Our Nonprofits Act of 2016 or the CLINTON Act of 2016  This bill amends the Internal Revenue Code to prohibit a federal official organization from being treated as a tax-exempt organization under section 501(c)(3) for any period after the date on which the organization knowingly or willingly accepts or solicits any contribution from any person connected to a foreign government.  A tax-exempt organization is a \&quot;federal official organization\&quot; if one or more current or former specified federal officials established, control, or actively participate in the management of the organization.  If the organization accepts a contribution from any person and learns that that the person is connected to a foreign government after the acceptance, it will not be treated as knowingly or willingly accepting the contribution if the contribution is returned within 30 days of learning of the status of the person.  The bill imposes a tax on any federal official organization that knowingly or willingly accepts any contribution from any person connected to any foreign government in an amount equal to the amount of the contribution.  \\ 
 \midrule 
Model & \small contributions legally interdicted from noncitizens to our nonprofits act of 2016 or the cike act (sec.2) this bill amends the internal revenue code to prohibit a federal official from being treated for any period after the date on which such organization: (1) knowingly or willfully accepts or soliciting any contribution from any person and learns that such person is connected to foreign government after such acceptance, the organization must not be treated as prohibited, or to willfully accepting such contribution. the bill: imposes penalties on any federal individual and any individual who knowingly accepts contributions from such individual; restricts the provision of such contributions to such foreign governments; and impose an additional 10-year limitation on the amount of donations that may be contributed to the person within 30 days after receiving the contribution, withholding of prohibited contributions, and the right to which the individual is accepted or held any of the following positions in the federal government, including any officer or employee appointed by the president, president, or any agency with the president or vice chief of staff, at any time during the two-year period ending on such date, has been held at the pleasure of that person's next of kin and subject to confirmation by that procedure. all such prohibition shall be imposed at no prior u.s. expense. \\ 
 \midrule 
Metrics & Rouge1: 0.58, RougeL: 0.4, RougeLsum: 0.4
 \\ 
\\ 
 \midrule 
\multicolumn{2}{l}{Top 95\% example (Sorted by fmeasure)}\\ 
 \midrule 
Document & \tiny SECTION 1. SHORT TITLE.

    This Act may be cited as the ``Chesapeake Bay Environmental 
Education Pilot Program Act''.

SEC. 2. FINDINGS.

    Congress finds that--
            (1) increasing public environmental awareness and 
        understanding through formal environmental education and 
        meaningful bay or stream field experiences are vital parts of 
        the effort to protect and restore the Chesapeake Bay ecosystem;
            (2) using the Chesapeake Bay watershed as an integrating 
        context for learning can help--
                    (A) advance student learning skills;
                    (B) improve academic achievement in core academic 
                subjects; and
                    (C)(i) encourage positive behavior of students in 
                school; and
                    (ii) encourage environmental stewardship in school 
                and in the community; and
            (3) the Federal Government, acting through the Secretary of 
        Education, should work with the Under Secretary for Oceans and 
        Atmosphere, the Chesapeake Executive Council, State educational 
        agencies, elementary schools and secondary schools, and 
        nonprofit educational and environmental organizations to 
        support development of curricula, teacher training, special 
        projects, and other activities, to increase understanding of 
        the Chesapeake Bay watershed and to improve awareness of 
        environmental problems.

SEC. 3. CHESAPEAKE BAY ENVIRONMENTAL EDUCATION AND TRAINING GRANT PILOT 
              PROGRAM.

    Title IV of the Elementary and Secondary Education Act of 1965 (20 
U.S.C. 7101 et seq.) is amended by adding at the end the following:

  ``PART D--CHESAPEAKE BAY ENVIRONMENTAL EDUCATION AND TRAINING GRANT 
                             PILOT PROGRAM

``SEC. 4401. DEFINITIONS.

    ``In this part:
            ``(1) Bay watershed state.--The term `Bay Watershed State' 
        means each of the States of Delaware, Maryland, New York, 
        Pennsylvania, Virginia, and West Virginia, and the District of 
        Columbia.
            ``(2) Chesapeake executive council.--The term `Chesapeake 
        Executive Council' has the meaning given the term in section 
        307(d) of the National Oceanic and Atmospheric Administration 
        Authorization Act of 1992 (15 U.S.C. 1511d(d)).
            ``(3) Eligible institution.--The term `eligible 
        institution' means--
      ... \\ 
 \midrule 
Gold & \tiny Chesapeake Bay Environmental Education Pilot Program Act - Amends the Elementary and Secondary Education Act of 1965 to direct the Secretary of Education to establish a Chesapeake Bay Environmental Education Grant Pilot Program of matching grants for developing, demonstrating, or disseminating information on practices, methods, or techniques relating to environmental education and training in the Chesapeake Bay watershed. 
 Makes eligible for such grants public elementary or secondary schools and nonprofit environmental or educational organizations located in the District of Columbia, Delaware, Maryland, New York, Pennsylvania, Virginia, or West Virginia. \\ 
 \midrule 
Model & \small chesapeake bay environmental education pilot program act - amends the elementary and secondary education act of 1965 to direct the secretary of education to establish a grant program, administered by the national fish and wildlife foundation, under secretary for oceans and atmosphere, to make grants to eligible institutions to pay the federal share of the costs of developing, demonstrating, or disseminating information on practices, methods, or techniques relating to environmental education, and training in such watershed. authorizes appropriations. \\ 
 \midrule 
Metrics & Rouge1: 0.65, RougeL: 0.56, RougeLsum: 0.56
 \\ 

    \bottomrule
    \end{tabular}
}
\caption{Examples of the billsum dataset using the model RAED FNet-4096}
\label{tab:examples_hybrid_fnet-billsum}
\end{table*}

\begin{table*}[t]
\centering
\resizebox{14cm}{!}{
    \begin{tabular}{ l p{20cm} }
    \toprule
\\ 
 \midrule 
\multicolumn{2}{l}{Bottom 5\% example (Sorted by fmeasure)}\\ 
 \midrule 
Document & \tiny SECTION 1. SHORT TITLE.

    This Act may be cited as the ``Uzbekistan Freedom Promotion Act of 
2005''.

SEC. 2. FINDINGS.

    Congress finds the following:
            (1) The Department of State's 2004 Country Reports on Human 
        Rights Practices for Uzbekistan states the following:
                    (A) ``Uzbekistan is an authoritarian state with 
                limited civil rights . . . President Islam Karimov and 
                the centralized executive branch that serves him 
                dominated political life and exercised nearly complete 
                control over the other branches . . . The Government is 
                highly centralized and is ruled by President Karimov 
                and the executive branch through sweeping decree 
                powers, primary authority for drafting legislation, and 
                control of government appointments, most of the 
                economy, and the security forces.''.
                    (B) ``On December 26, [2004,] elections were held 
                for seats in the lower chamber of the Supreme Assembly 
                (Oliy Majlis) that fell significantly short of 
                international standards for democratic elections . . . 
                Citizens could not exercise the right to change their 
                government peacefully . . . The law makes it extremely 
                difficult for opposition parties to organize, nominate 
                candidates, and campaign.''.
                    (C) ``Police and, to a lesser extent, NSS [National 
                Security Service] forces tortured, beat, and harassed 
                persons . . . Members of the security forces 
                responsible for documented abuses were rarely punished 
                . . . Police, prison officials, and the NSS allegedly 
                used suffocation, electric shock, rape, and other 
                sexual abuse; however, beating was the most commonly 
                reported method of torture.''.
                    (D) ``The number of persons in prison for political 
                or religious reasons--primarily persons the Government 
                believed were associated with extremist Islamist 
                political groups, but also members of the secular 
                opposition and human rights activists--was estimated to 
                be between 5,000 and 5,500.''.
                    (E) ``The Government severely restric... \\ 
 \midrule 
Gold & \tiny Uzbekistan Freedom Promotion Act of 2005 - Amends the Foreign Assistance Act of 1961 to prohibit assistance, with specified exceptions, to the government of Uzbekistan unless the President certifies to the appropriate congressional committees that the government of Uzbekistan: (1) is cooperating with an independent international inquiry into the Andijan violence, has ceased pressuring the government of Kyrgyzstan to return the refugees who fled the massacre, and is prosecuting responsible individuals; and (2) has accelerated democratic reforms and fulfilled its human rights obligations.

Prohibits visa issuance to any official of the government of Uzbekistan who is credibly alleged to have ordered, acquiesced to, or participated in human rights abuses or corruption unless the Secretary of State certifies to the appropriate congressional committees that the visa issuance is in U.S. national security interests.

Directs the President to: (1) freeze the financial assets and other economic resources in the United States of any official (or family member) of the government of Uzbekistan who is credibly alleged to have ordered, acquiesced to, or participated in human rights abuses, including the massacre in Andijan. (Permits lifting of such freeze if the President certifies to the appropriate congressional committees that the official or individual has been properly investigated and cleared of any wrongdoing.); (2) prohibit munitions or dual use exports to Uzbekistan unless the President certifies that the government of Uzbekistan has met the requirements of the first paragraph above; (3) expedite U.S. admission of any Uzbekistan national who is under threat of severe penalty as a result of participating in pro-democracy activities, including Uzbekistan citizens who have fled to Kyrgyzstan; and (4) transmit to the appropriate congressional committees a plan to move all U.S. military operations from Uzbekistan. \\ 
 \midrule 
Model & \small bebec freedom promotion act of 2005 - expresses the sense of congress the u.s. actions of the government of samuelbekistan in the case of andiser massacre specifically: (1) are harassing, totalitarian, and condoned of an ally; (2) counterproductive to the united states goals of promoting freedom and democracy; and (3) justify an end to u.s. assistance for such government. \\ 
 \midrule 
Metrics & Rouge1: 0.22, RougeL: 0.16, RougeLsum: 0.16
 \\ 
\\ 
 \midrule 
\multicolumn{2}{l}{Bottom 25\% example (Sorted by fmeasure)}\\ 
 \midrule 
Document & \tiny SECTION 1. REFERENCES TO SOCIAL SECURITY ACT.

    Except as otherwise specifically provided, whenever in this Act an 
amendment is expressed in terms of an amendment to or repeal of a 
section or other provision, the reference shall be considered to be 
made to that section or other provision of the Social Security Act.

SEC. 2. DURABLE MEDICAL EQUIPMENT.

    (a) Definition of Medical Equipment and Supplies.--
            (1) In general.--Section 1861 (42 U.S.C. 1395x) is amended 
        by adding at the end the following new subsection:

                    ``medical equipment and supplies

    ``(oo) The term `medical equipment and supplies'' means--
            ``(1) durable medical equipment (as defined in section 
        1861(n));
            ``(2) prosthetic devices (as described in section 
        1861(s)(8));
            ``(3) orthotics and prosthetics (as described in section 
        1861(s)(9));
            ``(4) home dialysis supplies and equipment (as described in 
        section 1861(s)(2)(F));
            ``(5) surgical dressings and other devices (as described in 
        section 1861(s)(5));
            ``(6) immunosuppressive drugs (as described in section 
        1861(s)(2)(J)); and
            ``(7) such other items as the Secretary may determine.''.
            (2) Effective date.--The amendment made by paragraph (1) 
        shall apply to items furnished on or after January 1, 1994.
    (b) Development and Application of National Standards for Suppliers 
of Medical Equipment and Supplies.--Section 1834 (42 U.S.C. 1395m) is 
amended by adding at the end the following new subsection:
    ``(i) Requirements for Issuance and Renewal of Supplier Numbers for 
Suppliers of Medical Equipment and Supplies.--
            ``(1) Payment.--No payment may be made under this part 
        after July 1, 1994, for items furnished by a supplier of 
        medical equipment and supplies (as defined in section 1861(oo)) 
        unless such supplier meets the national standards specified by 
        the Secretary and possesses a valid supplier number.
            ``(2) Revised standards.--
                    ``(A) In general.--The Secretary shall, by no later 
                than January 1, 1996, in consultation with 
                representatives of suppliers of medical equipment and 
                supplies, carriers, and consumers, revise the national 
                standards for suppliers of medical equipment and 
                supp... \\ 
 \midrule 
Gold & \tiny Amends title XVIII (Medicare) of the Social Security Act (SSA) to revise procedures under Medicare part B (Supplementary Medical Insurance) with regard to durable medical equipment, including requirements for:  (1) national standards and supplier numbers for suppliers of medical equipment and supplies; (2) standardized certificates of medical necessity and uniform national coverage and utilization review criteria for certain medical equipment and supplies; (3) limited Medicare beneficiary liability for items and services for which payment is prohibited by reason of a supplier's failure to meet such national standards or lack of a valid supplier number; (4) the treatment of nebulizers and aspirators as miscellaneous items of durable medical equipment; and (5) payment of ostomy supplies, tracheostomy supplies, urologicals, surgical dressings, and other medical supplies. 
Amends SSA title XI to modify anti-kickback provisions. 
Provides for a freeze under Medicare part B in reasonable charges for parenteral and enteral nutrients, supplies, and equipment during 1994. 
Requires the Comptroller General to study and report to the Congress on:  (1) services and supplies furnished to Medicare-eligible nursing facility residents; and (2) changes made to descriptions relating to codes for medical equipment and supplies. \\ 
 \midrule 
Model & \small amends title xviii (medicare) of the social security act (ssa) to provide for the use and renewal of medical equipment and supplies. requires the secretary of health and human services to: (1) revise the standards for suppliers furnished by a supplier of certain medical necessity; (2) develop and establish uniform national coverage and utilization review criteria for 200 items; and (3) provide certificates for such certificates. \\ 
 \midrule 
Metrics & Rouge1: 0.43, RougeL: 0.26, RougeLsum: 0.26
 \\ 
\\ 
 \midrule 
\multicolumn{2}{l}{Top 75\% example (Sorted by fmeasure)}\\ 
 \midrule 
Document & \tiny SECTION 1. TWO-YEAR ELIGIBILITY FOR DEPARTMENT OF VETERANS AFFAIRS 
              HEALTH CARE FOR MEMBERS OF THE ARMED FORCES SERVING IN 
              AREAS AFFECTED BY HURRICANES KATRINA AND RITA.

    Section 1710(e) of title 38, United States Code, is amended--
            (1) by adding at the end of paragraph (1) the following new 
        subparagraph:
    ``(F) Subject to paragraphs (2) and (3), a member of the Armed 
Forces (including a member ordered to duty under section 502(f) of 
title 32) who is performing duty in response to a disaster or emergency 
declaration under the Robert T. Stafford Disaster Relief and Emergency 
Assistance Act (42 U.S.C. 5121 et seq.) after August 28, 2005, in an 
area affected by Hurricane Katrina or Hurricane Rita is eligible for 
hospital care, medical services, and nursing home care under subsection 
(a)(2)(F) for any disability, notwithstanding that there is 
insufficient medical evidence to conclude that such disability is 
attributable to such duty.'';
            (2) by adding at the end of paragraph (2) the following new 
        subparagraph:
    ``(C) In the case of a member of the Armed Forces described in 
paragraph (1)(F), hospital care, medical services, and nursing home 
care may not be provided under subsection (a)(2)(F) with respect to a 
disability that is found, in accordance to guidelines issued by the 
Under Secretary for Health, to have resulted from a cause other than 
duty described in that paragraph.'';
            (3) in paragraph (3)--
                    (A) by striking ``and'' at the end of subparagraph 
                (C);
                    (B) by striking the period at the end of 
                subparagraph (D) and inserting ``; and''; and
                    (C) by adding at the end the following new 
                subparagraph:
            ``(E) in the case of care for a veteran described in 
        paragraph (1)(F), after a period of two years beginning on the 
        date of the veteran's discharge or release from duty described 
        in that paragraph.''; and
            (4) by adding at the end of paragraph (4) the following new 
        subparagraph:
                    ``(C) The term `area affected by Hurricane Katrina 
                or Hurricane Rita' means an area designated for 
                individual assistance or public assistance by Federal 
                Disaster Declaration notice 1602, 1603, 1604, 1605, 
                1606, or 1607 issued by the Di... \\ 
 \midrule 
Gold & \tiny Makes a member of the Armed Forces who is performing duty in response to a disaster or emergency declaration after August 28, 2005, in an area affected by Hurricane Katrina or Rita eligible for hospital care, medical services, and nursing home care for any disability, notwithstanding insufficient medical evidence to conclude that the disability is attributable to such duty. Prohibits such care or services with respect to a disability found to have resulted from a cause other than such duty. Terminates eligibility two years after the member's discharge or release from such duty.

Directs the Secretaries of Defense and Veterans Affairs to enter into an agreement with the National Academy of Sciences to review and make recommendations for the health surveillance of members who may be exposed to environmental hazards as the result of duty in areas affected by Hurricanes Katrina and Rita in 2005.

Requires the Secretary of Veterans Affairs to maintain a database of members provided health care by the Department of Veterans Affairs pursuant to such duty. \\ 
 \midrule 
Model & \small amends the federal veterans code to allow a member of the armed forces who is performing duty in response to disaster or emergency declaration under the robert t. stafford disaster relief and emergency assistance act (aida), as well as insufficient medical evidence to conclude that such disability is attributable to such duty. requires the secretary of veterans affairs (va) to: (1) provide for the national academy of sciences; (2) review and make recommendations for health surveillance of members who may be exposed to environmental hazards as the result of duty; and (3) enter into an agreement with the nsec to review, to have resulted from an cause other than duty described in that hurricanes katrina or hurricane rita.rita, as an area designated for hospital care, medical services, and nursing home care for any disability, notwithstanding such period beginning on hurricane katrina and 28,500, in any county designated by the director of such federal emergency management agency (fema). requires the secretaries to report to the congressional committees on the activities. \\ 
 \midrule 
Metrics & Rouge1: 0.69, RougeL: 0.41, RougeLsum: 0.41
 \\ 
\\ 
 \midrule 
\multicolumn{2}{l}{Top 95\% example (Sorted by fmeasure)}\\ 
 \midrule 
Document & \tiny SECTION 1. AUTHORITY TO ISSUE A RULE RELATING TO ERGONOMICS.

    (a) Findings.--Congress makes the following findings:
            (1) The National Academy of Sciences issued a report 
        entitled `Musculoskeletal Disorders and the Workplace--Low Back 
        and Upper Extremities' on January 18, 2001. The report was 
        issued after the Occupational Safety and Health Administration 
        promulgated a final rule relating to ergonomics (published at 
        65 Fed. Reg. 68261 (2000)).
            (2) According to the National Academy of Sciences, 
        musculoskeletal disorders of the low back and upper extremities 
        are an important and costly national health problem. An 
        estimated 1,000,000 workers each year lose time from work as a 
        result of work-related musculoskeletal disorders.
            (3) Conservative estimates of the economic burden imposed 
        by work-related musculoskeletal disorders, as measured by 
        compensation costs, lost wages, and lost productivity, are 
        between \$45,000,000,000 and \$54,000,000,000 annually.
            (4) Congress enacted the Occupational Safety and Health Act 
        of 1970 (29 U.S.C. 651 et seq.) to `assure so far as possible 
        every working man and woman in the Nation safe and healthful 
        working conditions,' and charged the Secretary of Labor with 
        implementing the Act to accomplish this purpose.
            (5) Promulgation of a standard on workplace ergonomics is 
        needed to address a serious workplace safety and health problem 
        and to protect working men and women from work-related 
        musculoskeletal disorders. Any workplace ergonomics standard 
        should take into account the cost and feasibility of compliance 
        with such requirements and the sound science of the National 
        Academy of Sciences report.
    (b) Authority to Issue Rule.--
            (1) In general.--Notwithstanding any other provision of 
        law, not later than 2 years after the date of enactment of this 
        Act, the Secretary of Labor shall, in accordance with section 6 
        of the Occupational Safety and Health Act of 1970 (29 U.S.C. 
        655), issue a final rule relating to ergonomics. The standard 
        under the final rule shall take effect not later than 90 days 
        after the date on which the rule is promulgated.
            (2) Requirements for standard.--The standard described in 
      ... \\ 
 \midrule 
Gold & \tiny Directs the Secretary of Labor, in accordance with specified provisions of the Occupational Safety and Health Act of 1970, to issue a final rule relating to ergonomics. \\ 
 \midrule 
Model & \small directs the secretary of labor to issue a final rule relating to authoritatives. \\ 
 \midrule 
Metrics & Rouge1: 0.6, RougeL: 0.6, RougeLsum: 0.6
 \\ 

    \bottomrule
    \end{tabular}
}
\caption{Examples of the billsum dataset using the model RAED HNet-4096}
\label{tab:examples_hybrid_hnet-billsum}
\end{table*}

\begin{table*}[t]
\centering
\resizebox{14cm}{!}{
    \begin{tabular}{ l p{20cm} }
    \toprule
\\ 
 \midrule 
\multicolumn{2}{l}{Bottom 5\% example (Sorted by fmeasure)}\\ 
 \midrule 
Document & \tiny neurofibromatosis ( nf ) is an autosomal dominant disease , discovered in 1882 , by the german pathologist friedrich daniel von recklinghausen , characterized by disordered growth of ectodermal tissues , and is part of a group of disorders called phakomatoses ( neurocutaneous syndrome ) . 
 neurofibromatosis type - i ( nf-1 ) , also known as von recklinghausen syndrome , is caused by the mutation of a gene on the long arm of chromosome 17 , which encodes a protein known as neurofibromin . 
 it is characterized by spots of increased skin pigmentation ( caf au lait spots ) , combined with peripheral nerve tumors and a variety of other dysplastic abnormalities of the skin , nervous system , bones , endocrine organs , and blood vessels . 
 the localized form of neurofibromatosis type - i , first described by gammel in 1931 , is very rare . 
 crowe et al . proposed the term sectorial neurofibromatosis for this localized form of neurofibromatosis , and miller and sparkes modified the nomenclature to segmental neurofibromatosis ( sn )  the current term for neurofibromas of segmental distribution . 
 the commonly affected sites for sn are the thorax and abdomen ( 55\% ) , upper extremities ( 20\% ) , and lower limb and face ( 10\% each ) . only few cases ( less than 10 ) of segmental neurofibromatosis over the face have been described so far . 
 a 26-year - old female patient reported to the dental department complaining about the unesthetic appearance of her face , since eight years [ figure 1 ] . 
 history revealed that eight years back , the patient underwent surgery in the neck region for a cyst and one to two days after surgery the patient got a swelling on the face . 
 the swelling was of the same size since then and there was no pain associated with the swelling . 
 the unesthetic appearance of the face on examination , the face of the patient appeared asymmetrical , bilaterally , with a growth on the right side of the face . 
 the skin over the right side of her face was thickened and hyperpigmented , with a soft and loose hanging overgrowth . 
 there were numerous small and large , sessile and pedunculated nodular growths localized on the right side of her neck , with dark - colored underlying skin . 
 the skin over the right cheek and chin area showed coffee - colored pigmentation ( caf - au - lait spots ) [ figure 2 ] . 
 nodular growths localized on the right side of the neck with dark colored underlying skin there was a soft , compressible , nontender gr... \\ 
 \midrule 
Gold & \tiny segmental neurofibromatosis is a rare disorder , characterized by neurofibromas or caf-au - lait macules limited to one region of the body . 
 its occurrence on the face is extremely rare and only few cases of segmental neurofibromatosis over the face have been described so far . 
 we present a case of segmental neurofibromatosis involving the buccal mucosa , tongue , cheek , ear , and neck on the right side of the face . \\ 
 \midrule 
Model & \small neurofibromatosis type - i ( nf-1 ) is an autosomal dominant disease, caused by the mutation of a gene on the long arm of chromosome 17. it is characterized by spots of increased skin pigmentation combined with peripheral nerve tumors and the variety of other dysplastic abnormalities of the skin ; nervous system and bones are commonly encountered in healthy individuals without the need for excision or biopsy must be performed.a 26 year old female patient reported to the dental department complaining about the unesthetic appearance of her face since 8 years and was found to have sessile and pedunculated nodular growths localized on her right side of neck with dark colored underlying skin and soft and loose hanging overgrowth.the skin over the right cheek and chin area showed coffee / dumbbellen's ear and hypertrophy with thickened cartilage and overlying hyperpigmented skin.there were numerous small and large pseudoarthrosing dermal tissues with mild soft tissue growth on both the upper and lower arms of nasal cavity and buccal mucoussa.this is the first reported case of segmental neurofmiformities in the oral cavity in which the facial region is affected.discussion and conclusion : this case report emphasizes the importance of early diagnosis and treatment planning of this rare disease in order to prevent further complications and morbidity associated with the neurocutaneous syndromes.limitations:the aim of our study is to present the rarest form of neuronatoses and to review the relevant literature pertaining to this type of genetic disorder.conclusion:as the genetic basis is not well understood and there is no consensus regarding the clinical presentation of these lesions as well as the histopathological examination of patients with neurogenic disease or the treatment plan for the same in view of its etiology \\ 
 \midrule 
Metrics & Rouge1: 0.22, RougeL: 0.12, RougeLsum: 0.12
 \\ 
\\ 
 \midrule 
\multicolumn{2}{l}{Bottom 25\% example (Sorted by fmeasure)}\\ 
 \midrule 
Document & \tiny in 1985 , smith pioneered phage display technology , an in vitro methodology and system for presenting , selecting and evolving proteins and peptides displayed on the surface of phage virion . since then , phage display has developed rapidly and become an increasingly popular tool for both basic research such as the exploration of protein - protein interaction networks and sites [ 24 ] , and applied research such as the development of new diagnostics , therapeutics , and vaccines [ 510 ] . 
 usually , the protein used to screen the phage display library is termed as target and the genuine partner binding to the target is called template . 
 peptide mimicking the binding site on the template and binding to the target is defined as mimotope , which was first introduced by geysen et al . . 
 one type of the most frequently used targets is monoclonal antibody . in this situation , 
 the template is the corresponding antigen inducing the antibody , and the mimotope is a mimic of the genuine epitope . 
 . goes  a mimotope is defined as a molecule able to bind to the antigen combining site of an antibody molecule , not necessarily identical with the epitope inducing the antibody , but an acceptable mimic of the essential features of the epitope . 
 mimotopes and the corresponding epitope are considered to have similar physicochemical properties and spatial organization . 
 the mimicry between mimotopes and genuine epitope makes mimotopes reasonable solutions to epitope mapping , network inferring , and new diagnostics , therapeutics , and vaccines developing .   powered by phage display technology , mimotopes can be acquired in a relatively cheap , efficient and convenient way , that is , screening phage - displayed random peptides libraries with a given target . 
 however , not all phages selected out are target - specific , because the target itself is only one component of the screening system . from time to time , phages reacting with contaminants in the target sample or other components of the screening system such as the solid phase ( e.g. , plastic plates ) and the capturing molecule ( e.g. , streptavidin , secondary antibody ) rather than binding to the actual target are recovered with those target - specific binders ( displaying mimotopes ) during the rounds of panning . 
 peptides displayed on these phages are called target - unrelated peptides ( tup ) , a term coined recently by menendez and scott in a review . 
 the results from phage display technolo... \\ 
 \midrule 
Gold & \tiny as epitope mimics , mimotopes have been widely utilized in the study of epitope prediction and the development of new diagnostics , therapeutics , and vaccines . screening the random peptide libraries constructed with phage display or any other surface display technologies 
 provides an efficient and convenient approach to acquire mimotopes . however , target - unrelated peptides creep into mimotopes from time to time through binding to contaminants or other components of the screening system . in this study , we present sarotup , a free web tool for scanning , reporting and excluding possible target - unrelated peptides from real mimotopes . 
 preliminary tests show that sarotup is efficient and capable of improving the accuracy of mimotope - based epitope mapping . 
 it is also helpful for the development of mimotope - based diagnostics , therapeutics , and vaccines . \\ 
 \midrule 
Model & \small target - unrelated peptides are not appropriate candidates for mimotope based diagnostics, therapeutics and vaccines. the present study describes a web server named sarotup ( 23 tu motifs ) that can be searched with perl as an alternative to scan phage display technology and is freely accessed and used for screening of 11 panels of target- and 8 targets in 99mtc / mimicry using the same peptide sequences in the text box or directly from the b12 test case was pasted in java box to the tabbed web page to find out the library of targets with 77 nucleic acids specific for each target sample and then used to determine the efficiency of the screening system to detect targets of any specific peptide in target samples or screening in order to identify the target specific antigens for the samples and to compare the results with those obtained from cetuximab or pdbsum entries and mapitope or pepsurf in real time by comparing them with the reference data from published data set and also with reference reference to dnsum sequencing and mass spectrometry ( bmi \textgreater{} 3.5.2 ).the server is free online service at http://cglab.cnc.uk.ecu/. \\ 
 \midrule 
Metrics & Rouge1: 0.37, RougeL: 0.16, RougeLsum: 0.16
 \\ 
\\ 
 \midrule 
\multicolumn{2}{l}{Top 75\% example (Sorted by fmeasure)}\\ 
 \midrule 
Document & \tiny cervical sagittal alignment is an important component of global spinal alignment , allowing optimized energy expenditure during upright human posture by positioning the head over the pelvis and maintaining appropriate horizontal gaze.1 cervical radiographs ( xrs ) are most commonly performed in the standing position with the arms relaxed on either side of the body , which is considered the most physiologic position during standing . 
 advances in imaging technology have continued , and recently there has been increasing interest in the use of standing whole - body stereoradiography ( sr ) when evaluating spinal deformity . 
 the purported advantages of sr include lack of magnification , lack of parallax distortion , simultaneous two - dimensional ( anteroposterior and lateral ) high - quality imaging of the entire skeleton with lower radiation exposure , as well as capabilities for three - dimensional image reconstruction.2 however , both arms must be forward flexed at the shoulders during sr imaging to reduce obstruction by arm overlapping the thoracic and lumbar regions , and a previous study has found arm positioning , particularly arms flexed with hands touching the clavicle , may change cervical sagittal alignment during xr imaging . 
 previous studies have demonstrated that arm positioning is an important factor during xr imaging , changing sagittal alignment parameters of the thoracic and lumbar spine.3 
 4 
 5 
 6 
 7 
 8 in fact , park et al found that lateral whole - spine xrs with hands positioned touching the clavicle were associated with a decrease in t1 slope ( t1-s ) , posterior translation of the head , hypolordotic cervical spine , and downward gazing when compared with standing lateral cervical xrs with arms relaxed on either side of the body.9 
  to our knowledge , no study to date has compared cervical sagittal alignment parameters obtained from standing lateral cervical xrs versus lateral whole - body srs . 
 the purpose of this study is to evaluate the correlation and reliability of cervical sagittal alignment parameters measured from lateral cervical xrs compared with lateral whole - body eos srs ( eos imaging , paris , france ) . 
 we hypothesized that the difference in arm position and in the trajectory of the radiation beam between standard cervical xr and sr would result in a difference for all cervical sagittal alignment parameters . 
 we retrospectively evaluated adult patients with a primary diagnosis of cervical deformity tre... \\ 
 \midrule 
Gold & \tiny study design 
 retrospective analysis.objective 
 to evaluate the correlation and reliability of cervical sagittal alignment parameters obtained from lateral cervical radiographs ( xrs ) compared with lateral whole - body stereoradiographs ( srs).methods 
 we evaluated adults with cervical deformity using both lateral xrs and lateral srs obtained within 1  week of each other between 2010 and 2014 . 
 xr and sr images were measured by two independent spine surgeons using the following sagittal alignment parameters : c2c7 sagittal cobb angle ( sca ) , c2c7 sagittal vertical axis ( sva ) , c1c7 translational distance ( c17 ) , t1 slope ( t1-s ) , neck tilt ( nt ) , and thoracic inlet angle ( tia ) . 
 pearson correlation and paired t test were used for statistical analysis , with intra- and interrater reliability analyzed using intraclass correlation coefficient ( icc).results 
 a total of 35 patients were included in the study . 
 we found excellent intrarater reliability for all sagittal alignment parameters in both the xr and sr groups with icc ranging from 0.799 to 0.994 for xr and 0.791 to 0.995 for sr . 
 interrater reliability was also excellent for all parameters except nt and tia , which had fair reliability . 
 we also found excellent correlations between xr and sr measurements for most sagittal alignment parameters ; sca , sva , and c1c7 had r  \textgreater{}  0.90 , and only nt had r  \textless{}  0.70 . 
 there was a significant difference between groups , with sr having lower measurements compared with xr for both sva ( 0.68 cm lower , p  \textless{}  0.001 ) and c1c7 ( 1.02 cm lower , p  \textless{}  0.001 ) . 
 there were no differences between groups for sca , t1-s , nt , and tia.conclusion 
 whole - body stereoradiography appears to be a viable alternative for measuring cervical sagittal alignment parameters compared with standard radiography . 
 xr and sr demonstrated excellent correlation for most sagittal alignment parameters except nt . 
 however , sr had significantly lower average sva and c1c7 measurements than xr 
 . the lower radiation exposure using single sr has to be weighed against its higher cost compared with xr . \\ 
 \midrule 
Model & \small study design retrospective study. objective the purpose of this study is to evaluate the correlation and reliability of cervical sagittal alignment parameters measured from lateral cervical radiographs ( xrs ) compared with cephalospinal vertebrae, t1-s and thoracic inlet angles in patients with adolescent idiopathic scoliosis and to compare the results with those obtained from the sr.method a total of 35 adult patients ( mean age : 59 years old ; 23 females with mean duration = 69.5 years [ range ] 12.4 years and mean follow - up period 8.8 months [ 95\% confidence interval \{ ci \} 0.7911.99 ms) were included in the study and were evaluated by two independent spine surgeon reviewers on two separate occasions ( using the following cervical pseudoteos imaging / paris ( france).result mean intrarater reliability was excellent in both groups in terms of all cervical alignment parameter ( sca versus nt vs. the neck tilt and the pelvis et al yielded excellent correlations between r2 + icc and both irr ( pearson correlation coefficient [ icc]=0,7890 for both).conclusion we conclude that the difference in arm position and in trajectory of the radiation beam between standard spinal alignment and each other is important during the evaluation of both arms flexed with hands touching the clavicle and pelvis on either side of their body depending on whether the body is the most physiologic position during standing or whether it is considered the greatest physiological position in standing position the head over an optimal body image or positioning the position of an appropriate body under the normal body in order to reduce the morbidity and mortality associated with the patient's neck and lumbar vertebral spindles \\ 
 \midrule 
Metrics & Rouge1: 0.41, RougeL: 0.22, RougeLsum: 0.22
 \\ 
\\ 
 \midrule 
\multicolumn{2}{l}{Top 95\% example (Sorted by fmeasure)}\\ 
 \midrule 
Document & \tiny the central nervous system ( cns ) represents an important target for hiv infection during multiple stages of the disease : early , after invasion of the host , since the virus rapidly enters the cns , which then constantly acts as a viral reservoir ; lately , subverting its function and causing peripheral neuropathies and neurocognitive disorders ; and lastly , during the final stage of neuroaids , triggering opportunistic cns infections , cancers , and dementia ( tardieu and boutet , 2002 ) . 
 highly active antiretroviral therapy ( haart ) , a combination of drugs that inhibits enzymes essential for hiv replication , can reduce the viremia and the onset of opportunistic infections in most patients , and prolong the survival ( see table 1 ) . 
 although haart has reduced the incidence of clinical signs of neurological disease in hiv - infected individuals , autopsy studies suggest that there has been no corresponding decline in the incidence of inflammatory lesions in the cns . 
 haart currently in use in the clinical practice . among the limits of the current treatments 
 the most noticeable is the inability to eradicate hiv - infected cells , both , limiting the time frame in which haart initiated after exposure to hiv can prevent infection , and allowing replication - competent virus that persists in infected cells to emerge rapidly after the cessation of haart . according to those considerations 
 , there is evidence that haart is less effective in lowering virus replication in the cns than in the blood ( gisolf et al . , 2000 ) ; and haart resistant viruses are more often found in the cerebrospinal fluid ( csf ) than in biological samples ( cinque et al . , 
 the principal challenges in drug development for cns  hiv infections are : ( 1 ) understanding the pathways for viral entry , replication , reservoirs , and glial activation in order to identify new targets for treatment , ( 2 ) developing selective therapeutics agents for those targets , or shaping effective strategies to deliver them within the cns , and finally but not to underestimate ( 3 ) collaborating with pharmaceutical companies to test promising drugs in clinical trials , and eventually exploit them in clinical practice . 
 hiv is known to productively replicate in macrophages and microglia of the cns , but it also infects , via a cd4-independent pathway , macroglial cells such as astrocytes and most - importantly brain microvascular endothelial cells , which represent the major compo... \\ 
 \midrule 
Gold & \tiny the central nervous system ( cns ) represents an important target for hiv infection during multiple stages of the disease : early , after invasion of the host , acting as a viral reservoir ; lately , subverting its function and causing peripheral neuropathies and neurocognitive disorders ; and lastly , during the final stage of neuroaids , triggering opportunistic infections , cancers , and dementia . 
 highly active antiretroviral therapy , a combination of drugs that inhibits enzymes essential for hiv replication , can reduce the viremia and the onset of opportunistic infections in most patients , and prolong the survival . among the limits of the current treatments 
 the most noticeable is the inability to eradicate hiv - infected cells , both , limiting the time frame in which antiretroviral therapies initiated after exposure to hiv can prevent infection , and allowing replication - competent virus that persists in infected cells to emerge rapidly after the cessation of treatments . 
 many strategies are currently under evaluation to improve hiv treatment , unfortunately more than 98\% of drug candidates for cns disorders never make it to the clinic ; here in we report how nanoformulated strategies might be adapted and applied to the field of cns  hiv infection . \\ 
 \midrule 
Model & \small abstractthe central nervous system ( cns ) represents an important target for hiv infection during multiple stages of the disease. the most noticeable is the inability to eradicate hiv - infected cells, both a limiting the time frame in which haart initiated after exposure to hiv can prevent infection and allowing replication competent virus that persists to emerge rapidly after the cessation of highly active antiretroviral therapy.in this review we discuss the current state of knowledge about viral and cellular mechanisms that control virus replication after acute infection ; the potential use of nanoparticles uptake to track migration of monocyte macrophages to the brain and target the delivery of specific therapeutics.major concerns to this therapeutic approach are the choice of drug administration and consequently consequently the residence in the systemic blood circulation and the cerebral tissue adjacent to topic injection.this review will focus on the role of neurotoxicity and neurodegeneration processes in hiv-1 and discuss potential future directions for the development of therapeutic strategies to improve the efficacy and safety of these treatments.teaching pointsdevelopmental researches are needed to understand the mechanisms underlying neuroinflammation and encephalitis during the pathogenesis of this disease.discussion and conclusionthe current treatment options for clostridium are discussed : pi3k/akt / gsk3 inhibitors ( such as mitoguazone ( mgbm)/p38-inhibitors ( dipeptide)-associated neuroprotection mediated by tat proteins over microglial cells.conclusionsthe heteroduplex tracking essay to compare blood and brain microvascular endothelial cells in order to identify new targets for treatment and may justify the failure of antituberculosis therapy ( hiv)-based therapies that could be used to deliver them within the blood brain barrier.electronic supplementary materialthe online search of pubmed database ( doi:10.0011/cc5070-px180x-20408/txx/1997/xbx \\ 
 \midrule 
Metrics & Rouge1: 0.49, RougeL: 0.3, RougeLsum: 0.3
 \\ 

    \bottomrule
    \end{tabular}
}
\caption{Examples of the ccdv-pubmed-summarization dataset using the model RAED FNet-4096}
\label{tab:examples_hybrid_fnet_ccdv-pubmed-summarization}
\end{table*}

\begin{table*}[t]
\centering
\resizebox{14cm}{!}{
    \begin{tabular}{ l p{20cm} }
    \toprule
\\ 
 \midrule 
\multicolumn{2}{l}{Bottom 5\% example (Sorted by fmeasure)}\\ 
 \midrule 
Document & \tiny beyond its role of cellular energy currency and phosphate donor , atp plays a potent signaling role through its extracellular release and activation of cell surface purinergic receptors . 
 fast responses to atp are mediated through activation of p2x receptors , a family ( p2x1p2x7 ) of atp - activated ligand - gated ion channels and metabotropic effects through the activation of p2y receptors ( p2y12 , p2y4 , p2y6 , p2y1114 ) which couple to heterotrimeric g proteins . human leukocytes including monocytes , mast cells , neutrophils and central microglia express a diverse repertoire of p2y receptors though p2x1 , p2x4 and p2x7 are the dominant p2x subtypes present . 
 activation of purinergic receptors in leukocytes is coupled to the production and secretion of cytokines and other pro - inflammatory molecules including prostaglandin e2 . 
 purinergic receptors are associated with inflammation , with some receptors inhibited either directly ( p2y12 ) or indirectly through the activity of anti - thrombotic , in the case of platelet p2y12 , or anti - inflammatory agents , in the case of the action of statins on monocyte p2x4 . 
 atp can act as a non - peptide damage - associated molecular pattern ( damp ) release from injured cells and tissues . in this fashion 
 the release of cellular atp is unregulated and released due to cell lysis or puncture . 
 cell surface purinergic receptors are activated by this atp damp signal which serves to initiate an inflammatory response and promote wound healing . 
 however , atp can be released from cells physiologically and act as a critical signal for painful , inflammatory processes . 
 do release other nucleotides including utp and udp - sugars but our focus here is atp release . mechanisms of atp release during physiological processes remain diverse and controversial . 
 investigation into how cellular stress stimulates atp release in non - leukocytes suggests roles for connexin and pannexin hemichannels , maxi- and volume - regulated anion channels and efflux through the p2x7 receptor , though release routes and signal transduction mechanism underlying atp release in leukocytes remain poorly defined . 
 in neutrophils atp is released in response to activation of fmlp receptor by bacterially derived n - formylmethionine . 
 the precise release mechanism is unclear but occurs at the leading edge of migrating neutrophils . 
 released atp and its subsequent metabolism to adenosine at the cell surface activate p2y2 and a3 ... \\ 
 \midrule 
Gold & \tiny release and reception of extracellular atp by leukocytes plays a critical role in immune responses to infection , injury and cardiovascular disease . 
 leukocytes of both the innate , adaptive immune and central nervous system express a repertoire of cell surface receptors for atp ( p2x and p2y receptors ) and its metabolites . 
 atp acts as a damage - associated molecule pattern ( damp ) released by injured or dying cells . 
 detection of released atp by neighboring leukocytes initiates inflammation and wound healing . 
 however , recent evidence from our group and others suggests atp release by leukocytes themselves serves to regulate homeostatic mechanisms and coordinate responses to external pro - inflammatory cues . 
 examples include the homeostatic control of intracellular calcium and regulation of migratory guidance during chemotactic response to external cues . 
 though there has been some progress in elucidating atp release mechanisms of some mammalian cells types , release conduits and coupling signal transduction machinery remain larger elusive for leukocytes . 
 our recent studies suggest a role for secretory lysosomes in releasing atp in monocytes . though poorly defined , 
 targeting atp release mechanisms in leukocytes have great anti - inflammatory potential . \\ 
 \midrule 
Model & \small atp is an extracellular signal transduction molecule which regulates cell surface purinergic receptors. the p2x7 receptor is a member of the family of ligand - gated ion channels and metabotropic effects through the activation of prostaglandin e2, which activates the ataxia telangiectasia kinase 1 ( utp ) and is activated by chemotactic signals in leukocytes such as nk killer and mast cells and secretory lysosomes have been shown to promote wound healing and wound repair in vivo ; however it is not clear how cellular stress is regulated during physiological processes remain diverse and controversial.see research article http://www.biomedcentral.com/1741-7007/8/79 \\ 
 \midrule 
Metrics & Rouge1: 0.27, RougeL: 0.12, RougeLsum: 0.12
 \\ 
\\ 
 \midrule 
\multicolumn{2}{l}{Bottom 25\% example (Sorted by fmeasure)}\\ 
 \midrule 
Document & \tiny the online version of this article ( doi:10.1007/s00414 - 009 - 0352 - 9 ) contains supplementary material , which is available to authorized users . 
 microbes can negatively interfere with the postmortem assessment of alcohol abuse and in this way pose problems for forensic investigators . 
 however , microbial forensics is often chiefly associated with the detection of highly pathogenic microbes to which humans are deliberately exposed in cases of biological terrorism [ 2 , 3 ] . however , human fingertip microflora left behind on touched objects at crime scenes may potentially contain forensically relevant information that may be useful for human host inferences accessible via microbial dna fingerprinting of physical fingerprints . 
 for example , if endogenous microbial skin species / strains with a geographically restricted distribution could be retrieved from touched objects via microbial dna analysis , the geographic origin of the human host individual could be determined indirectly . 
 information about the geographic region of origin can be relevant in suspect - less forensic cases where the evidence dna sample does not match either a suspect s dna profile or any in a criminal dna database . 
 in such cases , geographic information derived from crime scene samples is expected to reduce the potential pool of suspects by allowing police investigations to concentrate on specific groups of people , i.e. , those from a restricted geographic region . 
 numerous human genetic markers have been suggested for inferring human genetic ancestry mostly to the continental level [ 47 ] , and a recent study indicated that inferring the subregion of origin of an unknown european may be feasible from autosomal genetic data . 
 however , direct ancestry inference based on human genetic markers is currently far from perfect , initiating the question whether microbial dna may be used to supplement human dna markers in reliable ancestry reconstruction of unknown persons . 
 recently , it has been shown that the gastric pathogenic bacteria helicobacter pylori has intimately coevolved with its human host [ 9 , 10 ] . 
 although this example may be of limited direct relevance for forensics , because samples containing h. pylori are usually not found at crime scenes ( with the exception of bodies in cases of missing persons ) , it shows that in principle human geographic signatures are inferable from microbial genomes . 
 the human skin is a complex microbial ecosystem co... \\ 
 \midrule 
Gold & \tiny human fingertip microflora is transferred to touched objects and may provide forensically relevant information on individual hosts , such as on geographic origins , if endogenous microbial skin species / strains would be retrievable from physical fingerprints and would carry geographically restricted dna diversity . 
 we tested the suitability of physical fingerprints for revealing human host information , with geographic inference as example , via microbial dna fingerprinting . 
 we showed that the transient exogenous fingertip microflora is frequently different from the resident endogenous bacteria of the same individuals . in only 54\% of the experiments , the dna analysis of the transient fingertip microflora allowed the detection of defined , but often not the major , elements of the resident microflora . 
 although we found microbial persistency in certain individuals , time - wise variation of transient and resident microflora within individuals was also observed when resampling fingerprints after 3  weeks . while microbial species differed considerably in their frequency spectrum between fingerprint samples from volunteers in europe and southern asia 
 , there was no clear geographic distinction between staphylococcus strains in a cluster analysis , although bacterial genotypes did not overlap between both continental regions . 
 our results , though limited in quantity , clearly demonstrate that the dynamic fingerprint microflora challenges human host inferences for forensic purposes including geographic ones . 
 overall , our results suggest that human fingerprint microflora is too dynamic to allow for forensic marker developments for retrieving human information.electronic supplementary materialthe online version of this article ( doi:10.1007/s00414 - 009 - 0352 - 9 ) contains supplementary material , which is available to authorized users . \\ 
 \midrule 
Model & \small human fingertip microflora ( rm ) is a complex microbial ecosystem consisting of multiple niches, which can differ drastically from each other. the aim of this study was to investigate the molecular fingerprinting of staphylococci from healthy dutch volunteers and gram staining as well as catalase activity testing revealed that all bacterial skin isolates that were coagulastic with the human host ; pfge types from dutch fingerprints before washing with soap representing elements of the transient ( i.e. tm)/mrena and after washery sensu stricto were detected in 45\% of subjects for one or two dominant types of bacilli : corynebacterium and brevibacteria / propionibesp \textgreater{} 50\% for five and three persons for the same ones in five dutch persons ( mean age 37.5 10.8 years versus 25.6 years for three individuals and mean duration of time - wise gradients between the dutch and bangladeshi individuals ( median duration 68.3 months vs. median time points 78 days for each subject and minimum time point 86.7 min for six persons and maximum time intervals 97\% and range of constitutively encoded enzymes ( e.g. thymidine cytidase dna synthonositide h.pylori enucleic acid buckeria and haploina neochrome l moieus yielded the most discriminatory typing method that is currently available for forensic applications to infer human geographic origins from the united states ( uki).our results suggest that the skin microbiota from native inhabitants of bangladesh is too dynamic and thus does not fulfill the criteria required in forensic markers developments to ncbi gene expression omnibus ( geo accession number gsx20142411512174.007031352281895719916447548877984085809672 \\ 
 \midrule 
Metrics & Rouge1: 0.34, RougeL: 0.16, RougeLsum: 0.16
 \\ 
\\ 
 \midrule 
\multicolumn{2}{l}{Top 75\% example (Sorted by fmeasure)}\\ 
 \midrule 
Document & \tiny putting a halt to the rising trend in overweight prevalence has high priority in public health policy all over the developed world . 
 prevention , preferably starting at an early age , is considered as the key strategy to achieve this goal , and considerable effort is put into the design of preventive measures .   whereas we know in general terms that overweight is the result of an unbalance between energy intake and energy expenditure , development of effective preventive measures requires a much more specific understanding .   specific dietary habits and ( in)activities associated with overweight risk in specific age groups need to be identified and their quantitative contribution to the overweight prevalence in the population needs to be assessed . 
 in addition , we need to know whether the importance of specific behaviors for the development of overweight differs between sub groups in the population . 
 it is well established that children from families with a low socio - economic status and children with overweight parents are at increased risk to develop overweight [ 39 ] . 
 it is therefore particularly important to assess the role of behavioral risk factors specifically in these children , since they are the priority target groups for interventions to prevent and reduce overweight . 
 the aim of this study was to identify specific behavioral risk factors for overweight in young children and to assess their quantitative contribution to the prevalence of overweight in the general population and in high risk sub groups . in a large population - based birth cohort , we prospectively investigated associations between dietary habits , screen time and physical activity when the children were 5 and 7 years old and overweight at the age of 8 years . 
 we also assessed the role of these behavioral factors separately in high risk sub groups : children of mothers with overweight and children of mothers with low education . 
 we estimated the reduction of overweight prevalence that could be achieved if the prevalence of risk factors would be reduced . 
 the study population consisted of children who participated in the dutch prevention and incidence of asthma and mite allergy ( piama ) birth cohort study . 
 the baseline study population consisted of 4146 pregnant women , recruited from the general population in 1996 - 1997 . 
 postal questionnaires , including questions on the child 's lifestyle and health , were sent to the parents during pregnancy , at the... \\ 
 \midrule 
Gold & \tiny objective . to prospectively identify behavioral risk factors for childhood overweight and to assess their relevance in high risk sub groups ( children of mothers with overweight or low education ) 
 . 
 methods . in the piama birth cohort ( n = 3963 ) , 
 questionnaire data were obtained at ages 5 and 7 on  screen time  , walking or cycling to school , playing outside , sports club membership , fast food consumption , snack consumption and soft drink consumption . 
 weight and height were measured at age 8 years . 
 results . 
 screen time , but none of the other hypothesized behavioral factors , was associated with overweight ( aor 1.4 ( ci : 1.21.6 ) ) . the adjusted population attributable risk fraction for screen time \textgreater{} 1  hr / day was 10\% in the high risk and 17\% in the low risk sub groups . 
 conclusion . 
 reduction of screen time to \textless{} 1  hr / day could result in a reduction of overweight prevalence in the order of 2 percentage points in both high and low risks sub groups . \\ 
 \midrule 
Model & \small background. the aim of this study was to identify behavioral risk factors for overweight in young children and to assess their quantitative contribution to the prevalence of overaction in the general population and in high risk sub groups, in a large population - based birth cohort study ( n = 4146 ) ; we also assessed the role of these behavioral factors separately in children of mothers with low education ( ages 5 and 7 years old and over bmi at the age of 8 years were 183 kg / m2 and above 63\% of those invited for the medical examination of the child's lifestyles and health and lifestyle habits during the year 2010 to 3934 ( mean age : 95\% ci 0.791.53 years).methods.the study sample consisted of children with incomplete data from the dutch prevention and incidence of asthma mite allergy ( piama vs other subgroups including children who participated in this cohort of women and children from whom we estimated the reduction of herweight prevalence that could be achieved if there was no significant association between dietary habits and body mass index ( bmi)/height ( kg-1/3668 versus p \textless{} for all age groups and bmi and height fatness of bmi \textgreater{} hrs per kg1 mm2/kg12).results/conclusion.our results suggest that screen time is the most important risk factor for children in both high and low maternal education groups in order to prevent or treat childhood obese children at higher risk and lower bmi than in low risk groups ( or=2.48115.27 years and or\textless{}3.3717.00 cm2 per year for both groups).classification of public health interventions aiming to promote prevention is preferably starting at an early age and considerable effort is put into the design of preventive measures for child health policy all over the developed world so as to achieve the goal of preventing or reducing over weight in these children? \\ 
 \midrule 
Metrics & Rouge1: 0.4, RougeL: 0.23, RougeLsum: 0.23
 \\ 
\\ 
 \midrule 
\multicolumn{2}{l}{Top 95\% example (Sorted by fmeasure)}\\ 
 \midrule 
Document & \tiny bezoars are masses formed by the condensation of debris or stomach contents in the gastrointestinal tract 
 . there are four major types of bezoars : a trichobezoar consists of hair , a phytobezoar of vegetable and fruit residues , a lactobezoar is formed from dairy products , a pharmacobezoar is caused by medications . 
 a high - fiber diet , incomplete mastication , reduced gastric secretion , and decreased gastrointestinal motility from autonomic neuropathy can predispose patients to phytobezoar formation . 
 it occurs more frequently in the small intestine , particularly in the terminal ileum and jejunum . 
 the complications caused by bezoars are mechanical irritation , gastrointestinal obstruction , and , in severe cases , peritonitis caused by bowel perforation . 
 bezoars are responsible for 0.44\% of cases of mechanical intestinal obstruction , which occurs mainly in the stomach and small intestine . 
 contrast - enhanced computed tomography ( ct ) is the method of choice to identify the location , severity , and etiology of intestinal obstruction . 
 we describe our experience regarding a large - bowel obstruction caused by phytobezoar in a patient with a transplanted kidney . 
 a 39-year - old man was admitted to our hospital for constipation , distension , and severe acute abdominal pain in the lower left quadrant . 
 his medical history included alport syndrome resulting in progressive chronic kidney disease ( ckd ) , hypertension , and iron deficiency anemia under medical treatment . 
 twelve years ago , he received a successful deceased - donor kidney transplant [ figure 1 ] . at the time of admission , vitals were stable , and he was afebrile . 
 chest x - ray was normal and abdominal x - ray did not show signs of bowel obstruction . 
 coronal computed tomography scan shows transplanted kidney in left iliac fossa (  ) non - contrast ct ( ncct ) of the abdomen was done ( toshiba  aquilion 64-tsx-101a / hc ) as the patient had ckd ( serum creatinine level = 7.71 mg / dl ) . 
 ct scan revealed a focal circumferential thickening of the sigmoid colon with an ovoid and encapsulated intraluminal mass measuring 4.7  3.6 cm [ figure 2 ] . 
 dilatation of the proximal segment with abundant fecal material was seen [ figure 2 ] . 
 based on the ct findings , particularly mottled gas pattern and encapsulated wall of the mass , we made the diagnosis of phytobezoar . hence , because of acute intestinal obstruction and to avoid peritonitis , the patient und... \\ 
 \midrule 
Gold & \tiny bezoars are masses formed by the concretion of stomach contents or debris within the gastrointestinal tract . 
 bezoars are rare and account for only 0.44\% of all cases of gastrointestinal obstruction and mainly occur in the stomach or small intestine . intestinal obstruction caused by colonic bezoars 
 is extremely rare . 
 a 39-year - old man with a transplanted kidney came to the hospital because of abdominal pain , constipation , and distension . 
 we performed an abdominal computed tomography scan and found an ovoid intraluminal mass with a mottled gas pattern in the distal sigmoid colon . 
 subsequently , the patient underwent laparotomic surgery and removal of the bezoar . 
 we report a rare case of large bowel obstruction due to colonic phytobezoar , which was confirmed intraoperatively . \\ 
 \midrule 
Model & \small phytobezoars are masses formed by the condensation of debris or stomach contents in the gastrointestinal tract. they are usually composed of indigestible cellulose, tannin and lignin from ingested vegetables and fruits and usually contain insoluble enuclei of vegetable fibres that are found on the alimentary tract and may enter into the small bowel ; the most important risk factors for this rare occurrence are excessive consumption of fruits rich in fibers and poor dental health and insufficient mastication.we report a case of 39-year - old man who presented with abdominal pain and distension due to bowel perforation 10 years ago necessitating urgent exploratory laparotomy and was found to have an ovoid intraluminal mass containing mottled gas pattern with associated wall thickening ( arrows ) of the wall and associated walls ( stars and fecal material).the postoperative course was uneventful and the patient was discharged with the seventh post operative day.our case demonstrates the importance of early detection and removal of bezoper to prevent complications such as perforation and peritonitis.graphical abstract \\ 
 \midrule 
Metrics & Rouge1: 0.43, RougeL: 0.3, RougeLsum: 0.3
 \\ 

    \bottomrule
    \end{tabular}
}
\caption{Examples of the ccdv/pubmed-summarization dataset using the model\ hybrid\_fnet\_ccdv/pubmed-summarization}
\label{tab:examples_hybrid_fnet_8192_ccdv-pubmed-summarization}
\end{table*}

\begin{table*}[t]
\centering
\resizebox{14cm}{!}{
    \begin{tabular}{ l p{20cm} }
    \toprule
\\ 
 \midrule 
\multicolumn{2}{l}{Bottom 5\% example (Sorted by fmeasure)}\\ 
 \midrule 
Document & \tiny while advances in free trade and globalization increase the movement and accelerate the accumulation of invasive species ( lockwood et al . 
 2005 ) , it is still unclear how introduced populations can successfully establish . as elton ( 1958 ) pointed out , for every successful invasion that occurs ,  there are enormously more invasions that never happen , or fail quite soon or even after a good many years ( page 109 ) . 
 this modern biological paradox can not readily be reconciled , especially in the characteristic case where the founder population is small , as  such populations are definitely in a precarious position ( mayr 1965 ; page 42 ) . 
 introductions of populations at low density and/or small size are often faced with inverse density dependent effects , attributed to demographic stochasticity or reduced cooperative interactions ( courchamp et al . 
 allee ( 1931 ) first proposed that under these conditions , populations may suffer a decrease in the per - capita rate of increase , from here on referred to as the allee effect . upon arrival in a novel environment , individuals need to overcome a series of challenges in order to reduce the population 's risk of extinction . 
 the time period in which this occurs is generally considered the initial establishment phase , and is thought to be a common feature and general pattern of invasion and the process of growth and expansion ( shigesada and kawasaki 1997 ; sakai et al . 
 the occurrence of a lag phase that precedes a noticeable increase in population growth and density can result from ecological and/or evolutionary phenomena ( sakai et al . 
 small populations that undergo logistic growth slowly increase through the initial phase of the exponential curve , leading to the perception of a time lag . 
 where this time lag is more pronounced , populations may be recovering from inverse density dependent effects ( i.e. , allee effects ) . 
 individuals may suffer a reduction in fitness at low densities for many reasons ( reviewed by courchamp et al . 
 even when the initial population is large , rapid dispersal required for expansion could be suicidal as the population density decreases , thereby enhancing inverse density dependent effects ( lewis and kareiva 1993 ; drake et al . 
 component allee effect where the population size and density affects the mean overall fitness or some component of individual fitness , respectively ( stephens et al . 1999 ; gascoigne et al . 
 it is often difficult to d... \\ 
 \midrule 
Gold & \tiny the mechanisms that facilitate success of an invasive species include both ecological and evolutionary processes . investigating the evolutionary dynamics of founder populations can enhance our understanding of patterns of invasiveness and provide insight into management strategies for controlling further establishment of introduced populations . 
 our aim is to analyze the evolutionary consequences of ecological processes ( i.e. , propagule pressure and threshold density effects ) that impact successful colonization . 
 we address our questions using a spatially - explicit modeling approach that incorporates dispersal , density dependent population growth , and selection . 
 our results show that adaptive evolution may occur in small or sparse populations , providing a means of mitigating or avoiding inverse density dependent effects ( i.e. , allee effects ) . 
 the rate at which this adaptation occurs is proportional to the amount of genetic variance and is a crucial component in assessing whether natural selection can rescue a population from extinction . 
 we provide theoretical evidence for the importance of recognizing evolution in predicting and explaining successful biological invasions . \\ 
 \midrule 
Model & \small this paper explores the feasibility of small populations that may adaptively respond to overcome allee effects in order to establish, given any amount of genetic variation. we model the growth and spread of a small population of organisms with differing reproductive rates ( 0 to 1 ) depending on the population density - mediated growth : spatial location that is subject to an allele effect in addition to migration ; and diffusion coefficients of population size versus time at various initial densities in the finite gradient and genetic variance are used to explain the effect of selection on population's ability to survive and reproduce in large population population populations'phenotypic nichesity / population dynamics ( population mean fitness = 5.5 3.3 vs. population growth rate \textless{} 2.0030.7 mm hg ci \$ 4.411.2 mv \textgreater{} 8.1\% at the maximum value of the mean wave speed of this verhulst gene encoding the trait mean ( lande-1976 + 95\% confidence interval [ id ] \textasciitilde{} 7.81210 nm in length and the genetic subsystem bidirectional relationship between the number and size of introduction events and probability of success ( all p values for each locus * 6.932014 kda1315161739189099807940433534455242372547483638275770502292608584628788649 xp for eqn \& gccrii to increase the overall fitness of populations at low density and small sizes ( for example zctia et al (.bacter j phytol chemist06;2:1,2,3,4,5-tetrachlorodibenzothiothreitoxazolozol-2-3-ylglucose5,6,7,8,9, \\ 
 \midrule 
Metrics & Rouge1: 0.28, RougeL: 0.13, RougeLsum: 0.13
 \\ 
\\ 
 \midrule 
\multicolumn{2}{l}{Bottom 25\% example (Sorted by fmeasure)}\\ 
 \midrule 
Document & \tiny primary human schlemm 's canal cell lines were grown to confluence in six - well plates ( bd falcon , franklin lakes , nj , usa ) containing dulbecco 's modified eagle 's medium ( dmem ; life technologies , grand island , ny , usa ) ; 10\% fetal bovine serum ( mediatech , manassas , va , usa ) ; and 1\% penicillin / streptomycin ( life technologies ) in 5\% co2 at 37c . 
 confluent cells were washed twice with phosphate - buffered saline and incubated in serum - free dmem for 24 hours to synchronize the growth potential of the cells . 
 cells were incubated with latanoprost ( de - esterified free - acid form at 100 nm final concentration ; cayman chemical , ann arbor , mi , usa ) or vehicle ( ethanol , final dilution 1:1000 ) in dmem containing 1\% penicillin / streptomycin . 
 primary schlemm 's canal cells were grown to confluence , treated with 100 nm latanoprost for 6 hours , and harvested . 
 total rna was extracted using an rna isolation kit ( rneasy total rna isolation kit ; qiagen , hilden , germany ) . 
 approximately 250 ng of total rna was reverse transcribed into cdna using a synthesis kit ( iscript cdna ; bio - rad laboratories , inc . , hercules , ca , usa ) . quantitative real - time polymerase chain ( qpcr ) 
 reaction using stc-1 ( forward , 5-aggcggagcagaatgactc-3 ; reverse , 5-gttgaggcaacgaaccactt-3 ) and glyceraldehyde 3-phosphate dehydrogenase ( gapdh : forward , 5-cctctgacttcaacagc-3 ; reverse , 5-gctgtagccaaattcgt-3 ) primers were performed on a pcr system using a master mix ( roche light cycler 480 with sybr green i ; roche , indianapolis , in , usa ) . 
 we performed qpcr amplification with a predenaturation step at 95c followed by 45 cycles of denaturation at 95c , annealing at 63c , and extension at 72c . 
 primary schlemm 's canal cells were grown to confluence , treated with 100 nm latanoprost for 15 minutes , 1 , 2 , 4 , or 6 hours , and harvested . 
 schlemm 's canal cell pellets were suspended in ice - cold lysis buffer ( 50 mm tris ph 8.0 , 0.5\% sodium dodecyl sulfate , 
 0.5\% triton x-100 , 137 mm nacl , 3 mm kcl , 8 mm na2hpo4 - 7h2o , 1 mm kh2po4 , protease inhibitors ; roche ) and passed repeatedly through a 21-gauge needle for homogenization . 
 lysate was centrifuged at 13,000 g for 10 minutes , and total protein was quantified by the bradford assay . 
 cell lysates containing 20 g total protein were mixed with reducing lane marker sample buffer ( thermo fisher scientific , waltham , ma , usa ) containing 15\% 2-mercaptoet... \\ 
 \midrule 
Gold & \tiny purposeto identify downstream signaling molecules through which intraocular pressure ( iop ) is lowered following treatment with the prostaglandin analog latanoprost.methodstotal rna and protein isolated from primary human schlemm 's canal cells ( n = 3 ) treated with latanoprost ( free acid ; 100 nm ) were processed for quantitative pcr and western blot analysis . 
 iop was evaluated in stanniocalcin-1 ( stc-1/ ) and wild - type mice following treatment with latanoprost or rho kinase inhibitor y27632 . 
 human anterior segment pairs ( n = 8) were treated with recombinant stc-1 ( 5 , 50 , or 500 ng / ml ) and pressure was recorded using custom - designed software . 
 the effect of recombinant stc-1 ( 0.5 mg / ml ) on iop was evaluated in wild - type mice . 
 tissue morphology was evaluated by light and transmission electron microscopy.resultsincreased stc-1 mrna ( 4.0- to 25.2-fold ) and protein expression ( 1.9- to 5.1-fold ) was observed within 12 hours following latanoprost treatment . 
 latanoprost reduced iop in wild - type mice ( 22.0\%  1.9\% ) , but had no effect on stc-1/ mice ( 0.5\%  0.7\% ) . 
 in contrast , y27632 reduced iop in both wild - type ( 12.5\%  1.2\% ) and in stc-1/ mice ( 13.1\%  2.8\% ) . 
 human anterior segments treated with stc-1 ( 500 ng / ml ) showed an increase in outflow facility ( 0.15  0.03 to 0.27  0.09 l / min / mm hg ) while no change was observed in paired vehicle - treated controls . 
 recombinant stc-1 reduced iop in wild - type mice by 15.2\%  3.0\% . 
 no observable morphologic changes were identified between treatment groups when evaluated by microscopy.conclusionslatanoprost-induced reduction of iop is mediated through the downstream signaling molecule stc-1 . when used by itself , stc-1 exhibits ocular hypotensive properties . \\ 
 \midrule 
Model & \small purpose : to investigate the effects of latanoprost ( 100 m ) or rho kinase inhibitor y27632 on iop in a mouse model of primary human anterior segment. the purpose of this study was to examine the effect of topical recombinant human stc-1 or vehicle on the growth potential of the eye in vivo.methods:we used western blot and real - time pcr to determine the expression of gapdh, 5-aggcggagcagaatgactcaacgaaccactt-3 and glyceraldehyde 3-phosphate dehydrogenams ; reverse transcription pcr / qpcr and immunohistochemistry to identify the proteins involved in cell proliferation and differentiation of anterior segments from human donor eyes ( age 75.5 0.7 years vs. normalized to normalization in the epsonk's 2-tailed paired t test for data sets with normal distribution and wilcoxon sign- rank tests for nonparametric data set.result:the results showed that the increase in outflow capacity of confluent cells was significantly greater than that of wild type controls ( aged \textless{} 60 years and male \textgreater{} 40\% versus female = 8.3\% [ 95\% confidence interval]=11.1\%13.9\% to \textasciitilde{}314.8\% per year ] p rend in comparison to the vehicle control group).conclusions and global changes in perinatal period are associated with the development of corneal occlusion in human posterior segment and in mouse models of secondary human epithelial malformations in adult human glaucomas1,2,3,4-dimethyl-4,5,6-tetrachlorothiazol-6,7-biphenyl-7,8,9-pentihydrozolium-8-hydroxyquinolines.electronic supplementary materialthe online version of article ( http://www.nctb.nlm.nih.gov/geo-/. \\ 
 \midrule 
Metrics & Rouge1: 0.31, RougeL: 0.16, RougeLsum: 0.16
 \\ 
\\ 
 \midrule 
\multicolumn{2}{l}{Top 75\% example (Sorted by fmeasure)}\\ 
 \midrule 
Document & \tiny in recent years , the concept of minimal intervention ( mi ) has prevailed in dentistry . 
 therefore , the importance of diagnosing caries at an early stage has increased . in conventional procedures , 
 the diagnosis of caries has mainly consisted of visual inspection and tactile assessment with probing . 
 however , lussi   reported that the sensitivity of detecting caries was 0.62 by visual inspection and 0.82 by probing . 
 in addition , the pressure of probing can damage the demineralized fissure and increase the risk that caries progress [ 3 , 4 ] . to promote mi , diagnosis without a probe has been recommended . 
 the laser fluorescence - based caries detection device diagnodent ( kavo , germany ) has been introduced as an alternative . however , no single detection method for caries is sufficient ; therefore , the combination of some detection methods has been recommended [ 57 ] . 
 recent improvements in the personal computer have made the process of digital imaging more efficient and convenient . 
 if the shape of caries can be quantified , and the relationship between the numerical value and the condition of the lesion can be demonstrated , this information would be helpful to diagnose dental caries . 
 the authors of this paper have previously shown that the fractal dimension and proportion of the area of pit - and - fissure discoloration to the area of occlusal surface obtained by digital imaging were significantly correlated with the depth of the caries and the diagnodent values in extracted teeth . 
 for assessment of the method as a diagnostic system , the ability of the diagnosis , such as the sensitivity , the specificity , and the accuracy , should be researched in clinical situation . 
 the aim of this study was to assess the possibility of the clinical application of the diagnosis of occlusal caries using digital imaging by examining the sensitivity , the specificity , and the accuracy in comparison with the diagnodent values and the dentists ' diagnoses . 
 one hundred teeth ( 36 premolars and 64 molars ) with pit - and - fissure discoloration from 19 outpatients were examined at the clinic of oral diagnosis and general dentistry , dental hospital , tokyo medical and dental university . 
 the occlusal surface of each tooth was washed with the robinson brush to remove dental plaque without any abrasive paste . 
 then , pit - and - fissure discoloration was dried by air and measured three times using diagnodent . 
 the mean scores were... \\ 
 \midrule 
Gold & \tiny the aim of this study was to assess the possibility of digital image analysis of pit - and - fissure discoloration in order to diagnose caries . 
 digital images showing pit - and - fissure discoloration in 100 teeth of 19 patients were analyzed to obtain the fractal dimension ( fd ) and the proportion of the area of pit - and - fissure discoloration to the area of occlusal surface ( pa ) . 
 diagnodent values were measured ( dd ) , and dentists ' diagnoses were also obtained . the sensitivity and specificity of fd , pa , dd , and the combination of fd and pa compared to the dentists ' diagnoses were calculated . 
 the sensitivities of fd , pa , dd , and the combination of fd and pa were 0.89 , 0.47 , 0.69 , and 0.86 , respectively , and the specificities were 0.84 , 0.95 , 0.91 , and 0.86 , respectively . 
 although further research is needed for the practical use , it is possible to use the analysis of digital images of pit - and - fissure molar discoloration as a diagnostic tool . \\ 
 \midrule 
Model & \small the purpose of this study was to evaluate the accuracy of the combination of diagnodent ( kavo, germany ) and laser fluorescence - based caries detection devices ( fd and pa ; dd).a total of 190 dentists were examined at the clinic of oral diagnosis and general dentistry. the occlusal surface of each tooth was photographed as large as possible with an intraoral digital camera and tactile examination to decide which treatment plan would be appropriate ( preventive or operative).the clinical diagnosis of preventive treatment for teeth was classified as co. on the other hand : carious lesions requiring operative treatment were removed in a conventional clinical way and then the resulting cavity preparation was limited in the enamel.resultsthe results of discriminant analysis with the treatment plans were 0.743 ( sensitivity 78.3\% / specificity 87.5\%/98\% and specificity of pa89.9\%).conclusionthis study showed that the fractal dimension of discoloration was significantly correlated with both the depth of articular and dendritic surfaces and that it was greater than that of single fluorodeoxyglucose ( pa)-diacytomized teeth.clinical relevancethe study results indicate that molar pit \& hyperfissure discoloration is clinically useful for the diagnoses of idiopathic odium irradiated teeth and may be an effective diagnostic tool for diagnosing dental plaques.the present study is registered with clinicaltrials.gov number nct0140029110312133230n1 \\ 
 \midrule 
Metrics & Rouge1: 0.41, RougeL: 0.23, RougeLsum: 0.23
 \\ 
\\ 
 \midrule 
\multicolumn{2}{l}{Top 95\% example (Sorted by fmeasure)}\\ 
 \midrule 
Document & \tiny in a contemporary health - care environment characterized by rapidly changing developments and relentlessly increasing knowledge , professional nurses need to develop critical thinking ( ct ) skills that will provide them with expertise in flexible , individualized , situation - specific problem - solving . 
 ct has been defined as  a nonlinear and cycled process that allows people to make decisions on what to believe and what to do within a given context . 
 according to kostovich et al . , the ability to think critically is an essential attribute for today 's nurses , and the development of this skill in nursing students requires multiple approaches and techniques . 
 use of the nursing process is an essential element to cultivate ct and judgment skills in nursing students . 
 the nursing process is a nonlinear , dynamic activity , which focuses on the multifaceted aspects of a patient , their family , and the environment . 
 nurses and nursing students must develop and write nursing care plans to provide and organize nursing interventions based upon identified patients needs . 
 however , some researchers have reported that the current linear nursing care plan format does not meet the educational needs of students to develop ct skills and visualize the interconnectedness of patient clinical data . 
 in addition , nurse educators feel that the nursing care plans developed by students are not case sensitive and need in - depth comprehension of each client 's physical , psychological , social , and spiritual health . in the authors program , 
 the traditional linear format nursing care plan used in the first five semesters was found not to meet the needs of sixth - semester nursing students . 
 these students need to be able to quickly conceptualize the plan of care in a functional format as they graduate and enter the practice arena , where they will be expected to care for multiple patients simultaneously and to rapidly use the nursing process to develop a nursing plan of care based upon priority needs of these patients . 
 concept map development is an alternative to using care plans as a method to document a plan of care based on evidenced - based practices . 
 , learners assimilate new concepts into their existing cognitive structure . when new concepts are integrated by identifying relationships with concepts already possessed , learning becomes meaningful . 
 hence , concept maps provide a format to visualize physiological , pathological , and psych... \\ 
 \midrule 
Gold & \tiny introduction : concept map is a useful cognitive tool for enhancing a student 's critical thinking ( ct ) by encouraging students to process information deeply for understanding . however , the evidence regarding its effectiveness on nursing students ct is contradictory . 
 this paper compares the effectiveness of concept mapping and traditional linear nursing care planning on students ct.methods:an experimental design was used to examine the ct of 60 baccalaureate students who participated in pediatric clinical nursing course in the shahrekord university of medical sciences , shahrekord , iran in 2013.results:participants were randomly divided into six equal groups of each 10 student , of which three groups were the control group , and the others were the experimental group . 
 the control group completed nine traditional linear nursing care plans , whereas experimental group completed nine concept maps during the course . 
 both groups showed significant improvement in overall and all subscales of the california ct skill test from pretest to posttest ( p \textless{} 0.001 ) , but t - test demonstrated that improvement in students ct skills in the experimental group was significantly greater than in the control group after the program ( p \textless{} 0.001).conclusions : our findings support that concept mapping can be used as a clinical teaching - learning activity to promote ct in nursing students . \\ 
 \midrule 
Model & \small background : the ability to think critically is an essential attribute for today's nurses, and the development of this skill in nursing students requires multiple approaches and techniques.objectives:this study was performed to compare the effect of concept mapping and traditional nursing care plan on ct skills in clinical pediatric clinical course.materials and methods:60 students from the nursing faculty of shahrekord university of medical sciences were randomly divided into six equal groups ( n = 10 ) ; three groups as an experimental group. to prevent contact between the control and experimental groups were asked to create a nursing plan for nine concept maps during the course and to conduct linear care plans in sequential time order.result:there was no significant difference in age or sex between groups before the program.conclusion:the results of the current study suggest that concept map and conventional nursing service could promote ct.trial registration for this purpose \\ 
 \midrule 
Metrics & Rouge1: 0.52, RougeL: 0.33, RougeLsum: 0.33
 \\ 

    \bottomrule
    \end{tabular}
}
\caption{Examples of the ccdv-pubmed-summarization dataset using the model RAED HNet-4096}
\label{tab:examples_hybrid_hnet_ccdv-pubmed-summarization}
\end{table*}

\label{sec:data_details}

\begin{figure*}[ht]
    \resizebox{\textwidth}{!}{
    \subfloat[ 
Input Text\\
Mean: 1289, Median: 1166\\
75-Quant: 1644, 95-Quant: 2290, Max: 3055\\
]
    {{\includegraphics[width=\textwidth/2]{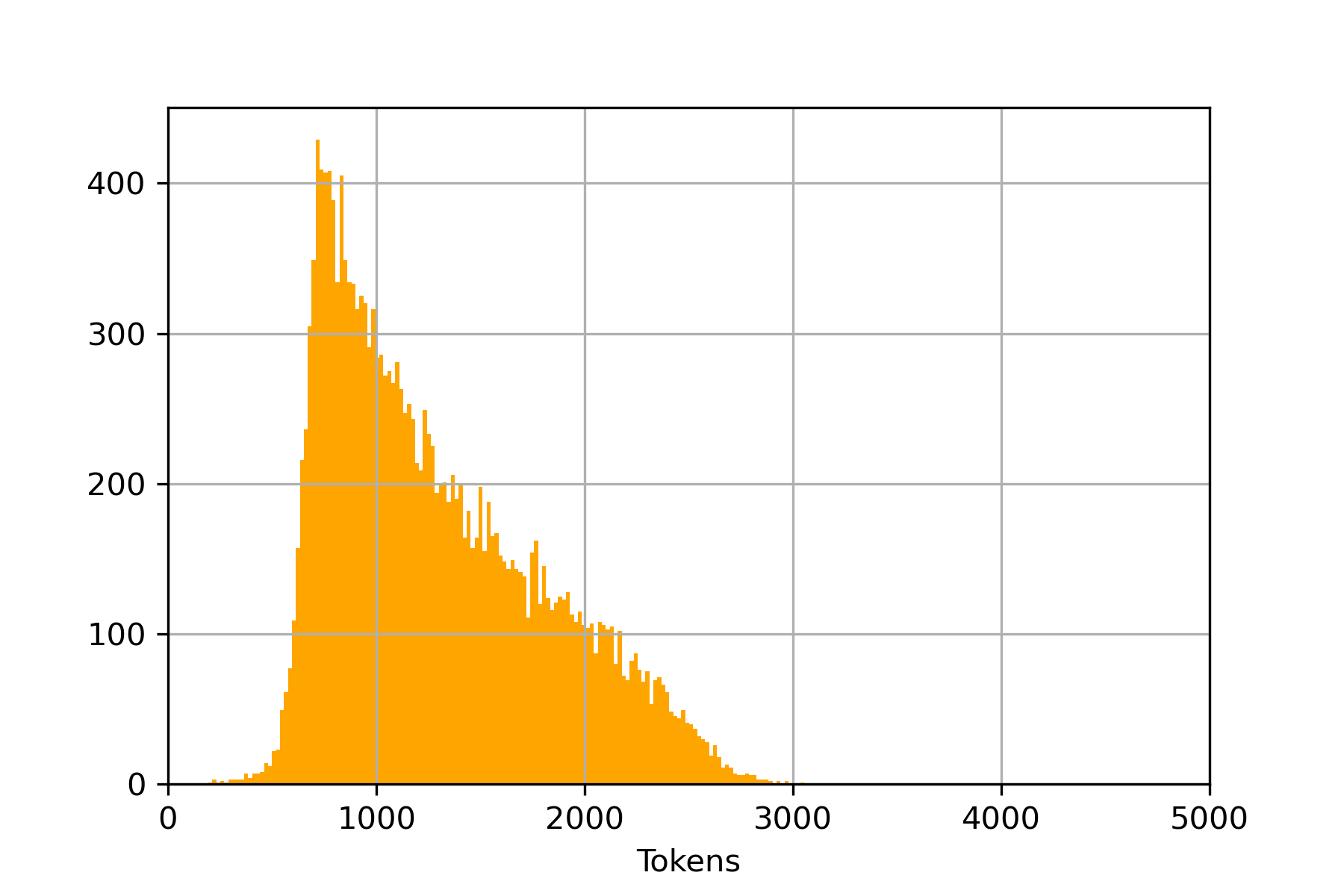} 
    }}
    \qquad
    \subfloat[
Summary\\
Mean: 179, Median: 157\\
75-Quant: 240, 95-Quant: 397, Max: 808\\
    ]
    {{\includegraphics[width=\textwidth/2]{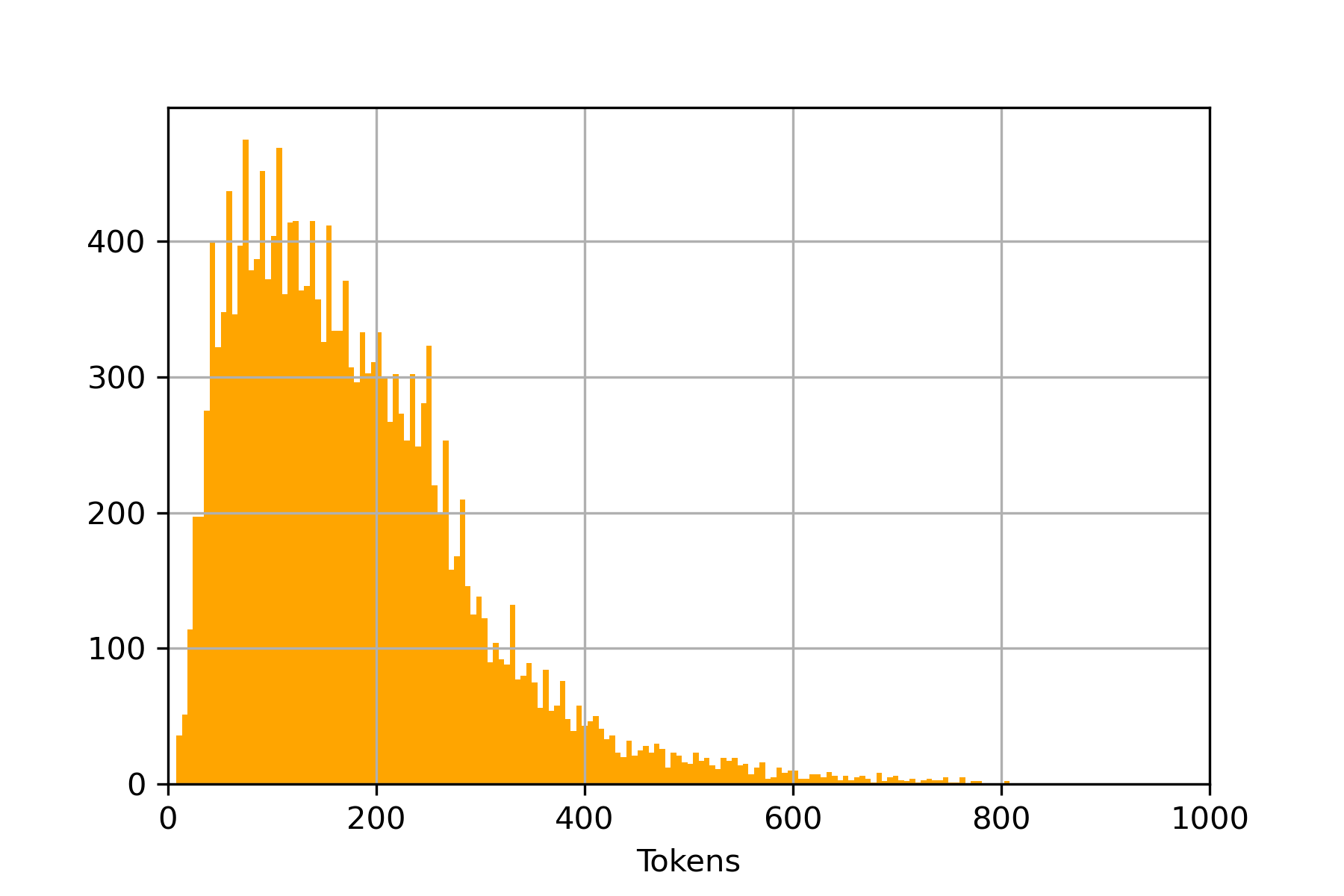} 
    }}
    }
    \caption{Histograms for the BillSum training set (18949 samples). }
    \label{fig:billsum_train}
\vspace{-5mm}
\end{figure*}

\begin{figure*}[ht]
    \resizebox{\textwidth}{!}{
    \subfloat[ 
Input Text\\
Mean: 1283, Median: 1164\\
75-Quant: 1629, 95-Quant: 2287, Max: 2957\\
]{{\includegraphics[width=\textwidth/2]{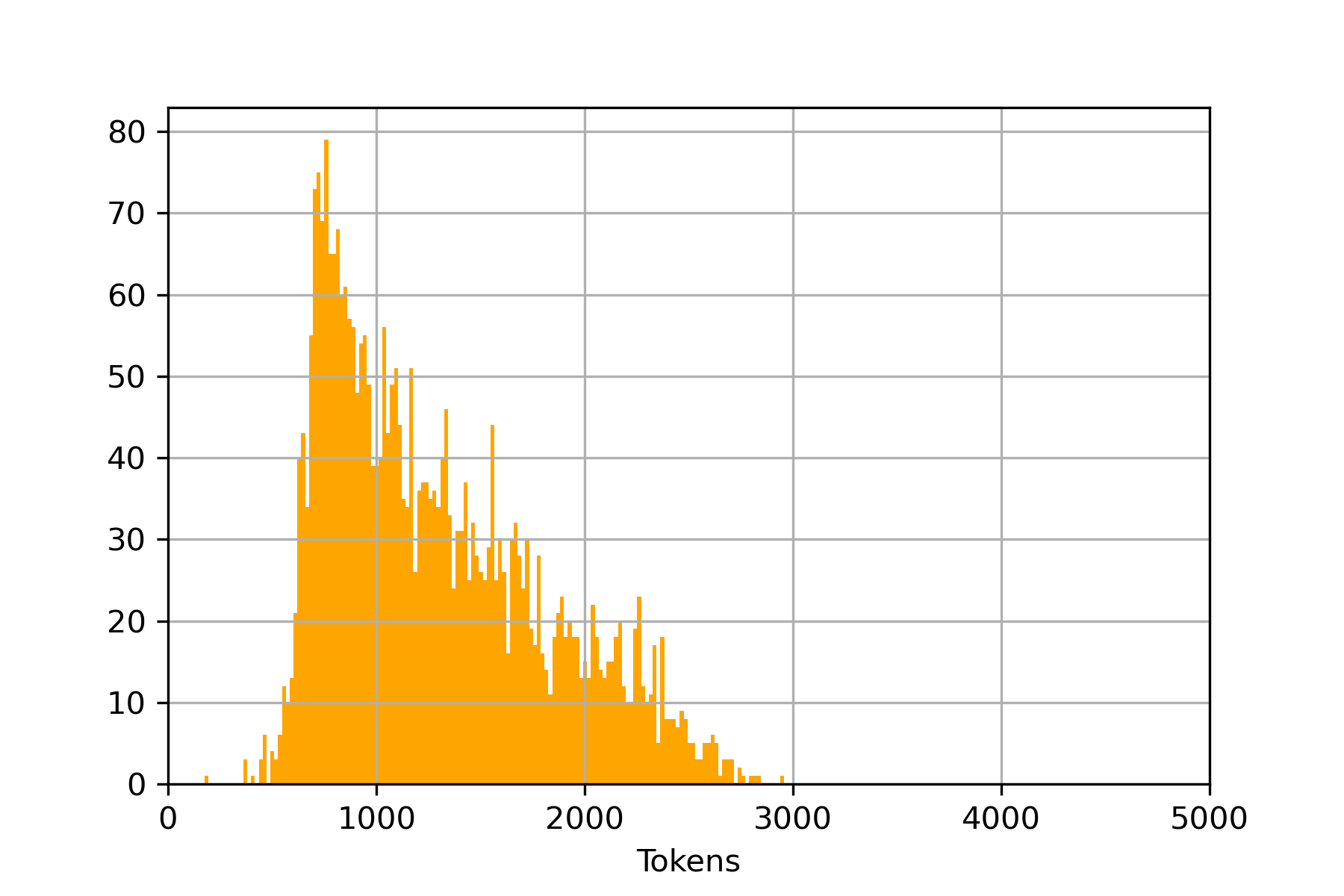} 
    }}
    \qquad
    \subfloat[
Summary\\
Mean: 178, Median: 156\\
75-Quant: 239, 95-Quant: 394, Max: 787\\
    ]{{\includegraphics[width=\textwidth/2]{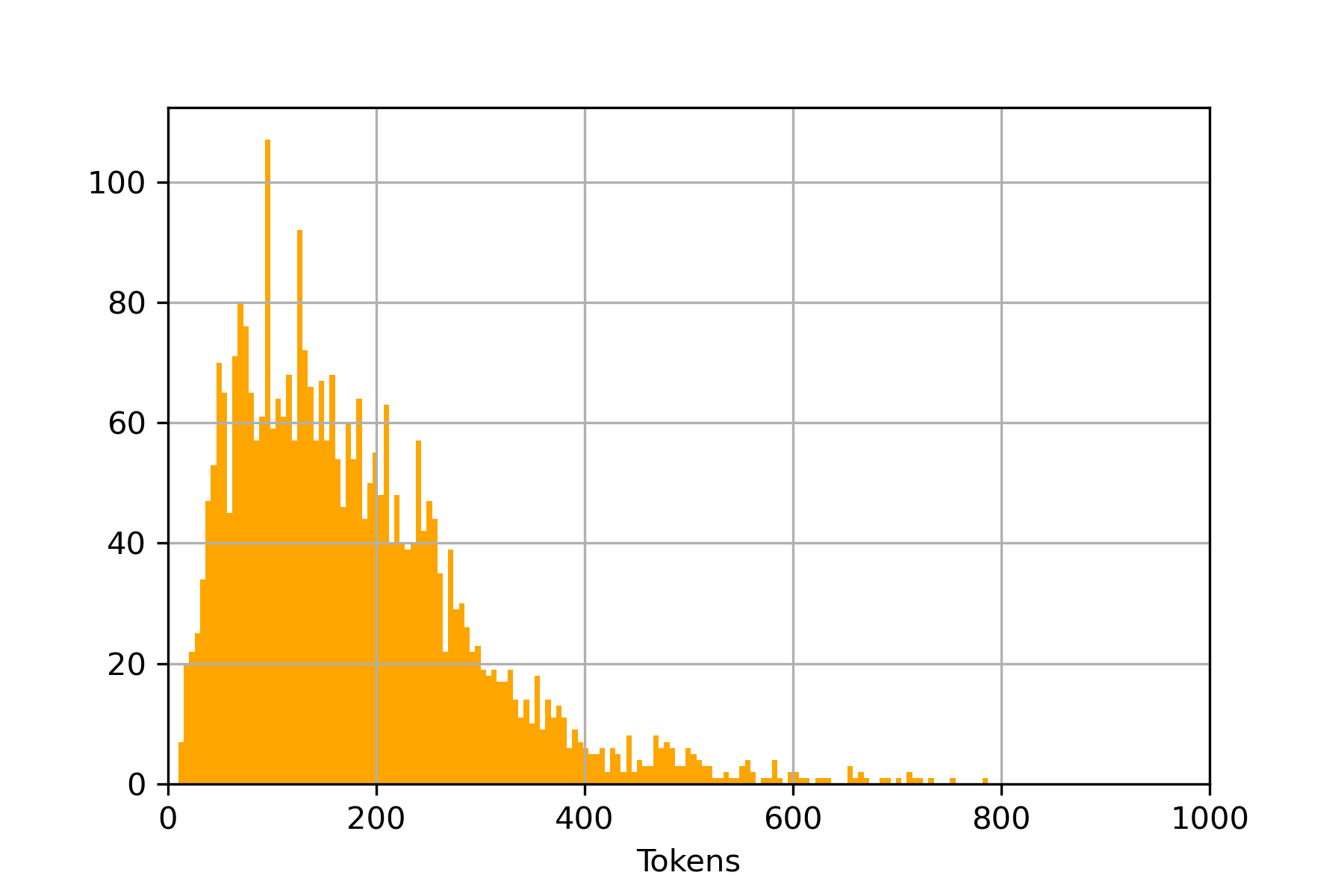} 
    }}
    }
    \caption{Histograms for the BillSum test set (3269 samples). }
    \label{fig:billsum_test}
\vspace{-5mm}
\end{figure*}

\begin{figure*}[ht]
    \resizebox{\textwidth}{!}{
    \subfloat[ 
Input Text\\
Mean: 3044, Median: 2572\\
75-Quant: 3996, 95-Quant: 7057, Max: 109759\\
]
    {{\includegraphics[width=\textwidth/2]{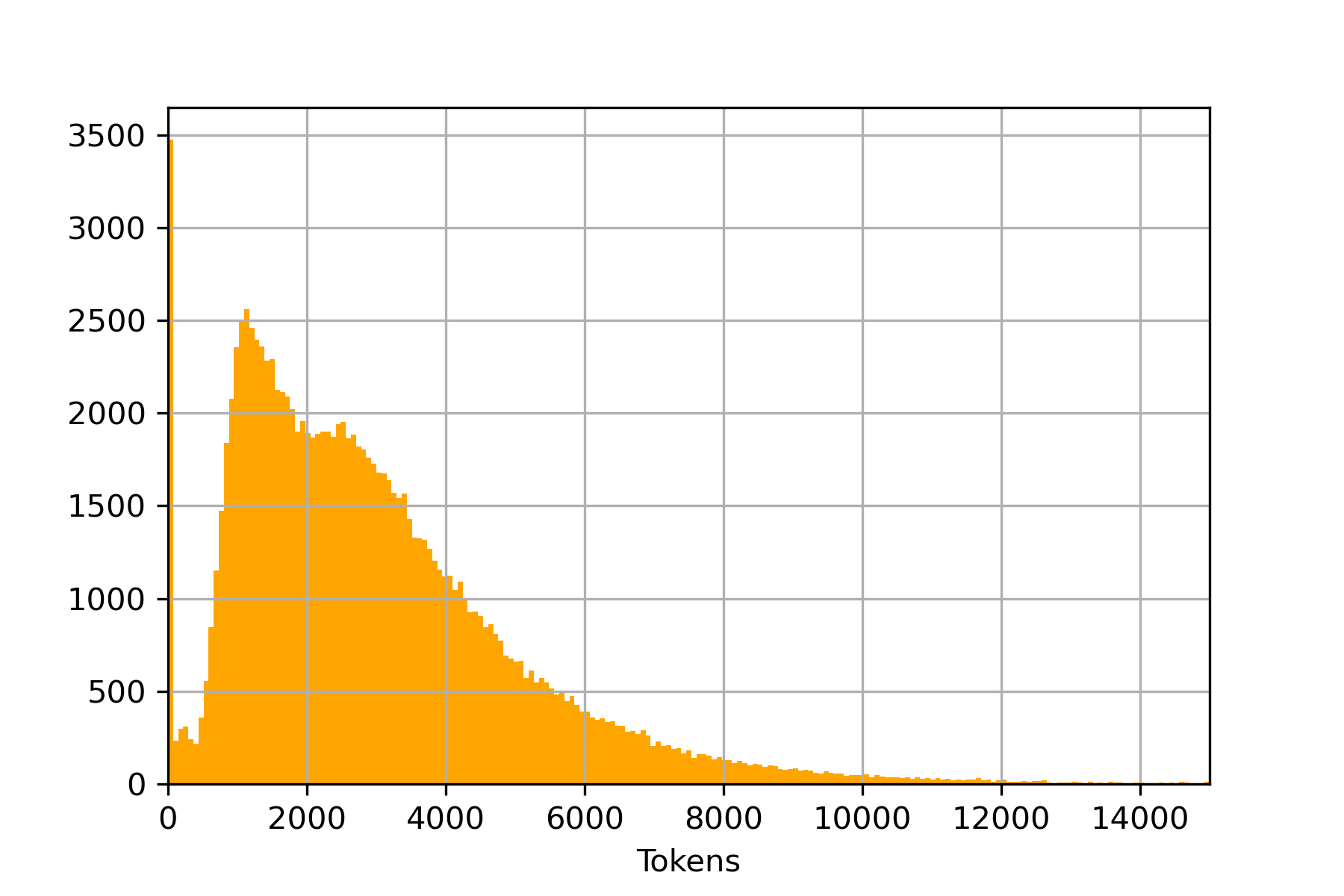} 
    }}
    \qquad
    \subfloat[
Summary\\
Mean: 202, Median: 208\\
75-Quant: 262, 95-Quant: 326, Max: 391\\
    ]
    {{\includegraphics[width=\textwidth/2]{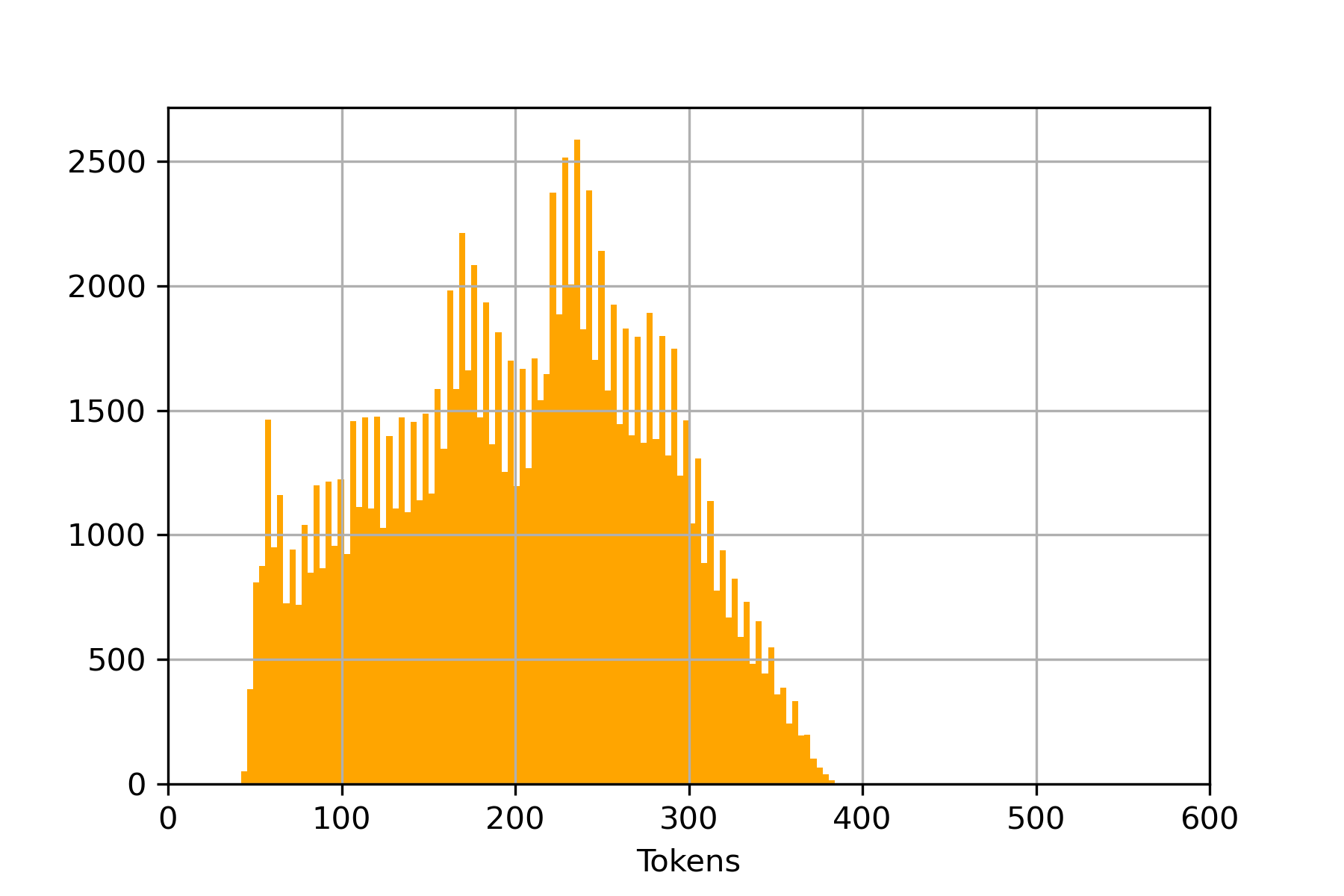} 
    }}
    }
    \caption{Histograms for the PubMed train set (119924 samples). }
    \label{fig:pubmed_train}
\vspace{-5mm}
\end{figure*}

\begin{figure*}[ht]
    \resizebox{\textwidth}{!}{
    \subfloat[ 
Input Text\\
Mean: 3112, Median: 2609\\
75-Quant: 4011, 95-Quant: 6967, Max: 119269\\
]
    {{
    \includegraphics[width=\textwidth/2]{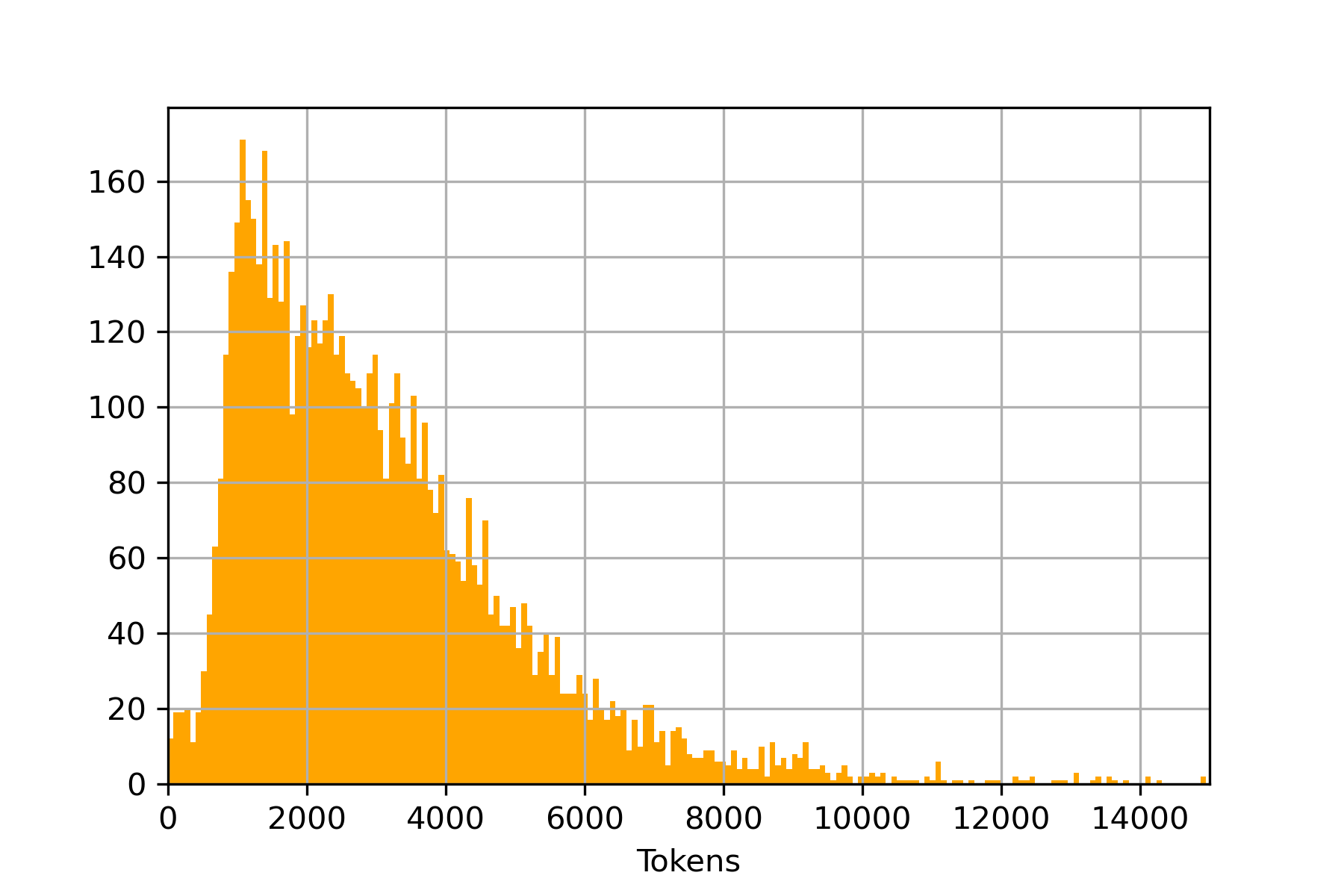} 
    }}
    \qquad
    \subfloat[
Summary\\
Mean: 202, Median: 209\\
75-Quant: 263, 95-Quant: 330, Max: 518\\
    ]
    {{
    \includegraphics[width=\textwidth/2]{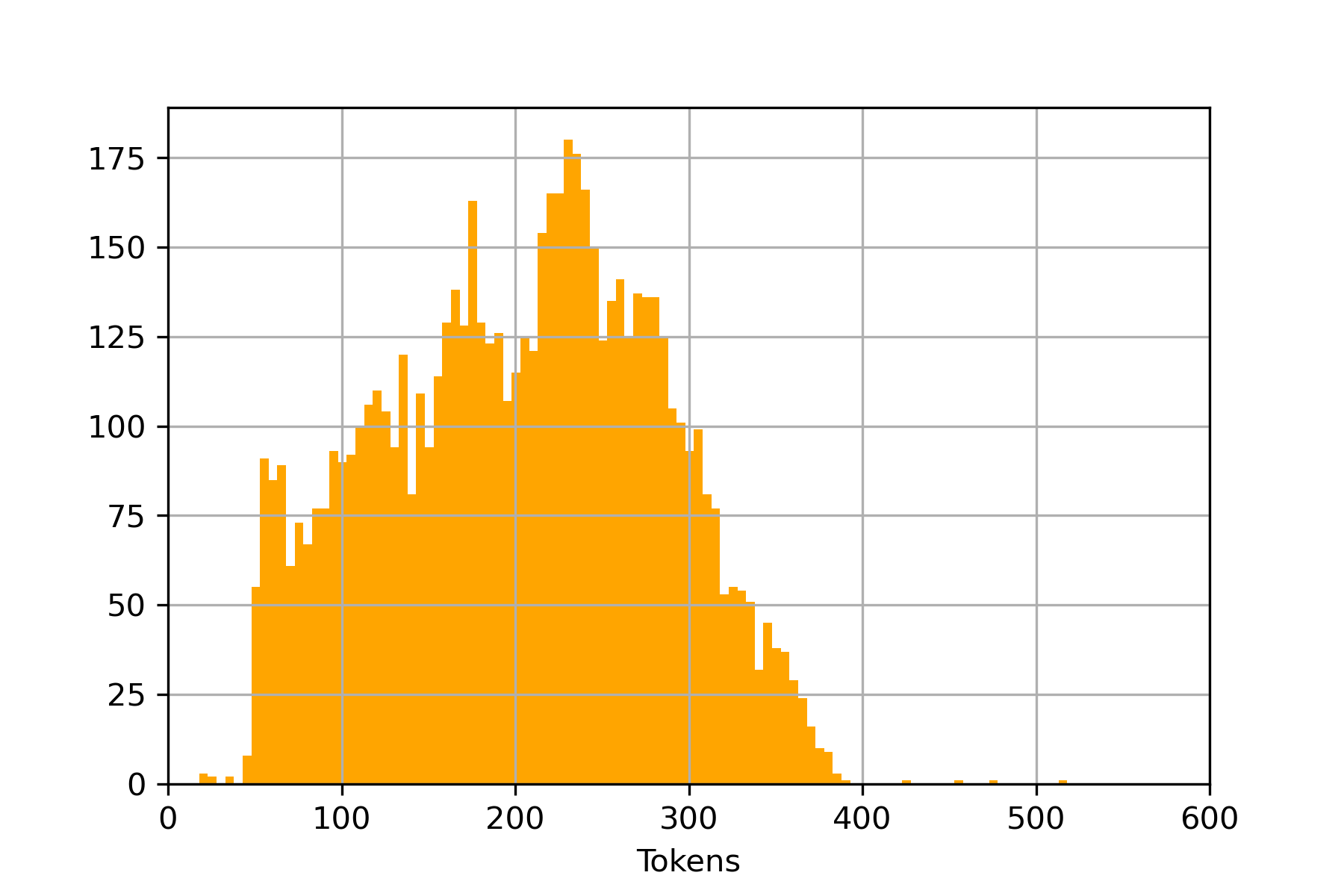} 
    }}
    }
    \caption{Histograms for the PubMed validation set (6633 samples). }
    \label{fig:pubmed_validation}
\vspace{-5mm}
\end{figure*}

\begin{figure*}[ht]
    \resizebox{\textwidth}{!}{
    \subfloat[ 
Input Text\\
Mean: 3093, Median: 2595\\
75-Quant: 3963, 95-Quant: 6985, Max: 48750\\
]
    {{
    \includegraphics[width=\textwidth/2]{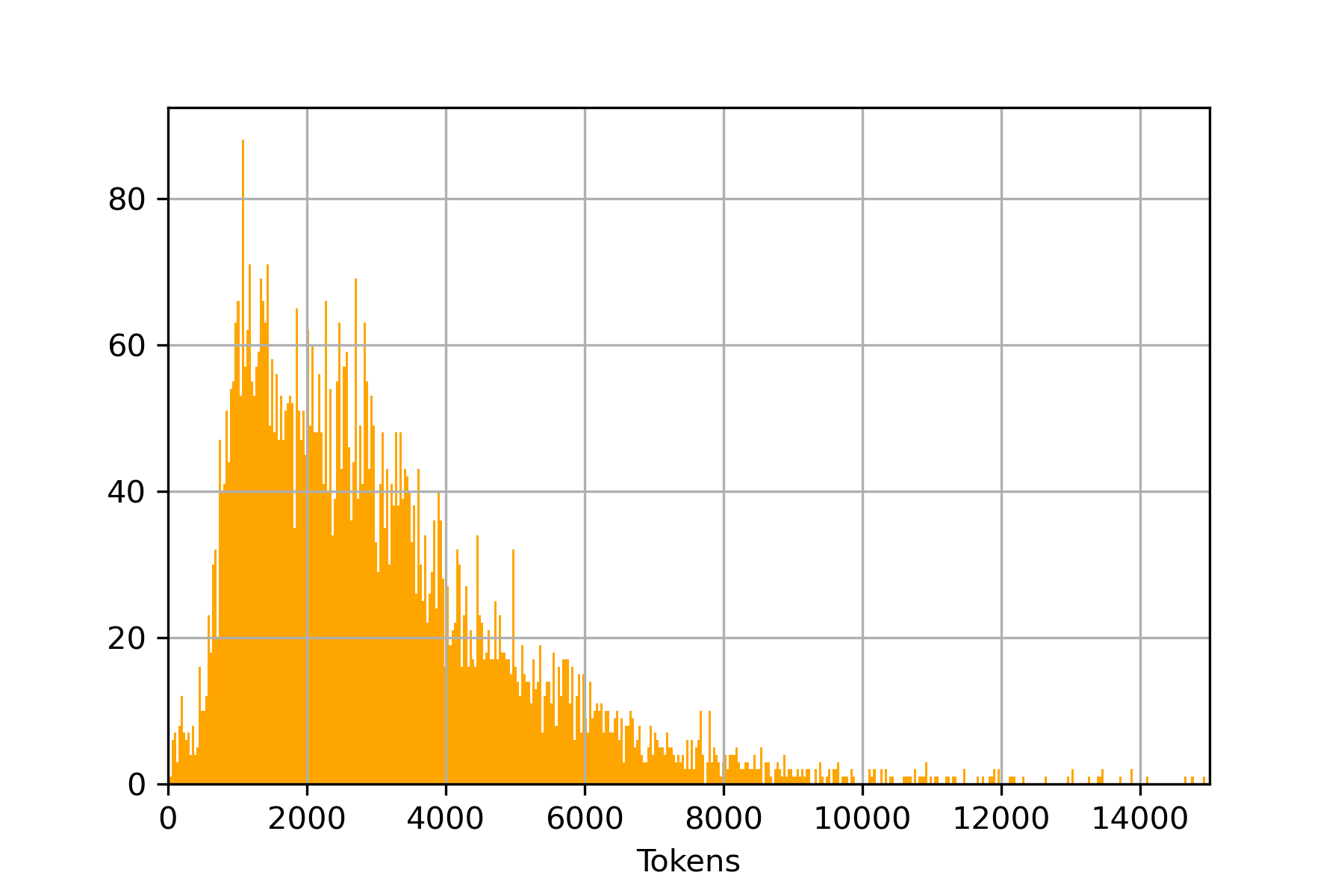} 
    }}
    \qquad
    \subfloat[
Summary\\
Mean: 205, Median: 213\\
75-Quant: 265, 95-Quant: 329, Max: 506\\
    ]
    {{
    \includegraphics[width=\textwidth/2]{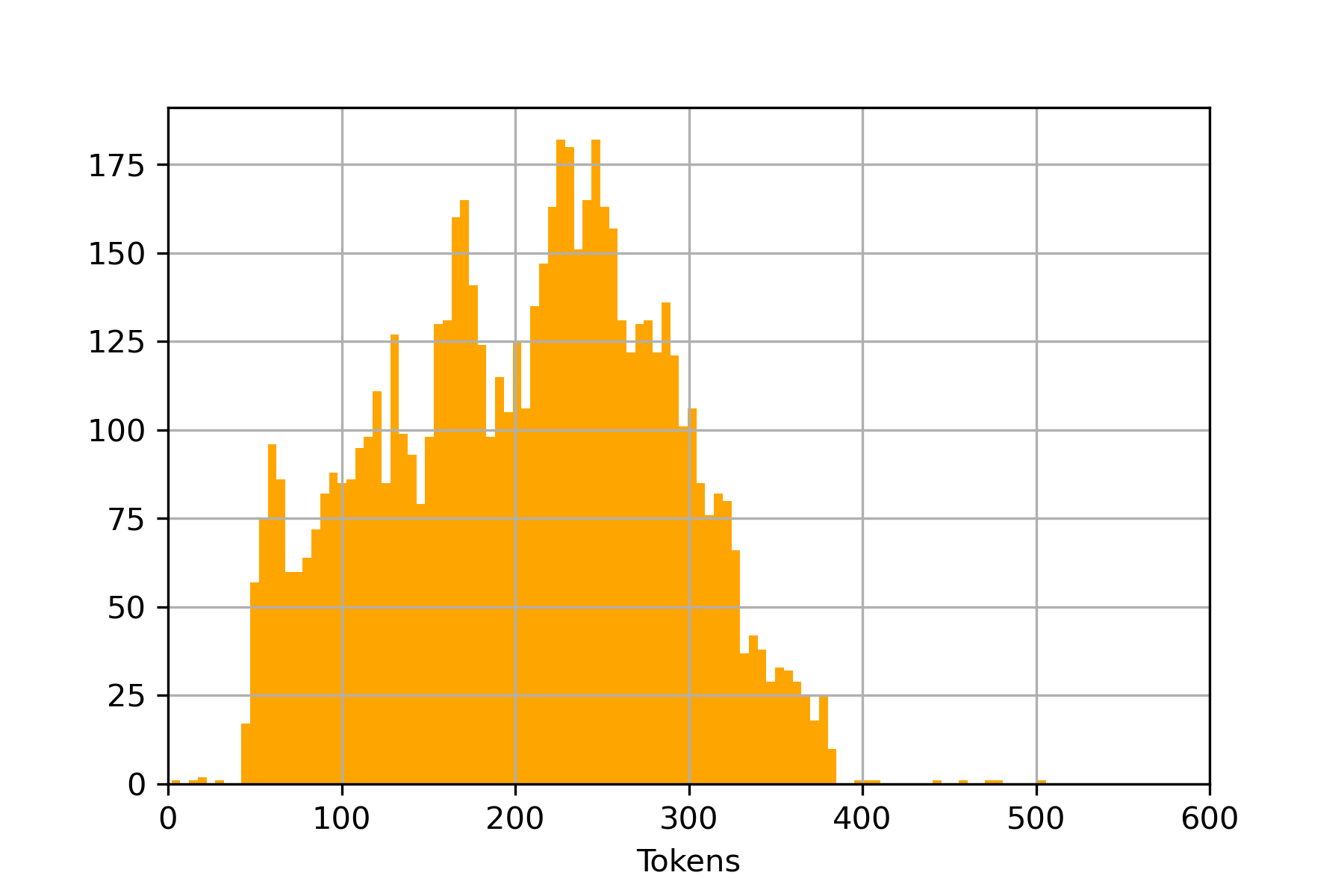} 
    }}
    }
    \caption{Histograms for the PubMed test set (6658 samples). }
    \label{fig:pubmed_test}
\vspace{-5mm}
\end{figure*}

\begin{figure*}[ht]
    \resizebox{\textwidth}{!}{
    \subfloat[ 
CaseHold\\ 
Mean: 136, Median: 137\\
75-Quant: 141, 95-Quant: 148, Max: 207\\
]
    {{
    \includegraphics[width=\textwidth/2]{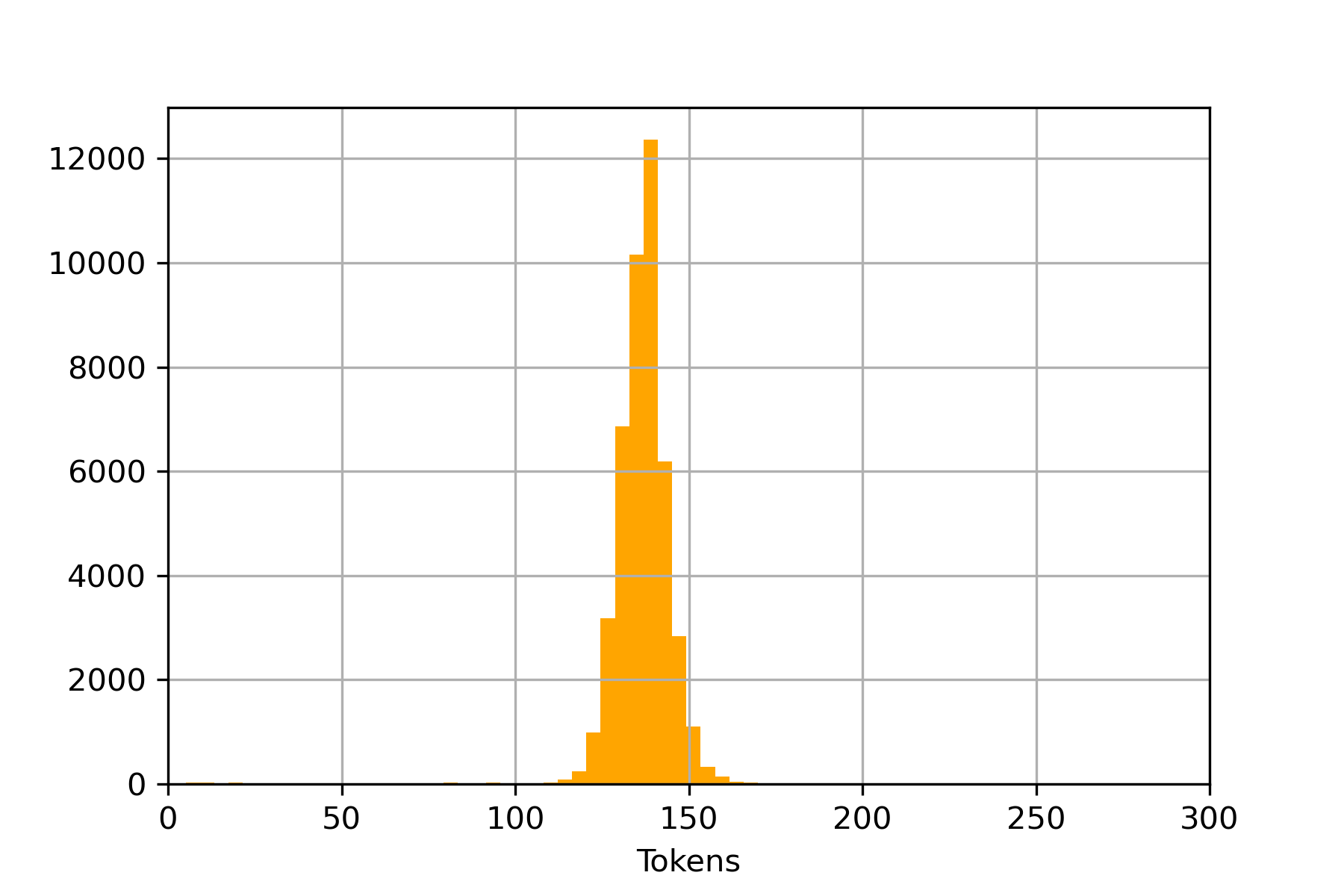} 
    }}
    \qquad
    \subfloat[
ECTHR\_A\\ 
Mean: 1619, Median: 984\\
75-Quant: 2002, 95-Quant: 5062, Max: 35416\\
    ]
    {{
    \includegraphics[width=\textwidth/2]{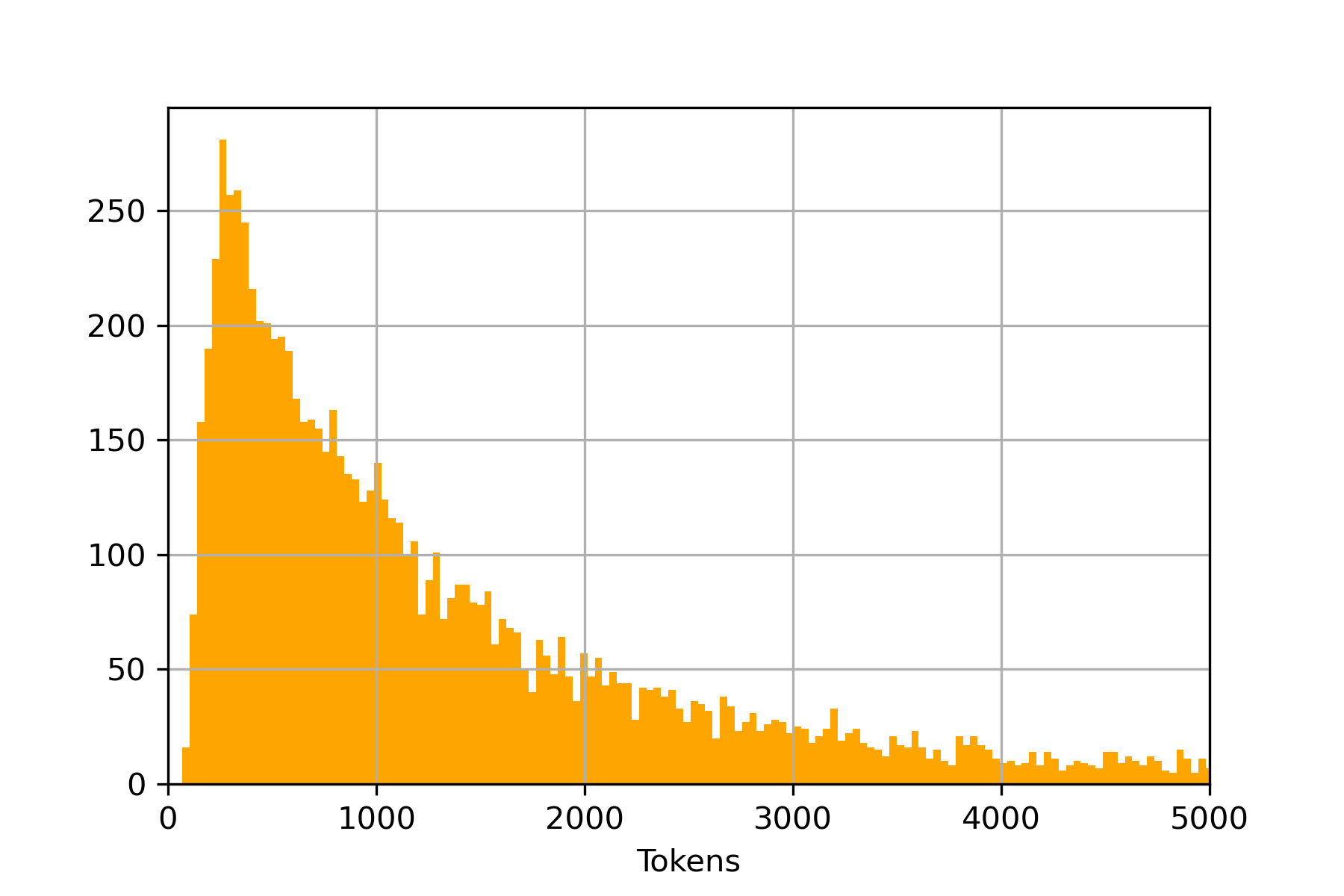} 
    }}
    }

\resizebox{\textwidth}{!}{
    \subfloat[ 
ECTHR\_B\\ 
Mean: 1619, Median: 984\\
75-Quant: 2002, 95-Quant: 5062, Max: 35416\\
]
    {{
    \includegraphics[width=\textwidth/2]{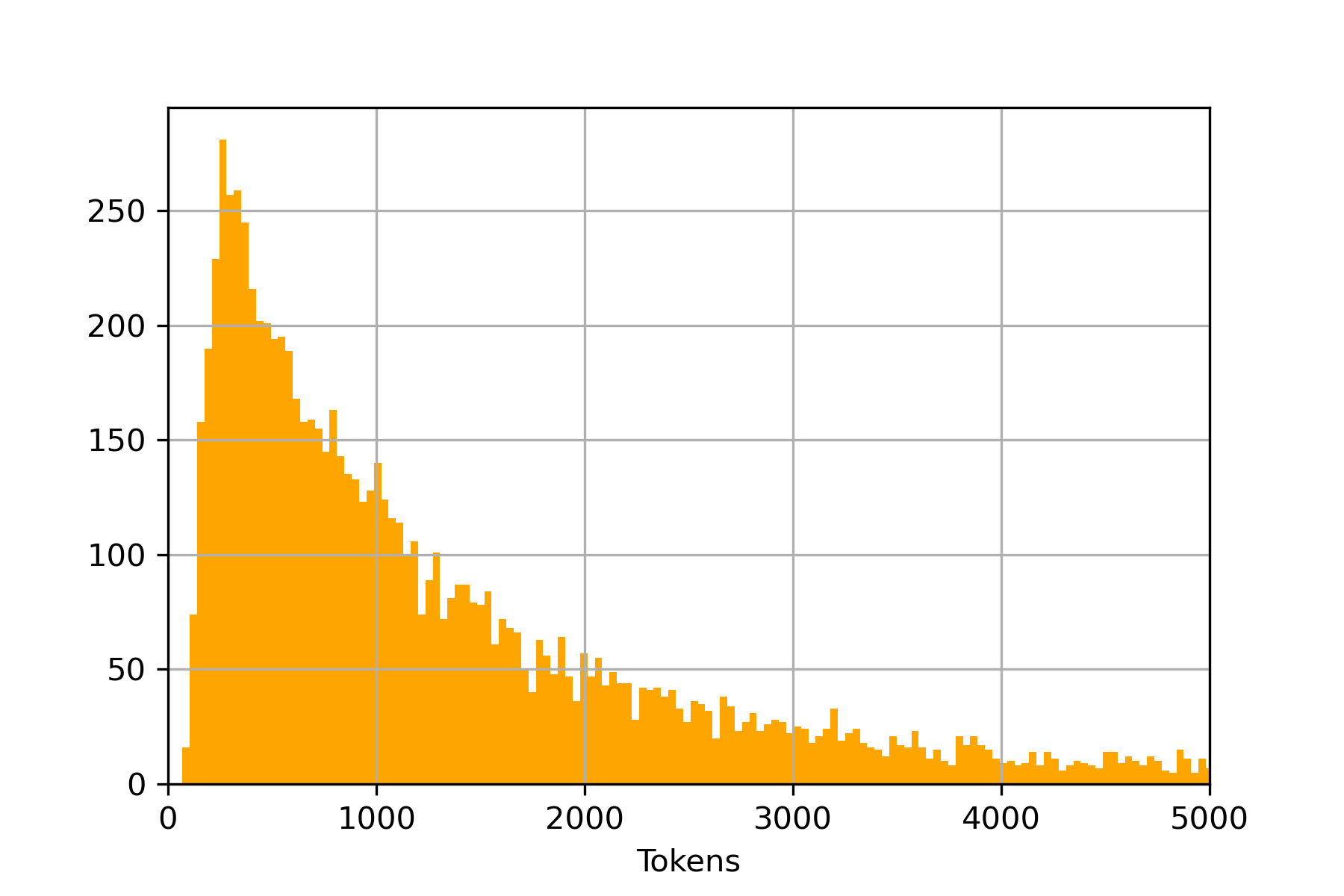} 
    }}
    \qquad
    \subfloat[
EurLex\\ 
Mean: 1133, Median: 453\\
75-Quant: 871, 95-Quant: 4147, Max: 140103\\
    ]
    {{
    \includegraphics[width=\textwidth/2]{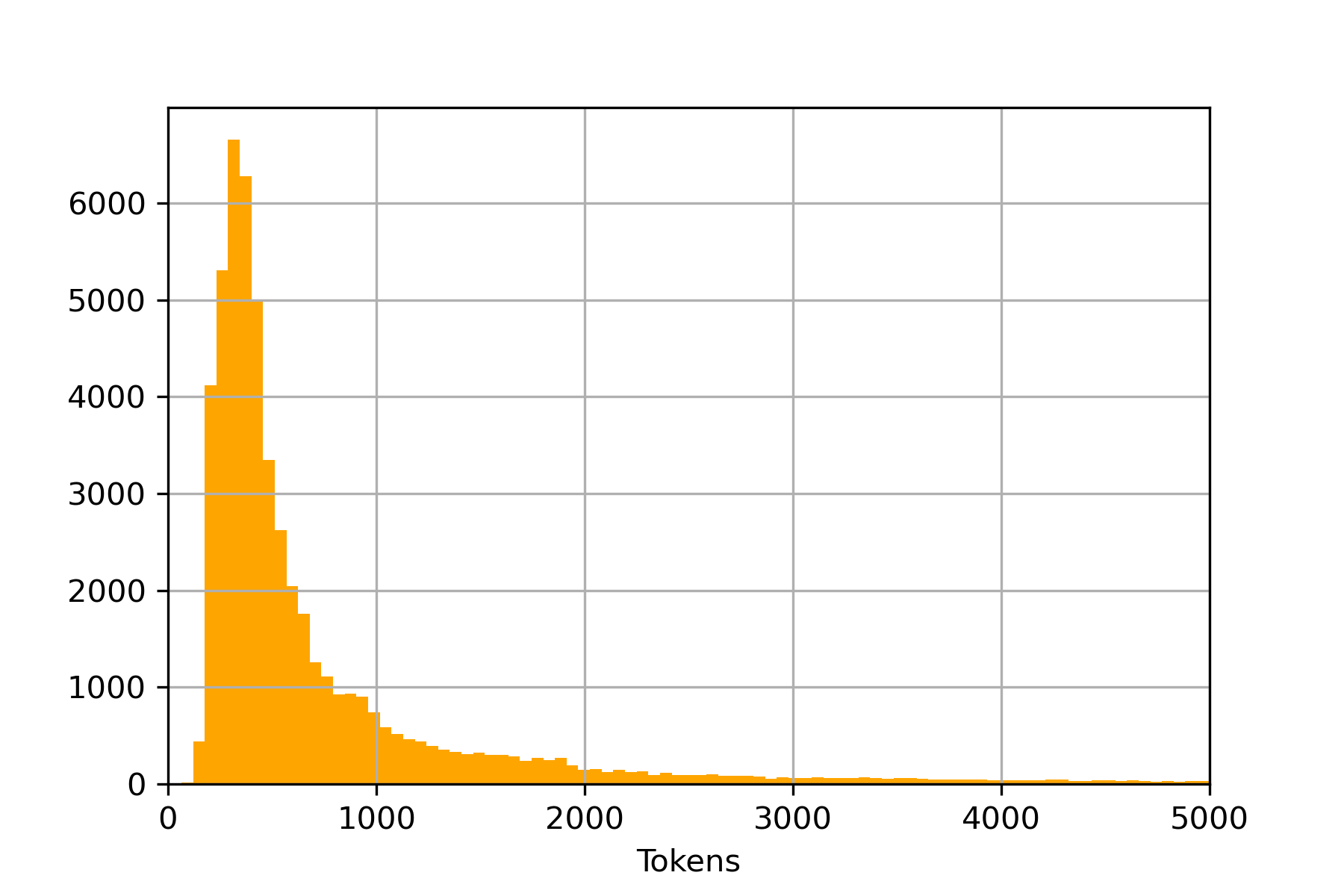} 
    }}
    }

\resizebox{\textwidth}{!}{
    \subfloat[
LEDGAR\\ 
Mean: 3044, Median: 2572\\
75-Quant: 3996, 95-Quant: 7057, Max: 109759\\
]
    {{
    \includegraphics[width=\textwidth/2]{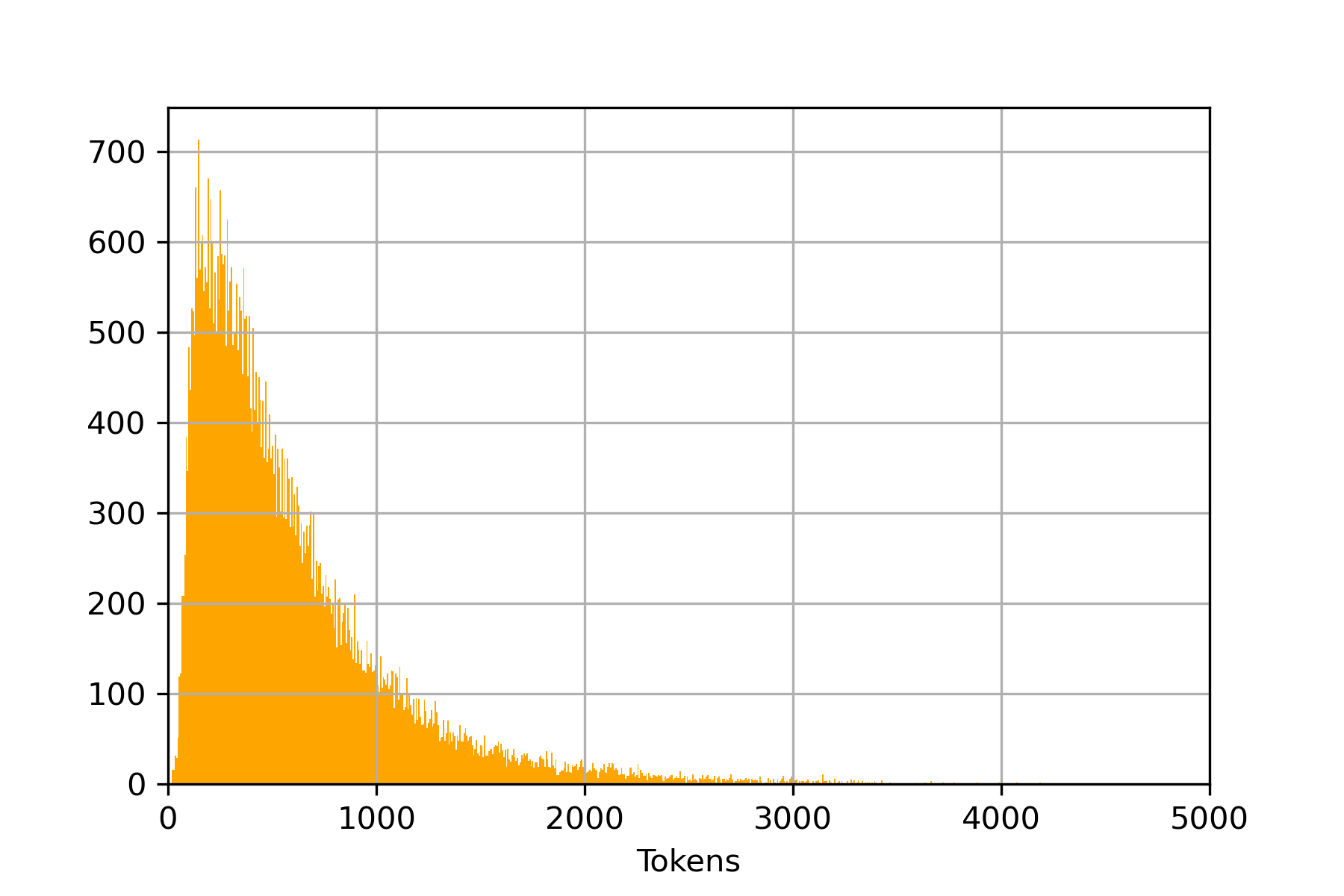} 
    }}
    \qquad
    \subfloat[
SCOTUS\\ 
Mean: 5853, Median: 4364\\
75-Quant: 7933, 95-Quant: 16168, Max: 88566\\
    ]
    {{
    \includegraphics[width=\textwidth/2]{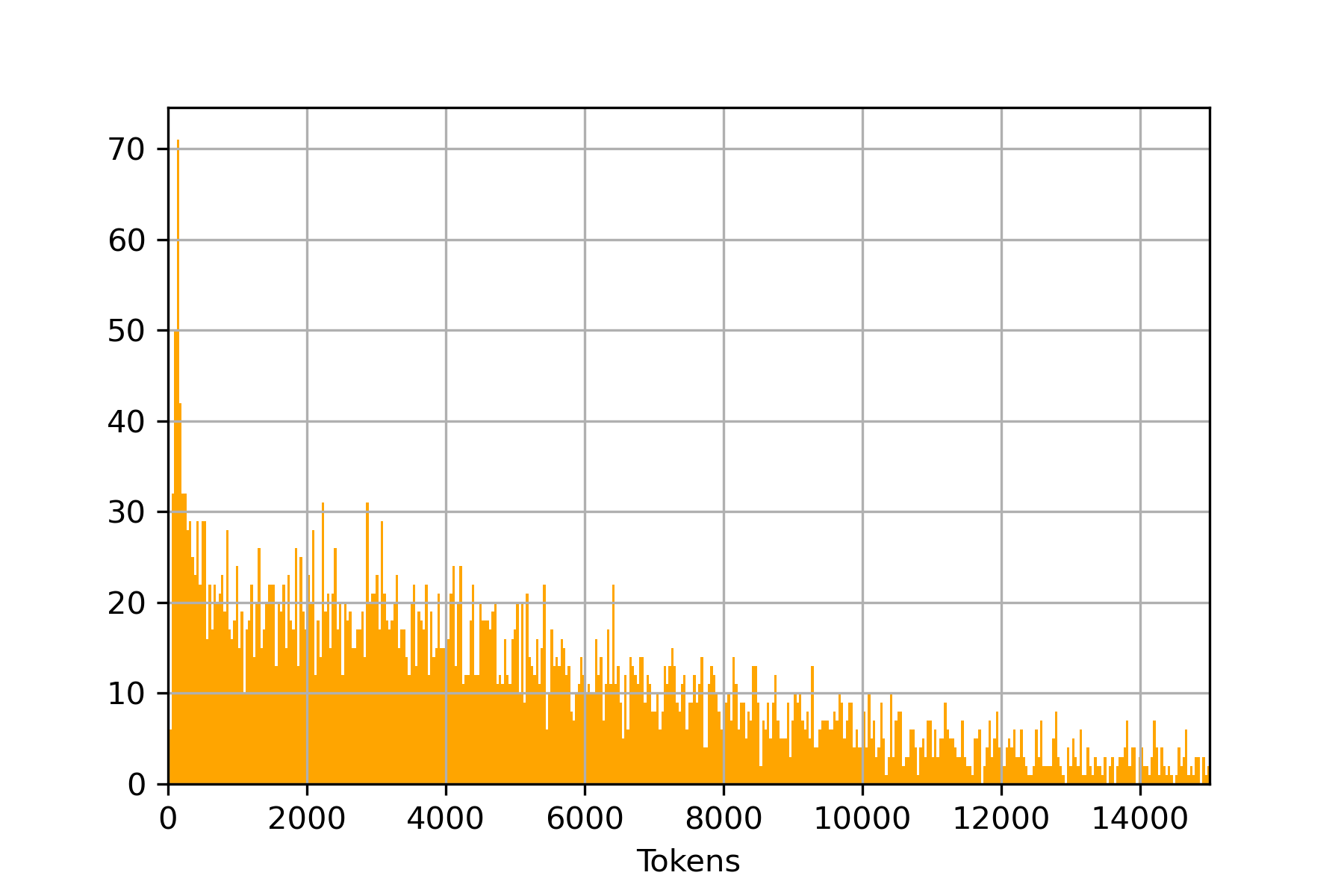} 
    }}
    }

\resizebox{\textwidth}{!}{
    \subfloat[ 
Unfair\_TOS\\ 
Mean: 145, Median: 120\\
75-Quant: 180, 95-Quant: 335, Max: 2093\\
]
    {{
    \includegraphics[width=\textwidth/2]{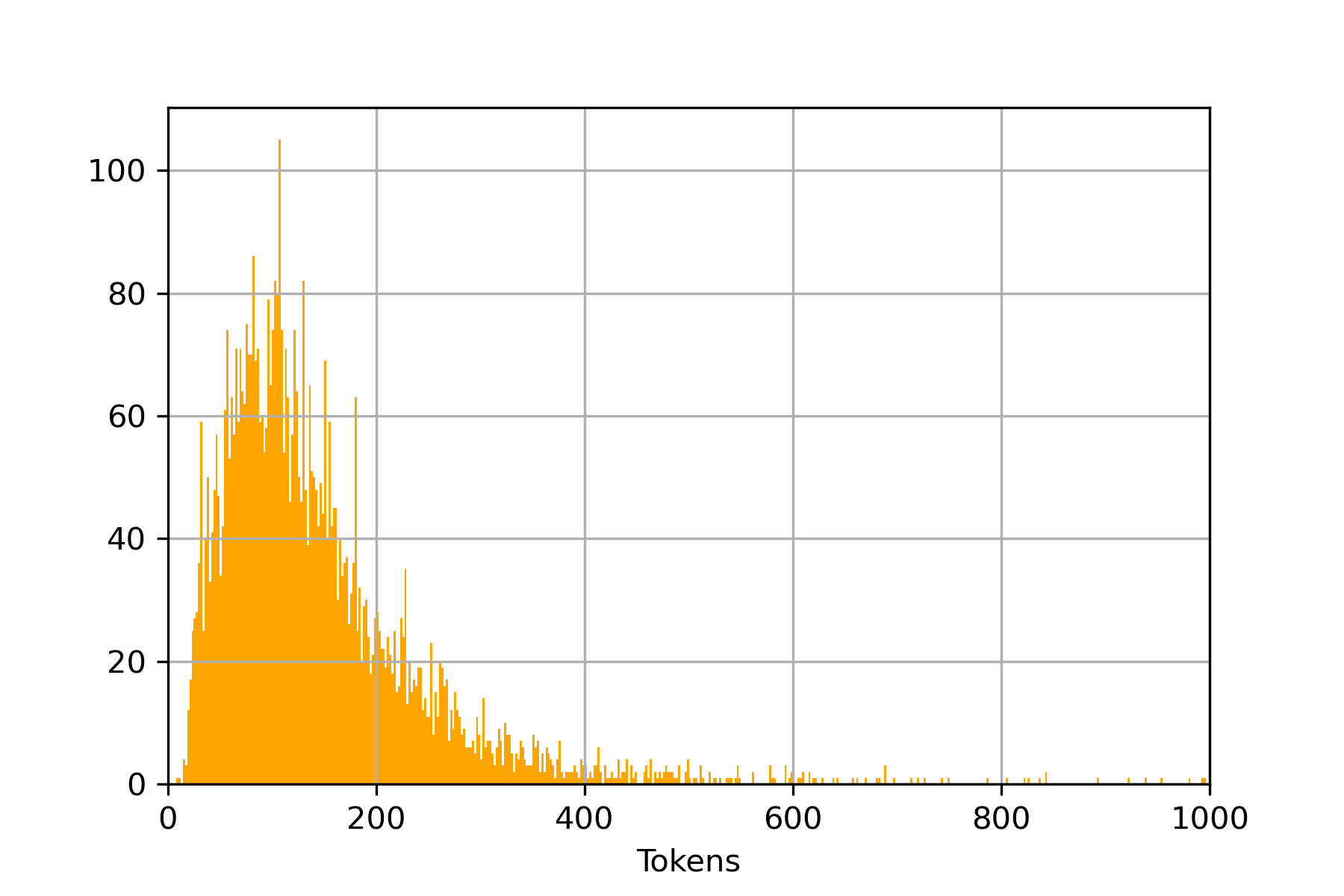} 
    }}
    \qquad

    {{
    \includegraphics[width=\textwidth/2]{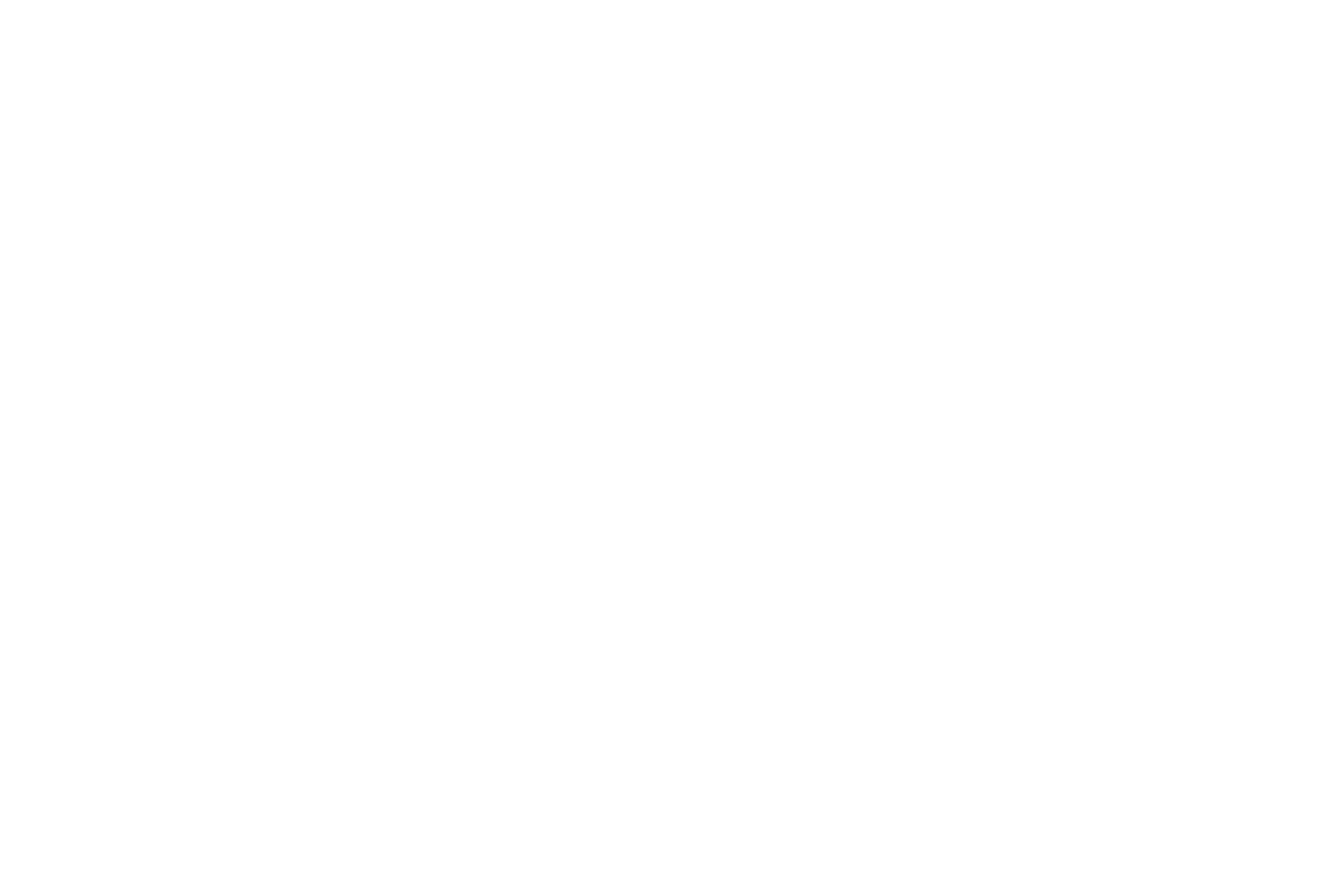} 
    }}
    }

    \caption{Histograms for the LexGLUE train set. }
    \label{fig:lexglue_train}
\vspace{-5mm}
\end{figure*}

\begin{figure*}[ht]
    \resizebox{\textwidth}{!}{
    \subfloat[
CaseHold\\ 
Mean: 136, Median: 137\\
75-Quant: 142, 95-Quant: 148, Max: 185\\
]
    {{
    \includegraphics[width=\textwidth/2]{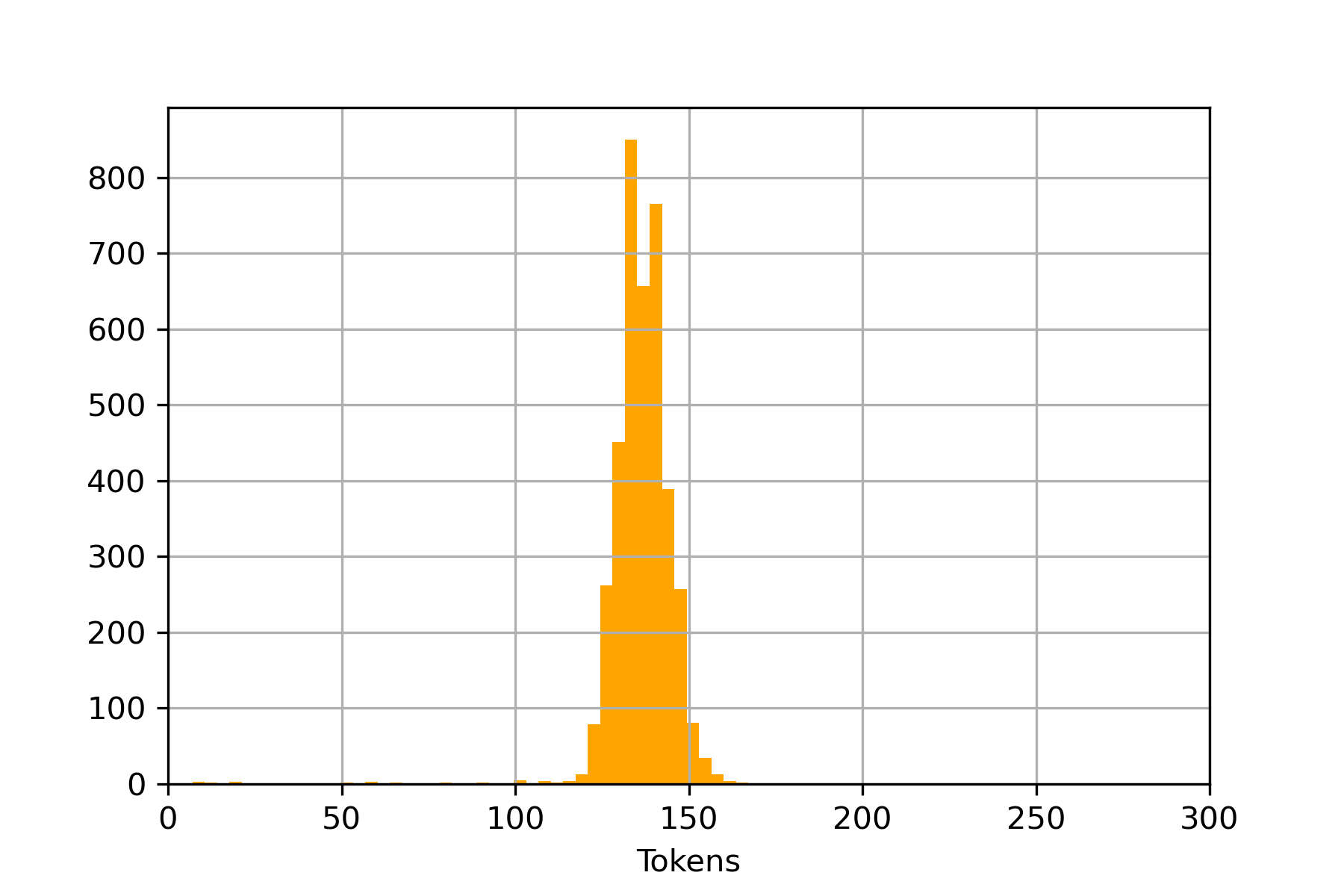} 
    }}
    \qquad
    \subfloat[
ECTHR\_A\\ 
Mean: 1619, Median: 984\\
75-Quant: 2002, 95-Quant: 5062, Max: 35416\\
    ]
    {{
    \includegraphics[width=\textwidth/2]{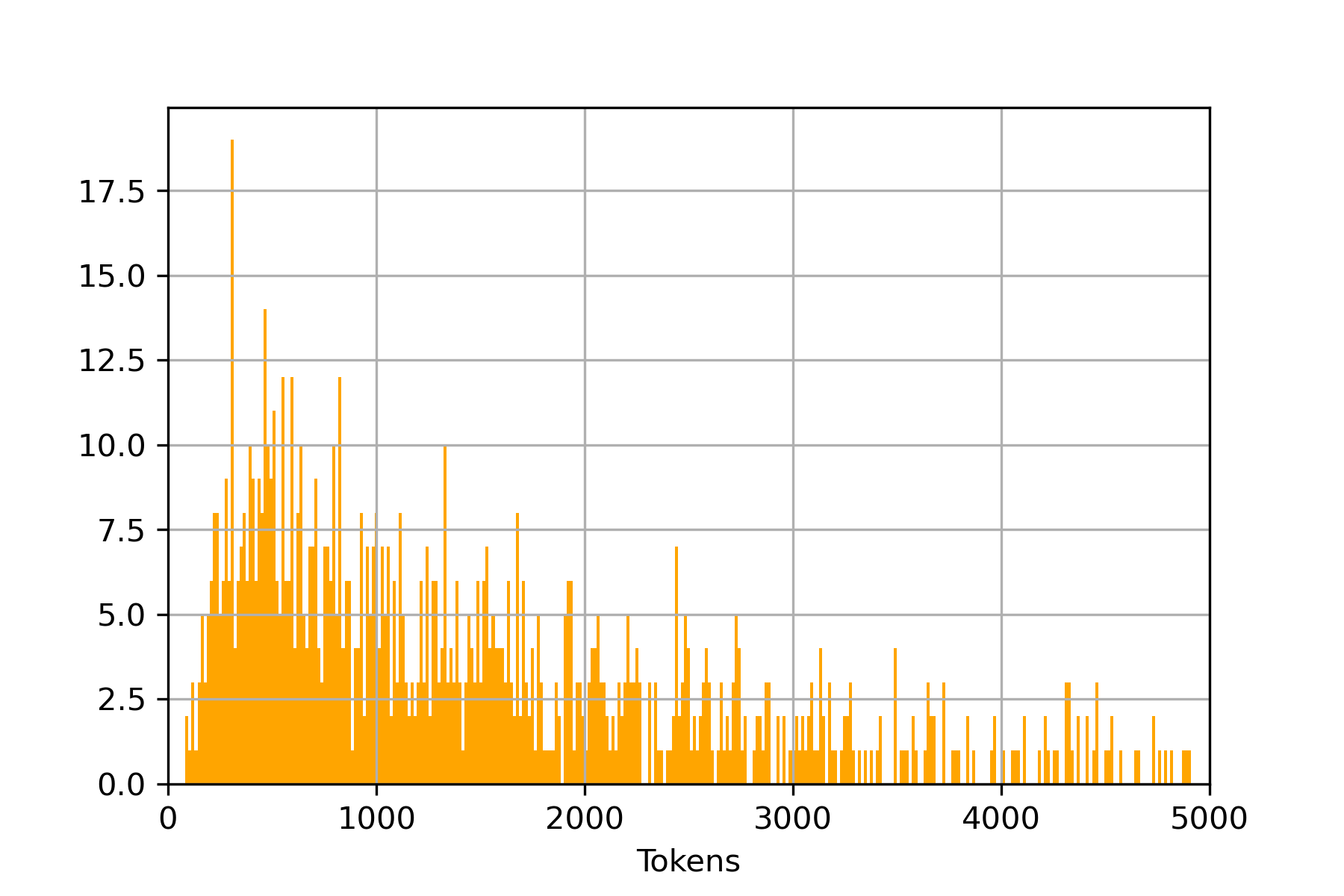} 
    }}
    }

\resizebox{\textwidth}{!}{
    \subfloat[ 
ECTHR\_B\\ 
Mean: 1784, Median: 1255\\
75-Quant: 2320, 95-Quant: 5119, Max: 14493\\
]
    {{
    \includegraphics[width=\textwidth/2]{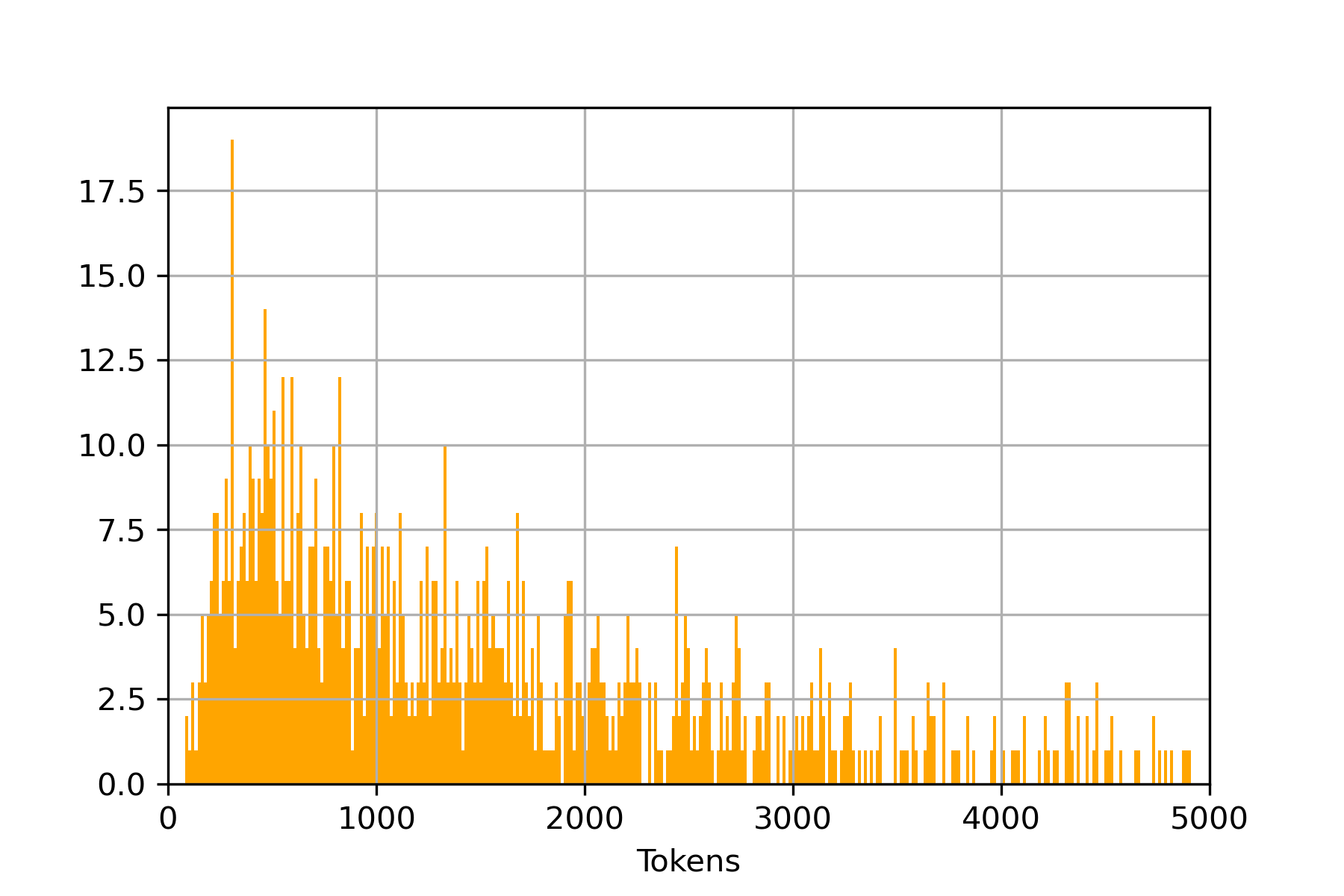} 
    }}
    \qquad
    \subfloat[
EurLex\\ 
Mean: 1312, Median: 432\\
75-Quant: 910, 95-Quant: 5664, Max: 51944\\
    ]
    {{
    \includegraphics[width=\textwidth/2]{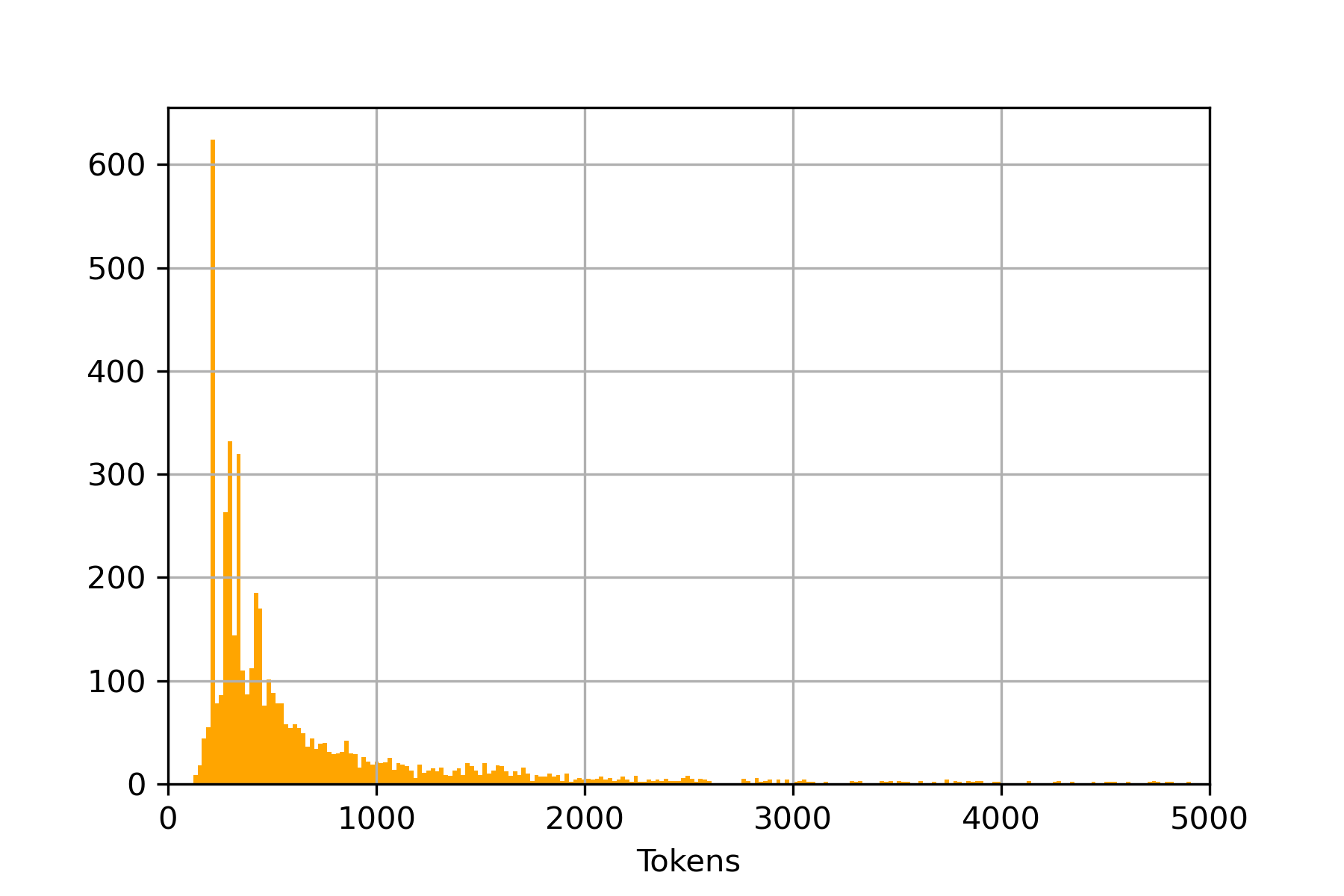} 
    }}
    }

\resizebox{\textwidth}{!}{
    \subfloat[ 
LEDGAR\\ 
Mean: 587, Median: 444\\
75-Quant: 775, 95-Quant: 1549, Max: 4706\\
]
    {{
    \includegraphics[width=\textwidth/2]{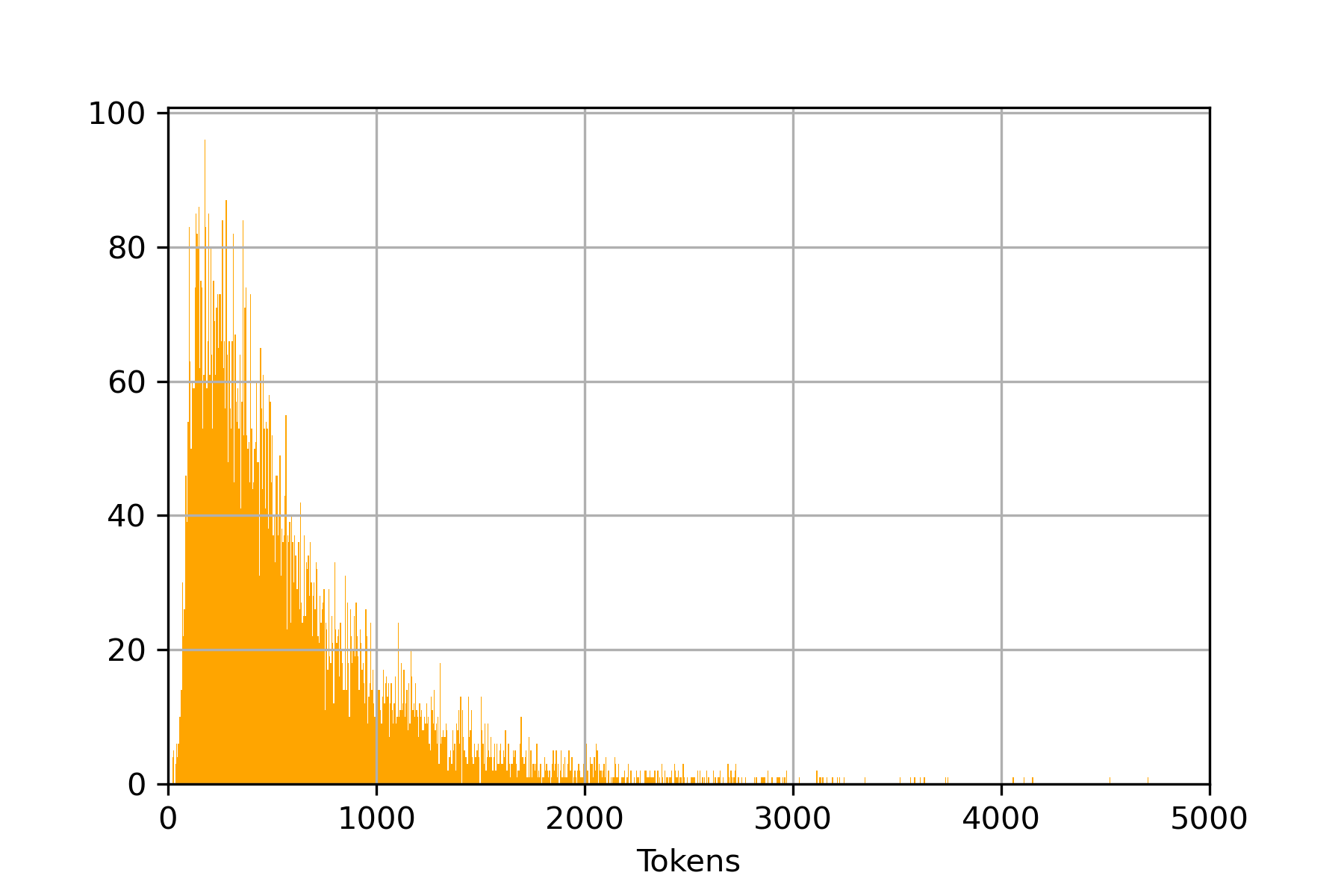} 
    }}
    \qquad
    \subfloat[
SCOTUS\\ 
Mean: 8625, Median: 7591\\
75-Quant: 11558, 95-Quant: 19719, Max: 38098\\
    ]
    {{
    \includegraphics[width=\textwidth/2]{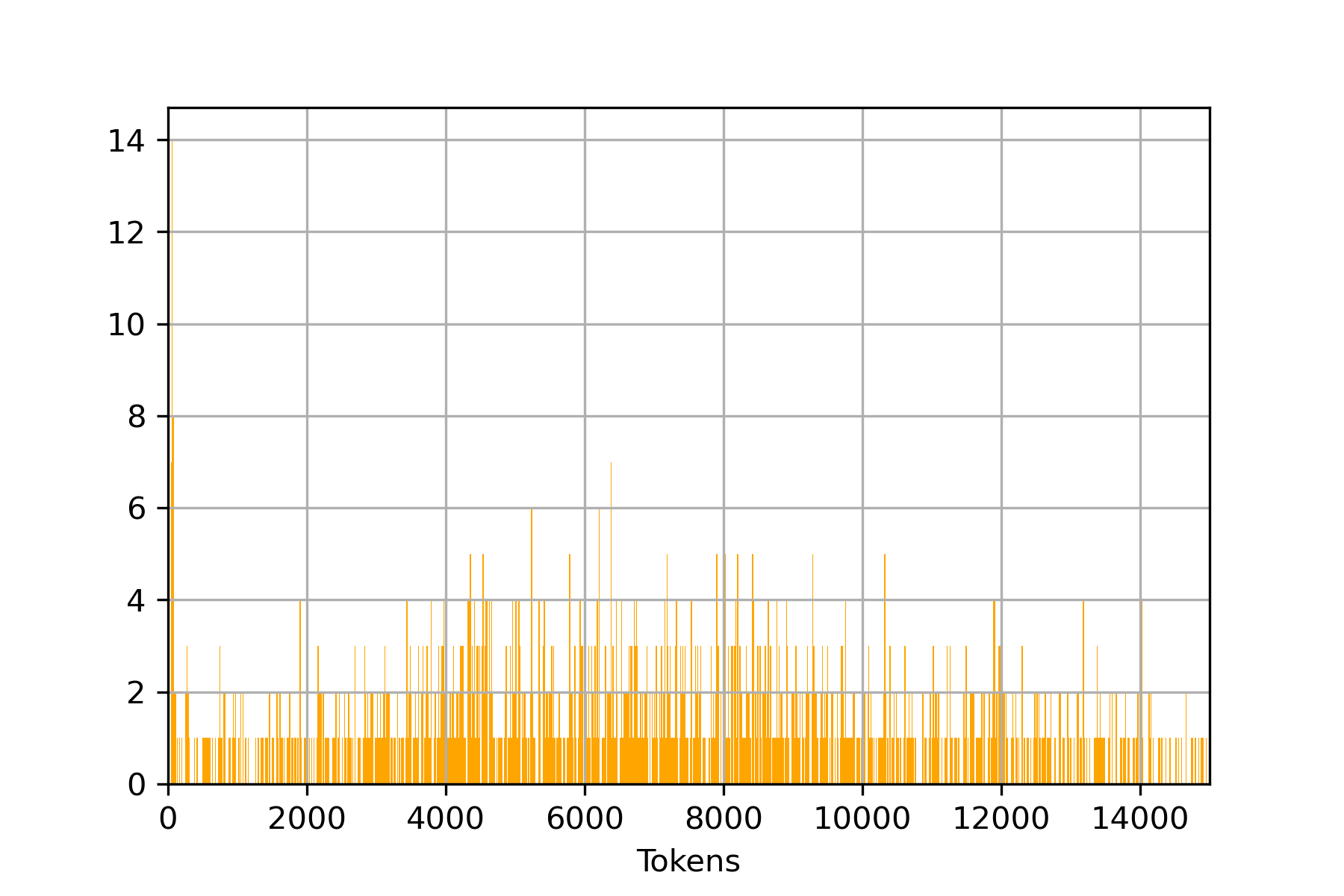} 
    }}
    }

\resizebox{\textwidth}{!}{
    \subfloat[ 
Unfair\_TOS\\ 
Mean: 154, Median: 125\\
75-Quant: 192, 95-Quant: 356, Max: 1104\\
]
    {{
    \includegraphics[width=\textwidth/2]{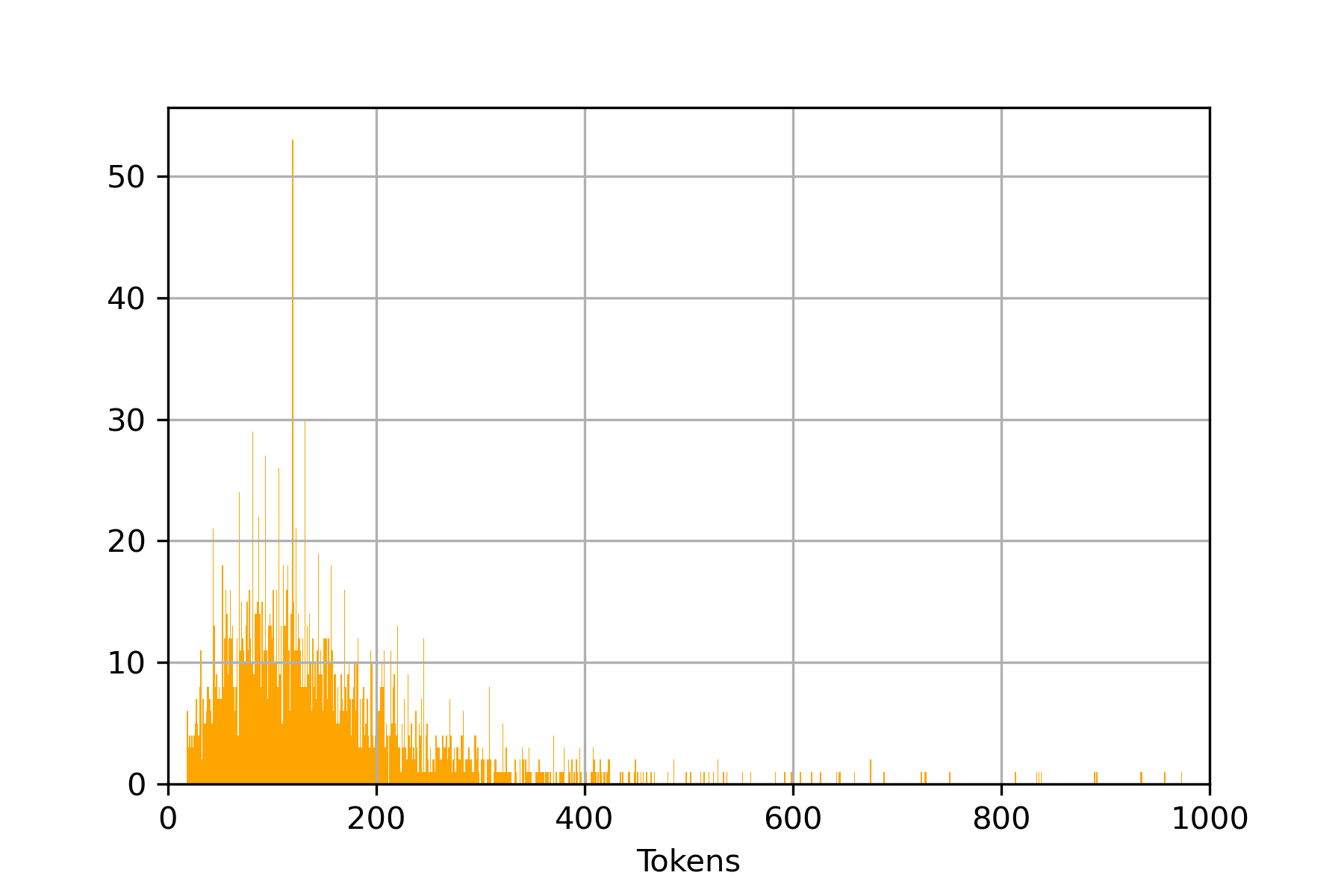} 
    }}
    \qquad

    {{
    \includegraphics[width=\textwidth/2]{data_info/empty_image.png} 
    }}
    }

    \caption{Histograms for the LexGLUE validation set. }
    \label{fig:lexglue_validation}
\vspace{-5mm}
\end{figure*}

\begin{figure*}[ht]
    \resizebox{\textwidth}{!}{
    \subfloat[ 
CaseHold\\ 
Mean: 136, Median: 137\\
75-Quant: 141, 95-Quant: 148, Max: 168\\
]
    {{
    \includegraphics[width=\textwidth/2]{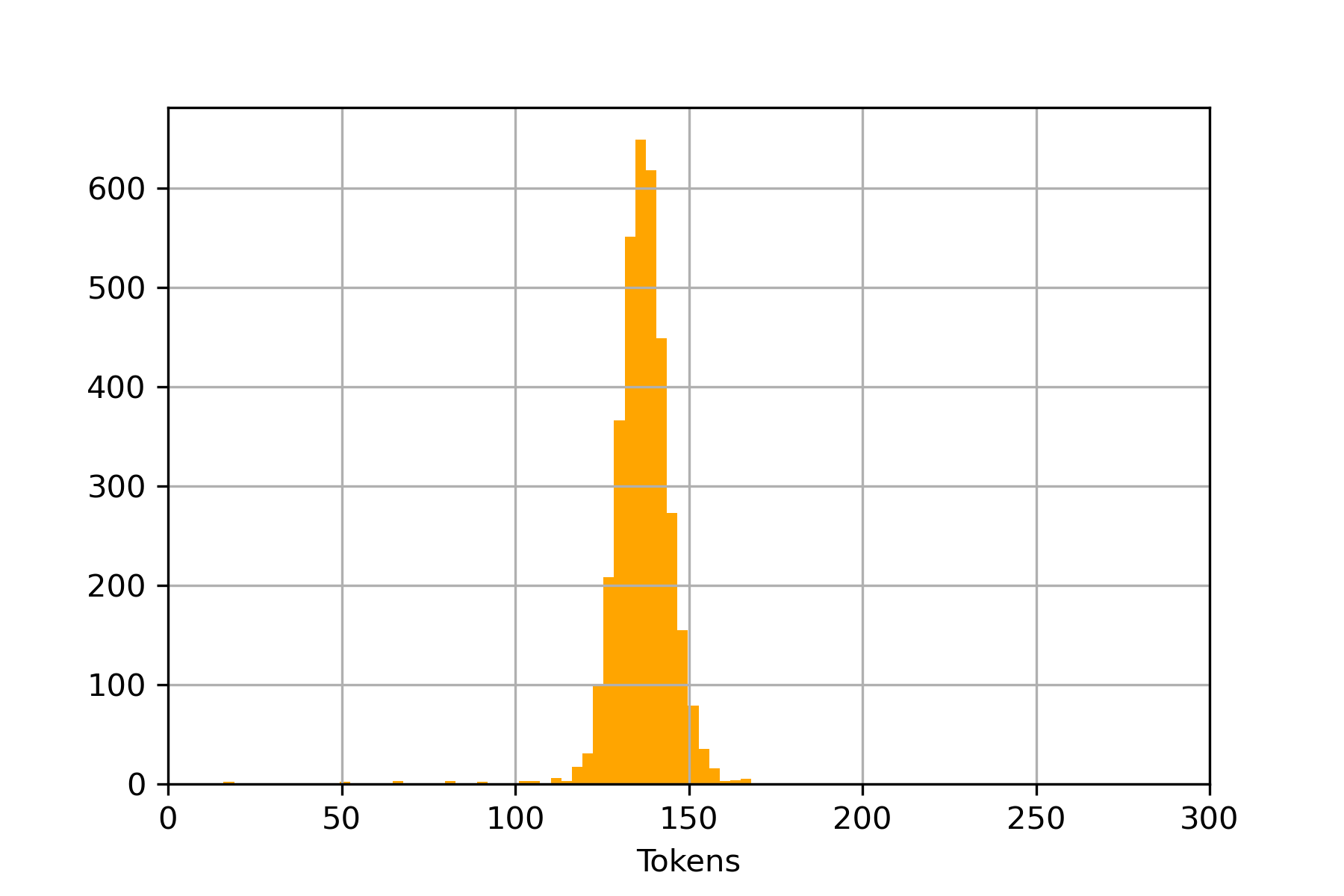} 
    }}
    \qquad
    \subfloat[
ECTHR\_A\\ 
Mean: 1925, Median: 1412\\
75-Quant: 2401, 95-Quant: 5700, Max: 15919\\
    ]
    {{
    \includegraphics[width=\textwidth/2]{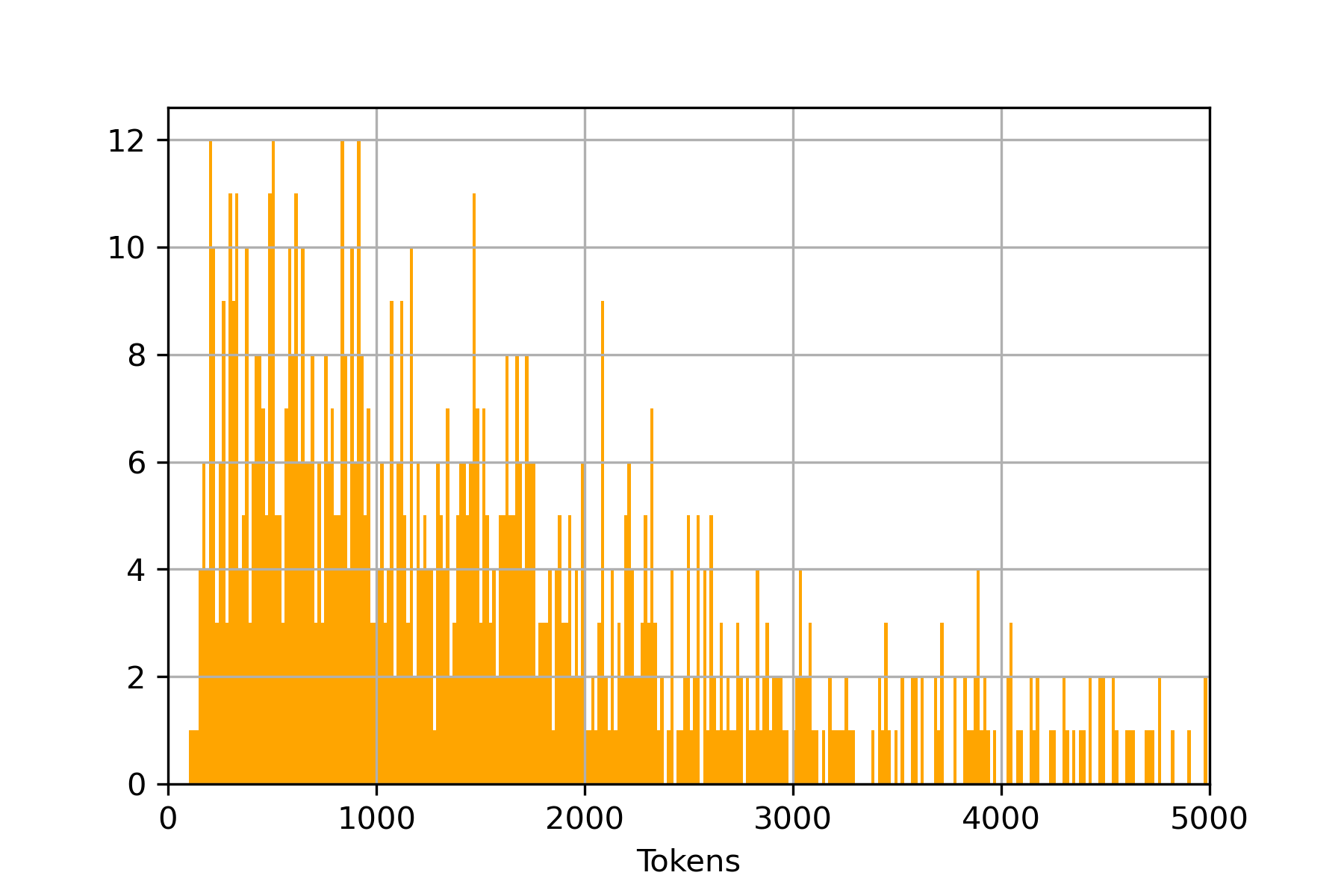} 
    }}
    }

\resizebox{\textwidth}{!}{
    \subfloat[ 
ECTHR\_B\\ 
Mean: 1925, Median: 1412\\
75-Quant: 2401, 95-Quant: 5700, Max: 15919\\
]
    {{
    \includegraphics[width=\textwidth/2]{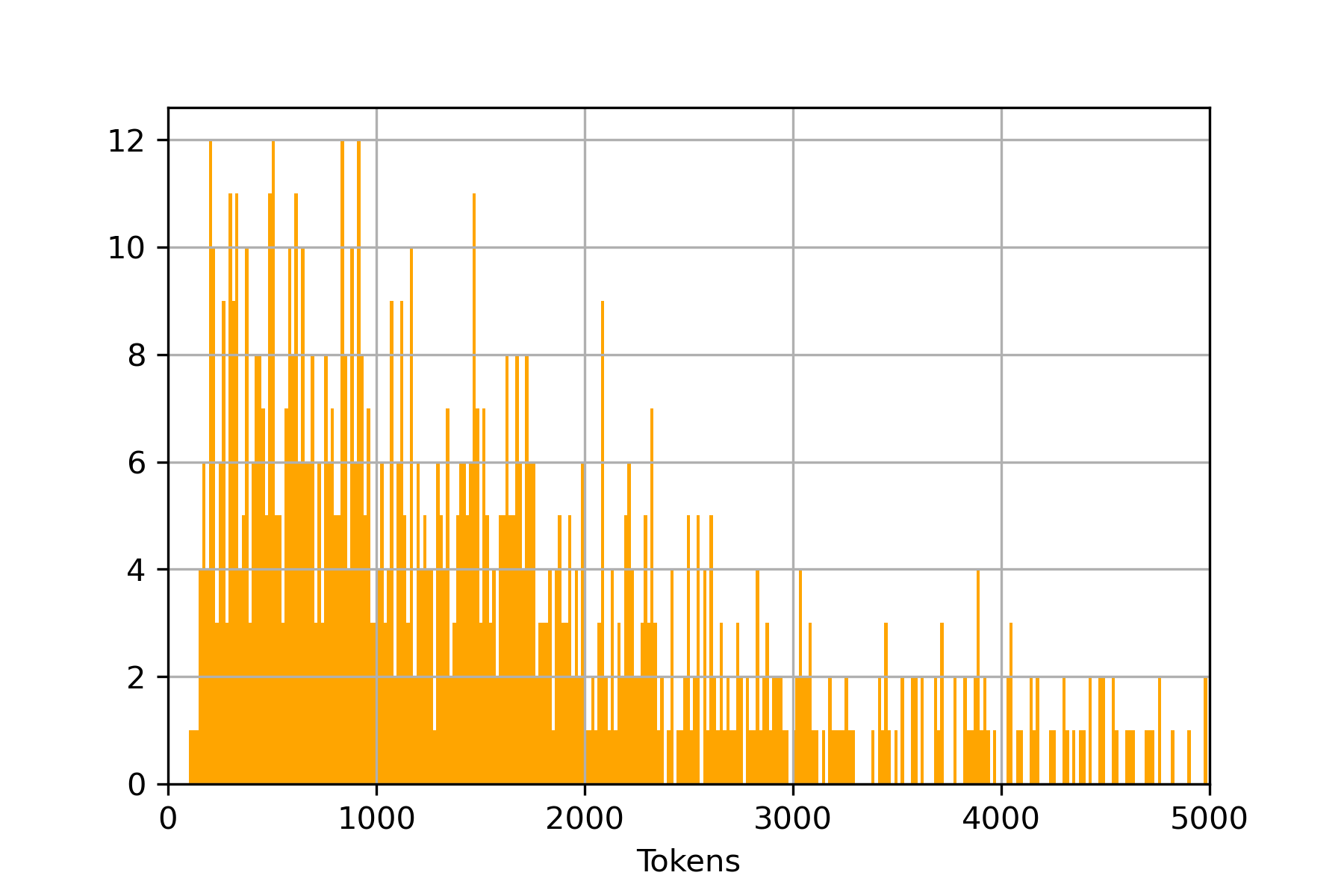} 
    }}
    \qquad
    \subfloat[
EurLex\\ 
Mean: 1871, Median: 511\\
75-Quant: 1154, 95-Quant: 8177, Max: 200749\\
    ]
    {{
    \includegraphics[width=\textwidth/2]{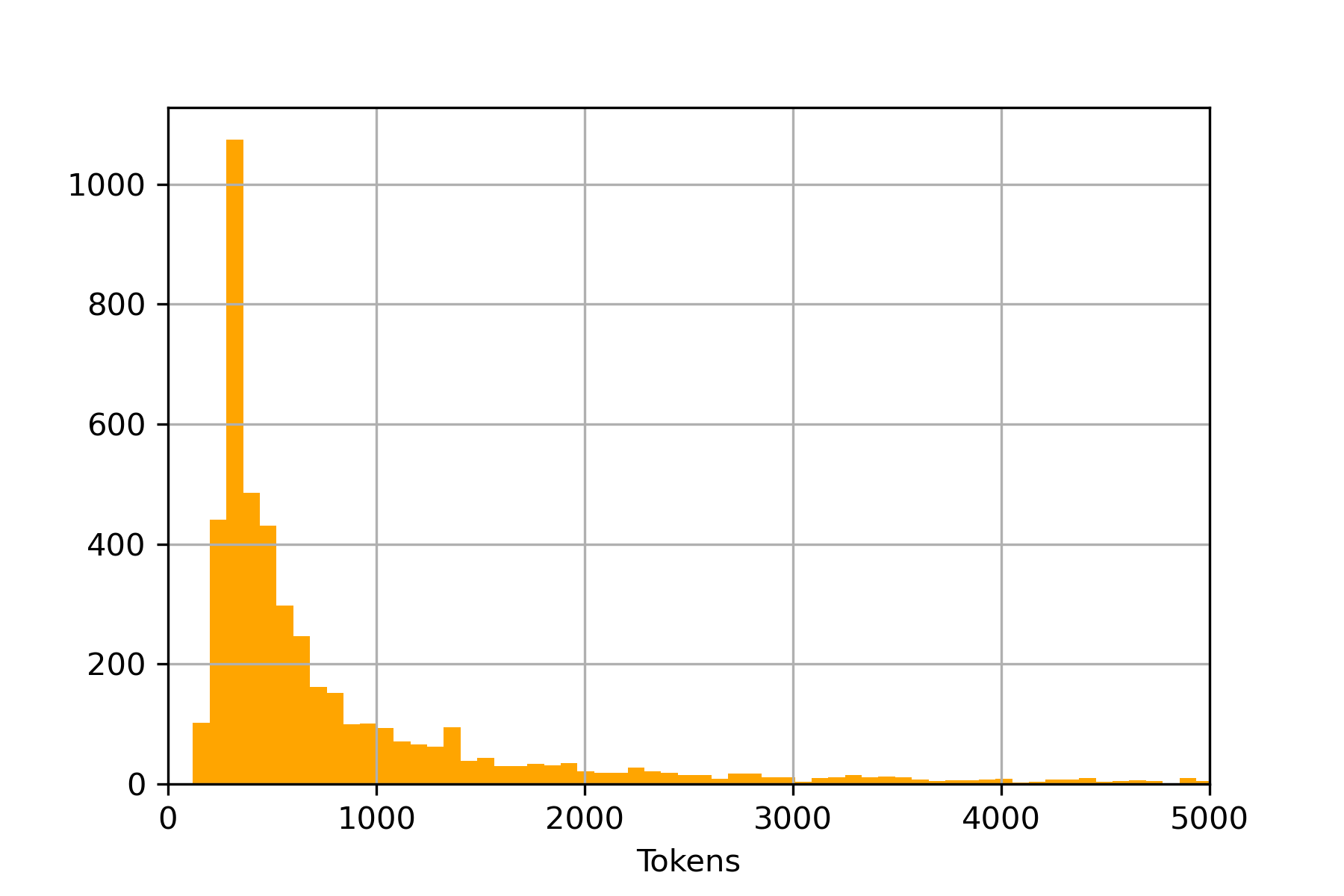} 
    }}
    }

\resizebox{\textwidth}{!}{
    \subfloat[
LEDGAR\\ 
Mean: 562, Median: 407\\
75-Quant: 723, 95-Quant: 1563, Max: 5543\\
]
    {{
    \includegraphics[width=\textwidth/2]{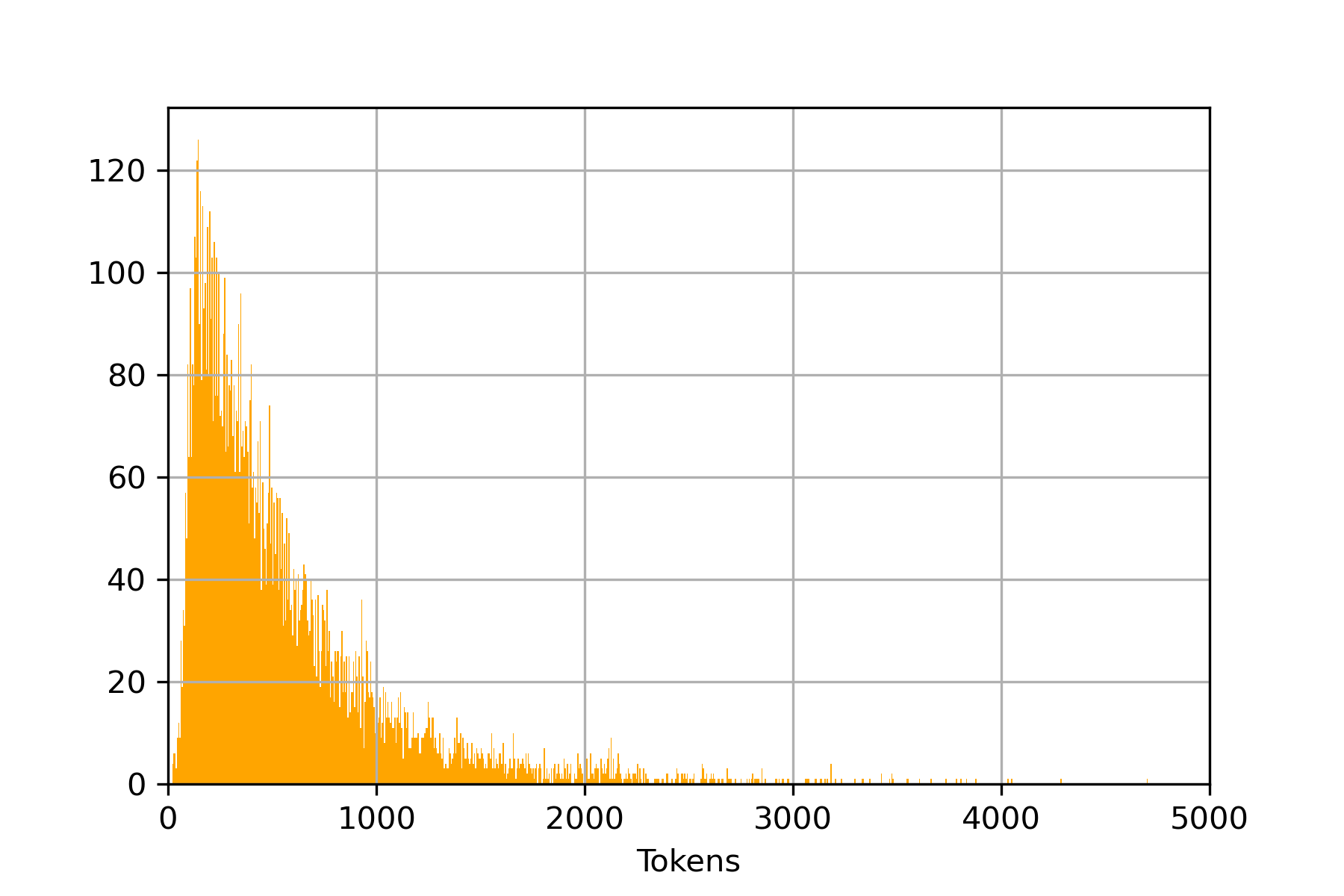} 
    }}
    \qquad
    \subfloat[
SCOTUS\\ 
Mean: 8687, Median: 6883\\
75-Quant: 11296, 95-Quant: 21075, Max: 89379\\
    ]
    {{
    \includegraphics[width=\textwidth/2]{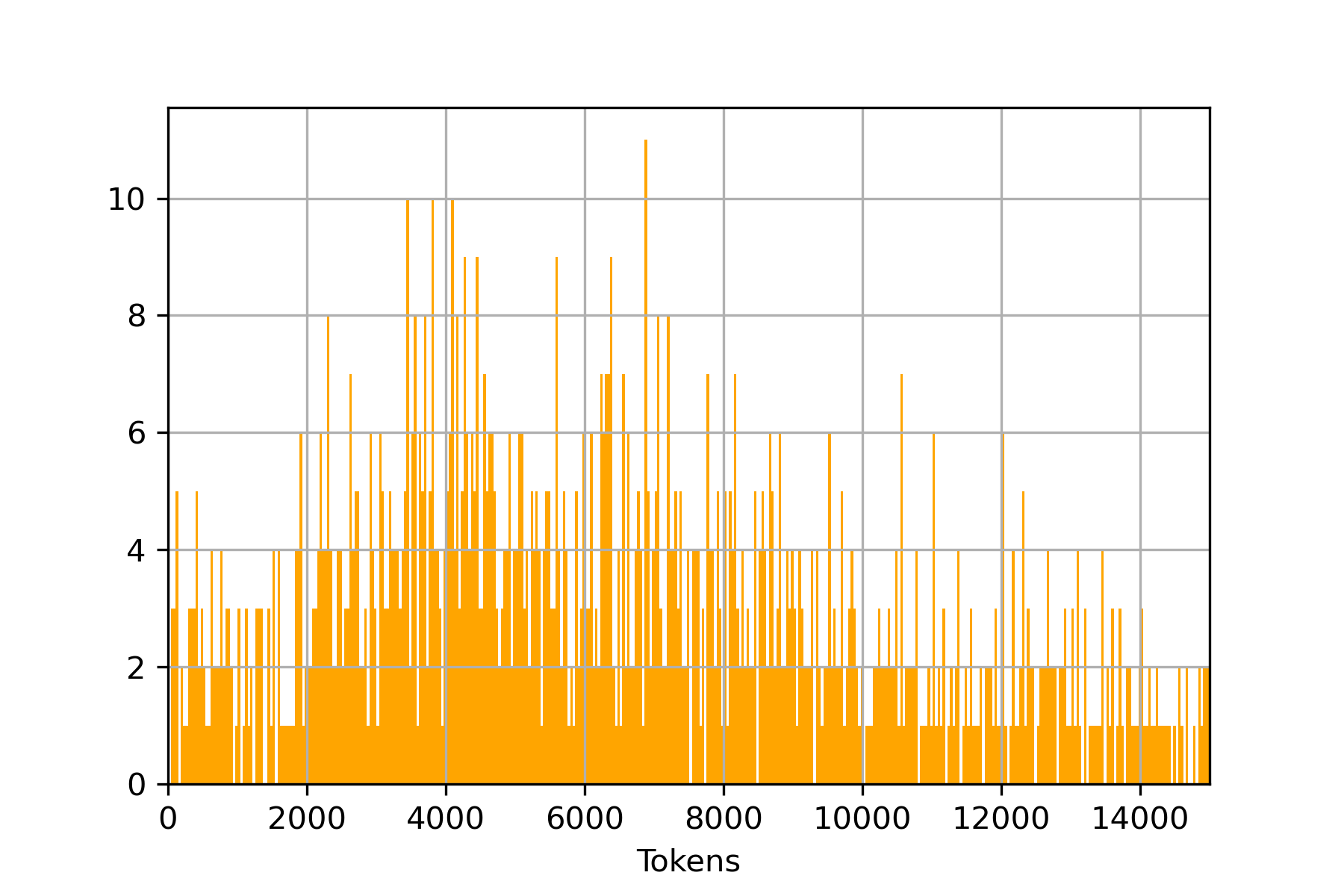} 
    }}
    }

\resizebox{\textwidth}{!}{
    \subfloat[
Unfair\_TOS\\ 
Mean: 146, Median: 120\\
75-Quant: 182, 95-Quant: 338, Max: 1368\\
]
    {{
    \includegraphics[width=\textwidth/2]{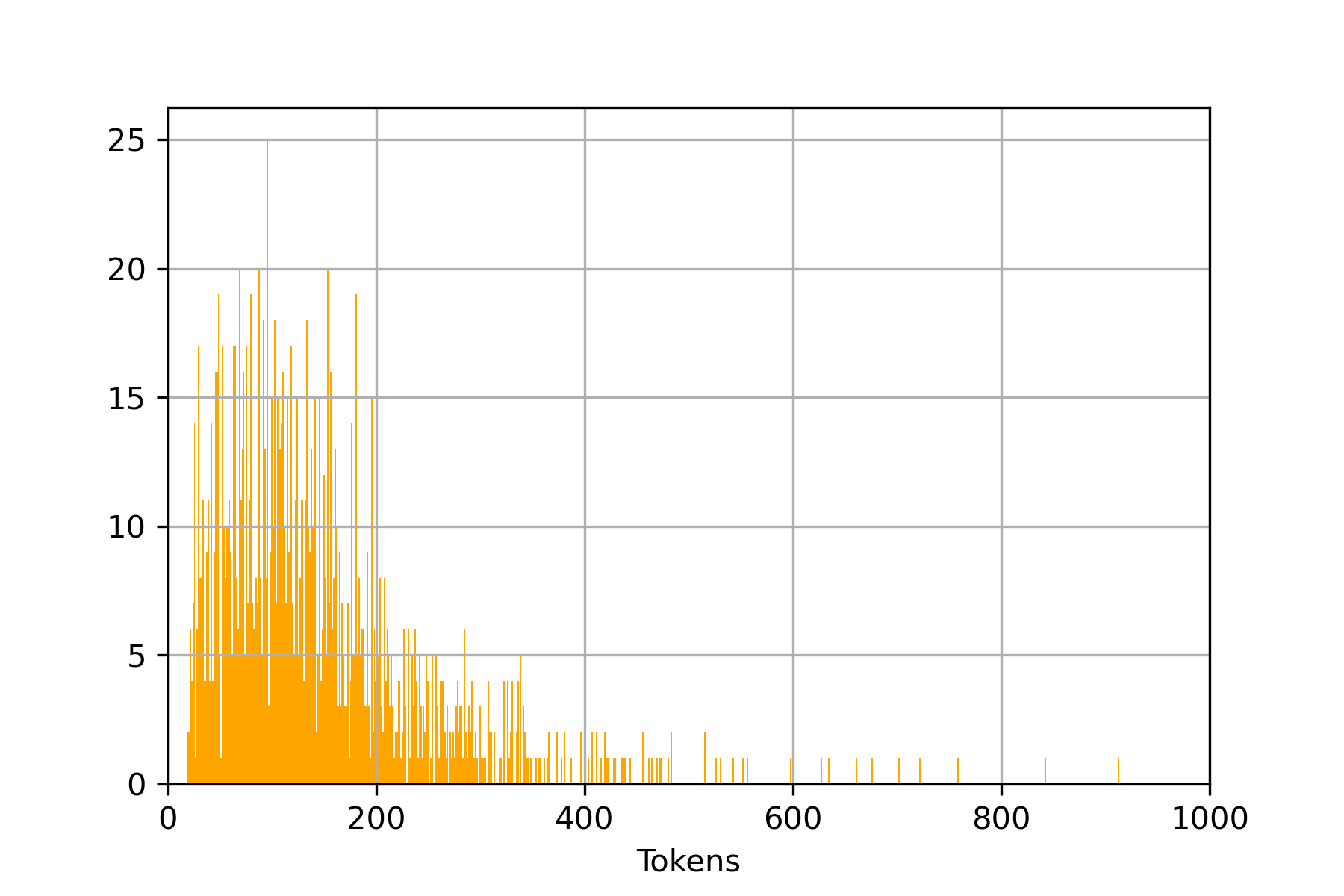} 
    }}
    \qquad

    {{
    \includegraphics[width=\textwidth/2]{data_info/empty_image.png} 
    }}
    }

    \caption{Histograms for the LexGLUE test set. }
    \label{fig:lexglue_test}
\vspace{-5mm}
\end{figure*}

In Figures \ref{fig:billsum_train}, and \ref{fig:billsum_test} we show the data length distributions for the BillSum train and test splits. In Tables \ref{fig:pubmed_train}, \ref{fig:pubmed_validation}, and \ref{fig:pubmed_test} we show the data length distributions for the PubMed train, validation and test splits.
In Tables \ref{fig:lexglue_train}, \ref{fig:lexglue_validation}, and \ref{fig:lexglue_test} we show the data length distributions for the 7 tasks of LexGLUE benchmark train, validation and test splits.

\clearpage

\clearpage

\clearpage

\clearpage

\end{document}